%% file: main.tex
\pdfoutput=1
\documentclass[11pt]{article}
\usepackage{acl}
\usepackage{times}
\usepackage{latexsym}
\usepackage[T1]{fontenc}
\usepackage[utf8]{inputenc}
\usepackage{microtype}
\usepackage{multicol}
\usepackage{multirow}
\usepackage{subcaption}
\usepackage{enumitem}
\usepackage{amsmath}
\usepackage{amssymb}
\usepackage{inconsolata}
\usepackage{graphicx}
\usepackage{booktabs}
\usepackage{fixmetodonotes}
\usepackage[title]{appendix}
\title{Can Transformer be Too Compositional?\\ Analysing Idiom Processing in Neural Machine Translation}

\author{Verna Dankers$^1$\textnormal{,} Christopher G. Lucas$^1$\textnormal{,} \textnormal{and} Ivan Titov$^{1,2}$ \\
$^1$ILCC, University of Edinburgh \\
$^2$ILLC, University of Amsterdam \\
\texttt{vernadankers@gmail.com},  \texttt{\{clucas2, ititov\}@inf.ed.ac.uk}}

\date{}

\begin{document}
\maketitle

\input{abstract}

\input{introduction}

\input{related_work}

\input{method}

\input{attention}

\input{hidden_states}

\input{amnesic_probing}

\input{conclusion}

\input{acknowledgements}

\bibliography{references}
\bibliographystyle{acl_natbib}
%TC:ignore
\appendix
\input{appendix}
%TC:ignore
\end{document}

%% file: abstract.tex
\begin{abstract}
Unlike literal expressions, idioms' meanings do not directly follow from their parts, posing a challenge for \textit{neural machine translation} (NMT). NMT models are often unable to translate idioms accurately and over-generate compositional, literal translations. In this work, we investigate whether the non-compositionality of idioms is reflected in the mechanics of the dominant NMT model, Transformer, by analysing the hidden states and attention patterns for models with English as source language and one of seven European languages as target language.
When Transformer emits a non-literal translation -- i.e. identifies the expression as idiomatic -- the encoder processes idioms more strongly as single lexical units compared to literal expressions. This manifests in idioms' parts being grouped through attention and in reduced interaction between idioms and their context.
In the decoder's cross-attention, figurative inputs result in reduced attention on source-side tokens. These results suggest that Transformer's tendency to process idioms as compositional expressions contributes to literal translations of idioms.
\end{abstract}

%% file: introduction.tex
\section{Introduction}

An idiom is a group of words of which the figurative meaning differs from the literal reading, such as ``kick the bucket,'' which means to die, instead of physically kicking a bucket. An idiom's figurative meaning is established by convention and is typically \textit{non-compositional} -- i.e. the meaning cannot be computed from the meanings of the idiom's parts.
Idioms are challenging for the task of \textit{neural machine translation} (NMT) \citep{barreiro2013multiwords,isabelle2017challenge,constant2017multiword,avramidis2019linguistic}. On the one hand, figures of speech are ubiquitous in natural language \citep{colson2019multi}. On the other hand, idioms occur much less frequently than their parts, their meanings need to be memorised due to the non-compositionality, and they require disambiguation before translation.
After all, not all \textit{potentially idiomatic expressions} (PIEs) are figurative -- e.g. consider ``When I kicked the bucket, it fell over''.
Whether PIEs should receive a figurative or literal translation depends on the context.
Yet, little is known about neural mechanisms enabling idiomatic translations and methods for improving them, other than data annotation \citep{zaninello2020multiword}.
Related work studies how idioms are represented by Transformer-based language models \citep[e.g.][]{garcia2021assessing,garcia2021probing},
but those models are not required to output a discrete representation of the idiom's meaning, which is a complicating factor for NMT models.

\begin{figure}[t]
    \centering\small
    \includegraphics[width=0.85\columnwidth]{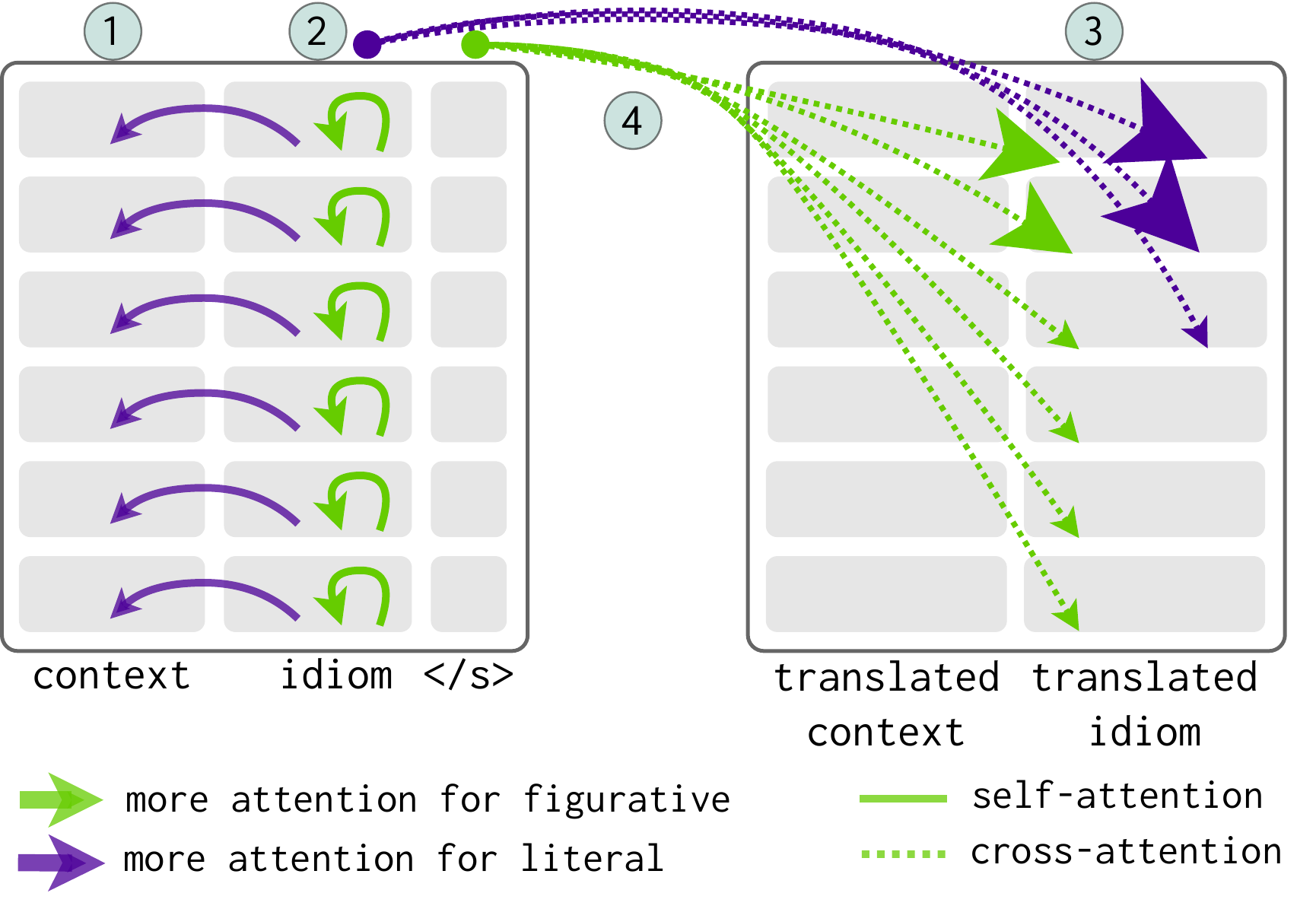}
    \caption{How do attention patterns of figurative PIEs that are paraphrased by the model compare to attention patterns of literal PIEs that are translated word for word? We find (1) decreased interaction between the PIE and its context, (2) increased attention within the PIE, (3) decreased cross-attention between the PIE and its paraphrase, (4) increased cross-attention from the paraphrase to \texttt{</s>}.}
    \label{fig:attention_flow}
    \vspace{-0.1cm}
\end{figure}

In this work, we analyse idiom processing for pre-trained NMT Transformer models \citep{vaswani2017attention} for seven European languages by comparing literal and figurative occurrences of PIEs. The comparison can help identify mechanics that underlie neural idiom processing to pave the way for methods that improve idiomatic translations.
Large-scale analyses of idiom translations suffer from a lack of parallel corpora \citep{fadaee2018examining}.
We, therefore, use a monolingual corpus, heuristically label Transformer's translations, and verify the heuristic works as intended through human evaluation, as described in \S\ref{sec:method}.
To understand how idioms are represented in Transformer, we firstly apply interpretability techniques to measure the impact of PIEs on the encoder's self-attention and the cross-attention mechanisms (\S\ref{sec:attention}), as well as the encoder's hidden representations (\S\ref{sec:hidden_states}).
Afterwards, in \S\ref{sec:amnesic_probing}, we intervene in the models while they process idiomatic expressions to show that one can change non-compositional translations into compositional ones.

The results indicate that Transformer typically translates idioms in a too compositional manner, providing a word-for-word translation.
Analyses of attention patterns  -- summarised in Figure~\ref{fig:attention_flow} -- and hidden representations point to the encoder as the mechanism grouping components of figurative PIEs.
Increased attention within the PIE is accompanied by reduced attention to context.
When translating figurative PIEs, the decoder relies less on the encoder's output than for literal PIEs. These patterns are stronger for figurative PIEs that the model paraphrases than for sentences that receive an overly compositional translation and hold across the seven European languages.
Considering that a recent trend in NLP is to encourage even more compositional processing in NMT \citep[][i.a.]{raunak2019compositionality, chaabouni2021can, li-etal-2021-compositional}, we recommend caution.
It may be beneficial to evaluate the effect of compositionality-favouring techniques on non-compositional phenomena like idioms to ensure their effect is not detrimental.

%% file: related_work.tex
\section{Related Work}\label{sec:related_work}

This section summarises work discussing human idiom comprehension, interpretability studies for NMT, and literature about figurative language processing in Transformer.

\paragraph{Idiom comprehension} Historically, idioms were considered non-compositional units \citep{swinney1979access}. Two main views (\textit{literal first} and \textit{direct access}) existed for how humans interpreted them. The former suggests humans attempt a compositional interpretation before considering the figurative interpretation in case of a contextual discrepancy \citep{bobrow1973catching,grice1975logic,grice1989studies}. The latter view suggests one can immediately retrieve the non-compositional meaning \citep{gibbs1994poetics}.
The more recent \textit{hybrid view} posits that idioms are simultaneously processed as a whole -- primed by a \textit{superlemma} \citep{kuiper2007slipping} -- and word for word \citep{caillies2007processing}. The processing speed and retrieval of the figurative meaning depend on the idiom's semantic properties and the context \citep{cain2009development,vulchanova2019boon}. Examples of semantic properties are the conventionality and decomposability of idioms \citep{nunberg1994idioms}.
We do not expect processes in Transformer to resemble idiom processing in humans.
Nonetheless, this work helps us determine our focus of study on the role of the surrounding context and the extent to which idioms' parts are processed as a whole.

Translating PIEs that are used figuratively is not always straightforward.
\citet{baker1992other} discuss strategies for human translators:
(i) Using an idiom from the target language of similar meaning and form, (ii) using an idiom from the target language with a similar meaning and a different form, (iii) copying the idiom to the translation, (iv) paraphrasing the idiom or (v) omitting it. In the absence of idioms with similar meanings across languages, (iv) is the most common strategy.
Our main focus is on literal translations (\textbf{word-for-word} translations), and \textbf{paraphrases}.

\paragraph{Interpreting Transformer}
Analyses of Transformer for NMT studied the encoder's hidden representations and self-attention mechanism \citep[e.g.][]{raganato2018analysis,tang2019understanding,voita2019bottom}, the cross-attention \citep[e.g.][]{tang2019encoders} and the decoder \citep[e.g.][]{yang2020sub}. The encoder is particularly important for the contextualisation of tokens from the source sentence; it acts as a feature extractor \citep{tang2019understanding}.
The encoder's bottom three layers better represent low-level syntactic features, whereas the top three layers better capture semantic features \citep{raganato2018analysis}.
As a result, one would expect the representations in higher layers to be more representative of idiomaticity.

Idioms are a specific kind of ambiguity, and whether a word is ambiguous can accurately be predicted from the encoder's hidden representations, as shown by \citet{tang2019encoders} for ambiguous nouns.
Transformer's cross-attention is not crucial for disambiguating word senses \citep{tang2018analysis}, but the encoder’s self-attention does reflect ambiguity through more distributed attention for ambiguous nouns \citep{tang2019encoders}.

\paragraph{Tropes in Transformer}
Various studies examine the Transformer-based language model BERT's \citep{devlin2019bert} ability to capture tropes like metonyms \citep{pedinotti2020don}, idioms \citep{kurfali2020disambiguation}, and multiple types of figurative language \citep{shwartz2019still}. \citet{kurfali2020disambiguation} detect idioms based on the dissimilarity of BERT's representations of a PIE and its context, assuming that contextual discrepancies indicate figurative usage. \citet{pedinotti2020don} measure whether BERT detects meaning shift for metonymic expressions but find cloze probabilities more indicative than vector similarities. 
\citet{shwartz2019still} find that BERT is better at detecting figurative meaning shift than at predicting implicit meaning -- e.g. predicting that ``a hot argument'' does not involve temperature.

The most recent work studies properties of hidden representations of noun-noun compounds (NCs) and verb-noun compounds (VCs): \citet{garcia2021probing} examine (contextualised) word embeddings, including BERT, to compare figurative and literal NC \textit{types}. They investigate the similarities between (1) NCs and their synonyms, (2) NCs and their components, (3) in-context and out-of-context representations, and (4) the impact of replacing one component in the NC.
Surprisingly, idiomatic NCs are quite similar to their components and are less similar to their synonym compared to literal NCs.
Moreover, the context of the NC hardly contributes to how indicative its representation is of idiomaticity, which was also shown by \citet{garcia2021assessing}, who measured the correlation between \textit{token}-level idiomaticity scores and NCs' similarity in- and out-of-context.

In search of the \textit{idiomatic key} of VCs (the part of the input that cues idiomatic usage), \citet{nedumpozhimana2021finding} train a probing classifier to distinguish literal usage from figurative usage. They then compare the impact of masking the PIE to masking the context on the classifier's performance and conclude that the idiomatic key mainly lies within the PIE itself, although there is some information coming from the surrounding context.

%% file: method.tex
\section{Method}
\label{sec:method}
We use Transformer models \citep{vaswani2017attention} with English as the source language and one of seven languages as the target language (Dutch, German, Swedish, Danish, French, Italian, Spanish).\footnote{Our figures refer to these languages using their ISO 639-1 codes, that are \texttt{nl}, \texttt{de}, \texttt{sv}, \texttt{da}, \texttt{fr}, \texttt{it} and \texttt{es}, respectively.}
Transformer contains encoder and decoder networks with six self-attention layers each and eight heads per attention mechanism. The models are pre-trained by \citet{tiedemann2020opus} with the Marian-MT framework \citep{junczys2018marian} on a collection of corpora (OPUS).\footnote{The models are available via the \href{https://huggingface.co/transformers/}{\texttt{transformers} library} \citep{wolf-etal-2020-transformers}.}
We extract hidden states and attention patterns for sentences with PIEs.
The analyses presented are detailed for Dutch, after which we explain how the results for the other languages compare to Dutch.\footnote{
The data and code are available via the \href{https://github.com/vernadankers/mt_idioms}{\texttt{mt\_idioms} github repository}.}

Parallel PIE corpora are rare, exist for a handful of languages only, and are limited in size \citep{fadaee2018examining}. Rather than rely on a small parallel corpus, we use the largest corpus of English PIEs to date and annotate the translations heuristically. This section provides corpus statistics and discusses the heuristic annotation method.

\paragraph{MAGPIE corpus}
\label{subsec:data}
The MAGPIE corpus presented by \citet{haagsma2020magpie} contains 1756 English idioms from the Oxford Dictionary of English with 57k occurrences.
MAGPIE contains identical PIE matches and morphological and syntactic variants, through the inclusion of
common modifications of PIEs, such as passivisation (``the beans were spilled'') and word insertions (``spill all the beans'').\footnote{Available via the \href{https://www.github.com/hslh/magpie-corpus}{\texttt{MAGPIE} github repository}.}
We use 37k samples annotated as fully \textbf{figurative} or \textbf{literal}, for 1482 idioms that contain nouns, numerals or adjectives that are colours (which we refer to as \textbf{keywords}).
Because idioms show syntactic and morphological variability, we focus mostly on the nouns. Verbs and their translation are harder to identify due to the variability. Moreover, idiom indexes are also typically organised based on the nominal constituents, instead of the verbs \citep{piirainen2013widespread}.
Only the PIE and its sentential context are presented to the model.
We distinguish between PIEs and their context using the corpus's word-level annotations.

\paragraph{Heuristic annotation method}
The MAGPIE sentences are translated by the models with beam search and a beam size of five.
The translations are labelled heuristically.
In the presence of a literal translation of at least one of the idiom's keywords, the entire translation is labelled as a \textbf{word-for-word} translation, where the literal translations of keywords are extracted from the model and Google translate.
When a literally translated keyword is not present, it is considered a \textbf{paraphrase}.\footnote{The annotation does not evaluate whether paraphrases are correct, which requires expert idiom knowledge in both languages. A paraphrase being provided is a first step to adequately translating idioms and, at present, the only way to detect how the model approaches the task for large datasets.}
\citet{shao2018evaluating} previously analysed NMT translations of 50 Chinese idioms using a similar method and manually curated lists of literal translations of idioms' words to detect literal translation errors. \citet{dankers2022paradox} use a similar method for 20 English idioms, to track when a word-for-word translation changes into a paraphrased one during training for an English-Dutch (\texttt{En-Nl}) NMT model.

\begin{table}
    \centering\small\setlength{\tabcolsep}{4pt}
    \resizebox{\columnwidth}{!}{\begin{tabular}{l|cccccccc}
    \toprule
    \textbf{Category} & \texttt{nl} & \texttt{de} & \texttt{sv} & \texttt{da} & \texttt{fr} & \texttt{it} & \texttt{es} \\\midrule
    Figurative, paraphrase    & 20 & 20 & 24 & 18 & 19 & 20 & 24 \\
    Figurative, word for word & 80 & 80 & 76 & 82 & 81 & 80 & 76 \\\midrule
    Literal, paraphrase       & 5  & 6  & 8  & 5  & 7  & 9  & 7  \\
    Literal, word for word    & 95 & 94 & 92 & 95 & 93 & 91 & 93 \\
    \bottomrule
    \end{tabular}}
    \caption{Distribution of the heuristically assigned labels for translations of MAGPIE sentences in percentages, expressed within category (figurative / literal).}
    \label{tab:distribution}
    \vspace{-0.1cm}
\end{table}

\begin{figure}
    \centering
    \includegraphics[width=0.49\columnwidth]{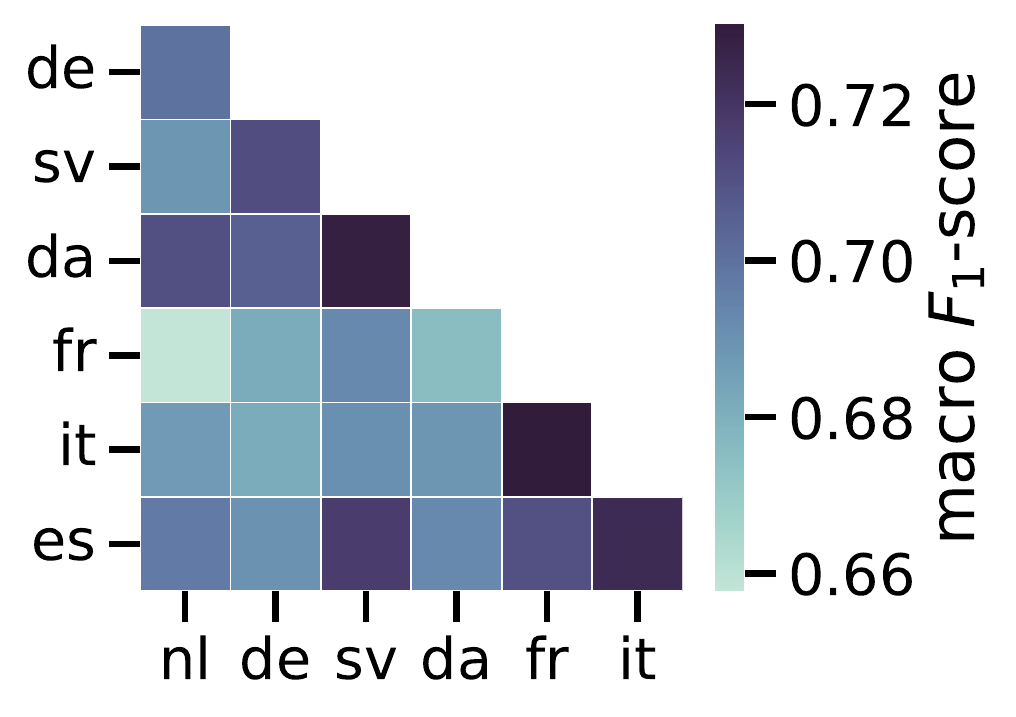}
    \includegraphics[width=0.49\columnwidth]{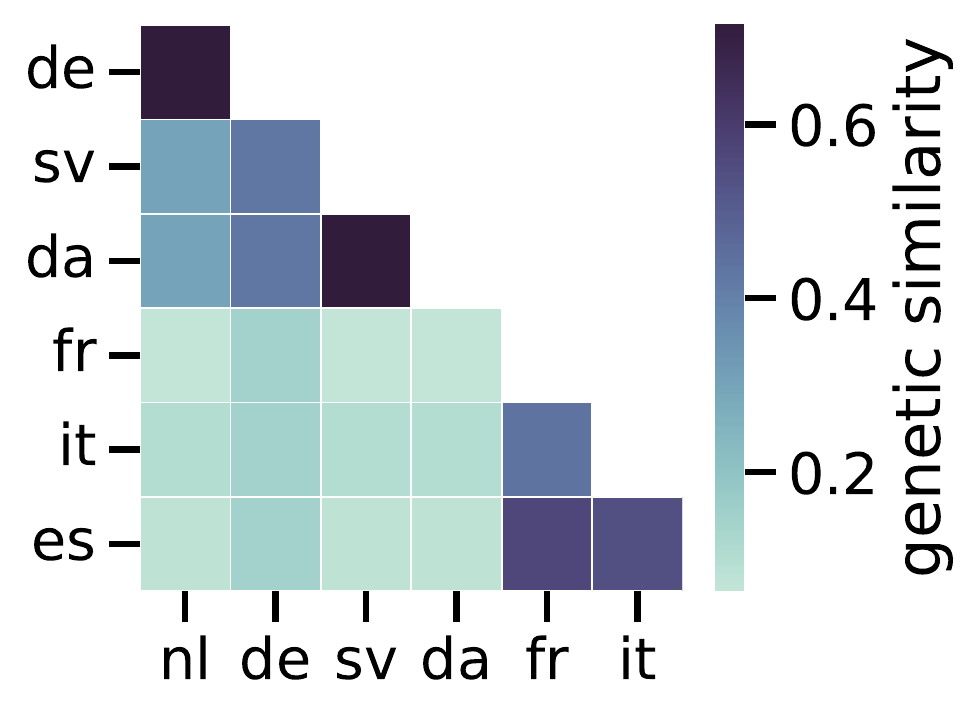}
    \caption{The macro-averaged $F_1$-score of translation labels (paraphrase vs word for word) for figurative PIEs and languages' genetic similarity visualised (Pearson's $r{=}0.61$,  $p<0.005$).}
    \label{fig:agreement}
    \vspace{-0.1cm}
\end{figure}

Table~\ref{tab:distribution} summarises the distribution of these categories for all languages, for the subsets of figurative and literal examples from MAGPIE.
Generally, paraphrased translations of figurative PIEs are more appropriate than word-for-word translations, whereas literal PIEs can be translated word for word \citep{baker1992other}. The vast majority of literal PIEs indeed result in word-for-word translations.
The subset of figurative samples results in more paraphrases, but $\geq 76\%$ is still a word-for-word translation, dependent on the language.
Although the statistics are similar across languages, there are differences in which examples are paraphrased.
Figure~\ref{fig:agreement} illustrates the agreement by computing the $F_1$-score when using the predictions for figurative instances of one language as the target, and comparing them to predictions from another language. The agreement positively correlates with genetic similarity as computed using the Uriel database \citep{littell2017uriel}.

\begin{table}
    \centering\small\setlength{\tabcolsep}{3pt}
    \begin{tabular}{l|c|cccccccc}
    \toprule
    \textbf{Category} & \# & \texttt{nl} & \texttt{de} & \texttt{sv} & \texttt{da} & \texttt{fr} & \texttt{it} & \texttt{es} \\\midrule
    Fig., paraphrase     & 116 & 88 & 84 & 75 & 81 & 78 & 78  & 87  \\
    Fig., word for word  & 103 & 95 & 92 & 95 & 74 & 96 & 97  & 82 \\\midrule
    Lit., paraphrase     & 28  & 54 & 71 & 43 & 82 & 43 & 32  & 50 \\
    Lit., word for word  & 103 & 98 & 89 & 97 & 89 & 98 & 100 & 94 \\
    \bottomrule
    \end{tabular}
    \caption{Survey statistics: the number of sentence pairs used (\#), and the percentage of labels for which the annotator and the algorithm agreed per language.}
    \label{tab:study_stats}
    \vspace{-0.1cm}
\end{table}

To assess the quality of the heuristic method, one (near) native speaker per target language annotated 350 samples, where they were instructed to focus on one PIE keyword in the English sentence.
Annotators were asked whether (1) the English word was present in the translation (initially referred to as ``copy''), (2) whether there was a literal translation for the word, or (3) whether neither of those options were suited, referred to as the ``paraphrase''.\footnote{Annotators were not involved in the research. Except for Swedish, annotators were native in the target language. For ethical considerations and more details, see Appendix~\ref{ap:survey}.}
Due to the presence of cognates in the ``copy'' category, that category was merged with the ``word for word'' category after the annotation.
Table~\ref{tab:study_stats} summarises the accuracies obtained. Of particular interest are samples that are figurative and paraphrased, since they represent the translations that are treated non-compositionally by the model, as well as instances that are literal and translated word for word, since they represent the compositional translations for non-idiomatic PIE occurrences.
These categories have annotation accuracies of $\geq 75\%$ and $\geq 89\%$, respectively.
During preliminary analyses, an annotation study was conducted for Dutch by annotators from the crowd-sourcing platform Prolific.
The annotators and the heuristic method agreed in 83\% of the annotated examples, and for 77\% of the samples an average of 4 annotators agreed on the label unanimously (see Appendix~\ref{ap:survey} for more details).

Sentences containing idioms typically yield lower BLEU scores \citep{fadaee2018examining}.
MAGPIE is a monolingual corpus and does not allow us to compute BLEU scores, but we refer the reader to Appendix~\ref{ap:idioms_in_opus} for an exploratory investigation for MAGPIE's idioms using the \texttt{En-Nl} training corpus.\looseness=-1

%% file: attention.tex
\section{Attention}
\label{sec:attention}
We now turn to comparing how literal and figurative PIEs are processed by Transformer.
Whether a PIE is figurative depends on the context -- e.g. compare 
``in culinary school, I felt \textit{at sea}'' to ``the sailors were \textit{at sea}''.
Within Transformer, contextualisation of input tokens is achieved through the attention mechanisms, which is why they are expected to combine the representations of the idioms' tokens and embed the idiom in its context.
This section discusses the impact of PIEs on the encoder's self-attention and the encoder-decoder cross-attention.
To assert that the conclusions drawn in this section are not simply explained by shallow statistics of the data used, we recompute the results in Appendix~\ref{ap:attention} for (1) a data subset excluding variations of PIEs' standard surface forms, (2) a data subset that includes PIEs that appear in both figurative and literal contexts, (3) a data subset that controls for the number of tokens within a PIE.
Qualitatively, these results lead to the same findings.

\begin{figure}[!t]
\hspace{0.2cm}\begin{subfigure}[b]{0.435\columnwidth}\centering
    \includegraphics[width=\textwidth]{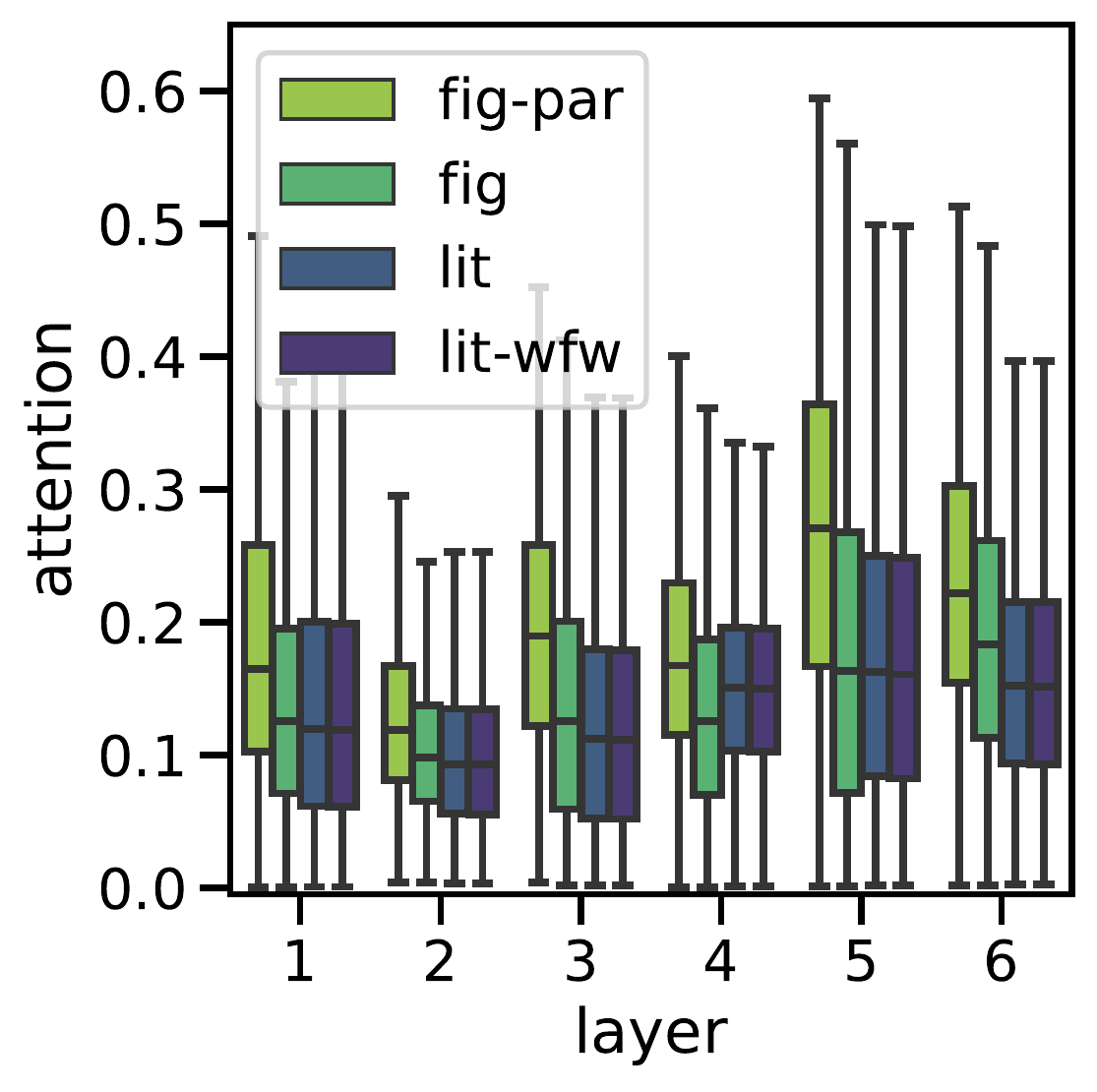}
    \caption{PIE to PIE}
    \label{fig:att_idi2idi}
\end{subfigure}
\begin{subfigure}[b]{0.46\columnwidth}\centering
    \includegraphics[width=\textwidth]{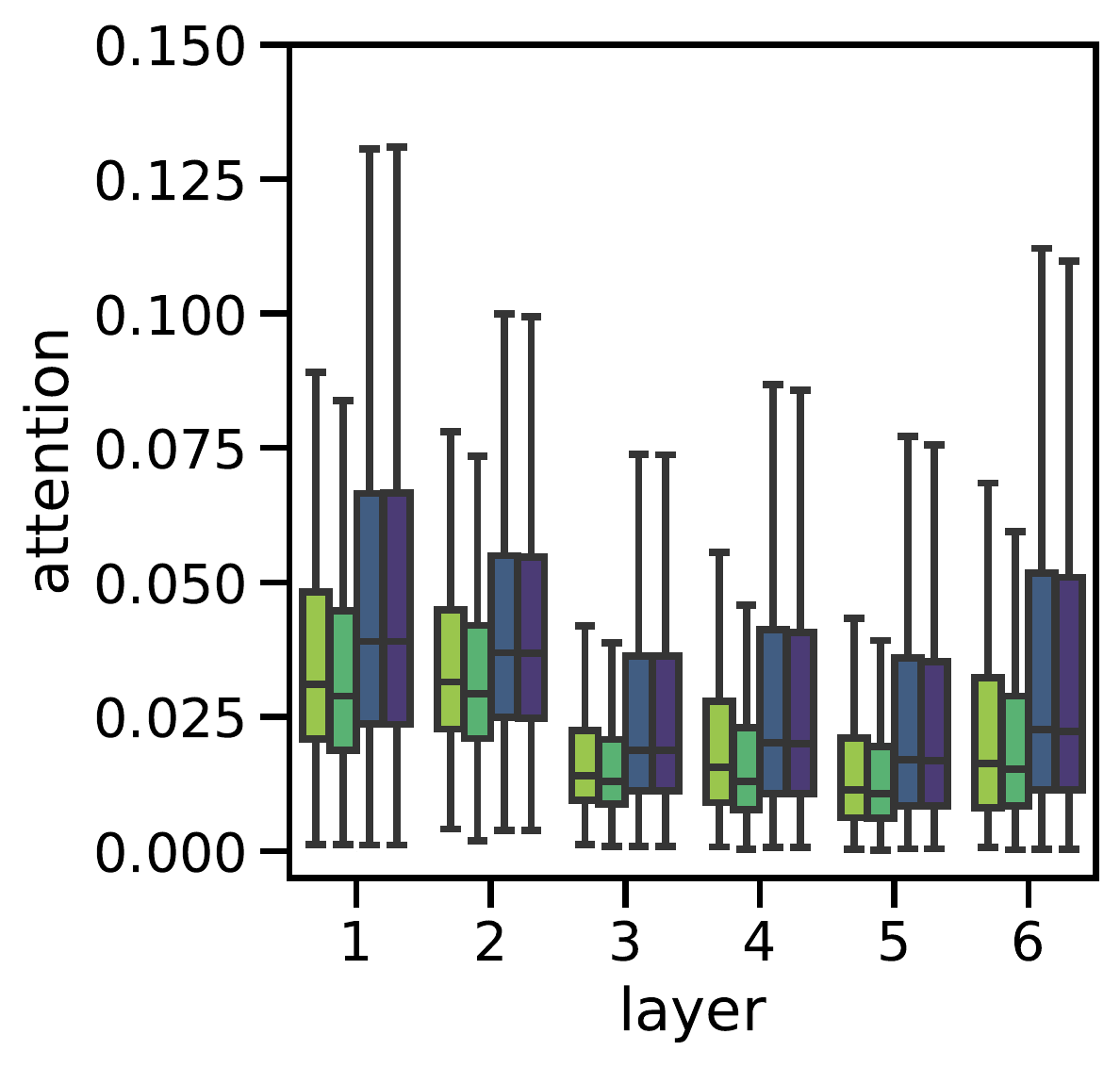}
    \caption{PIE to context}
    \label{fig:att_idi2con}
\end{subfigure}

\begin{subfigure}[b]{0.46\columnwidth}\centering
    \includegraphics[width=\textwidth]{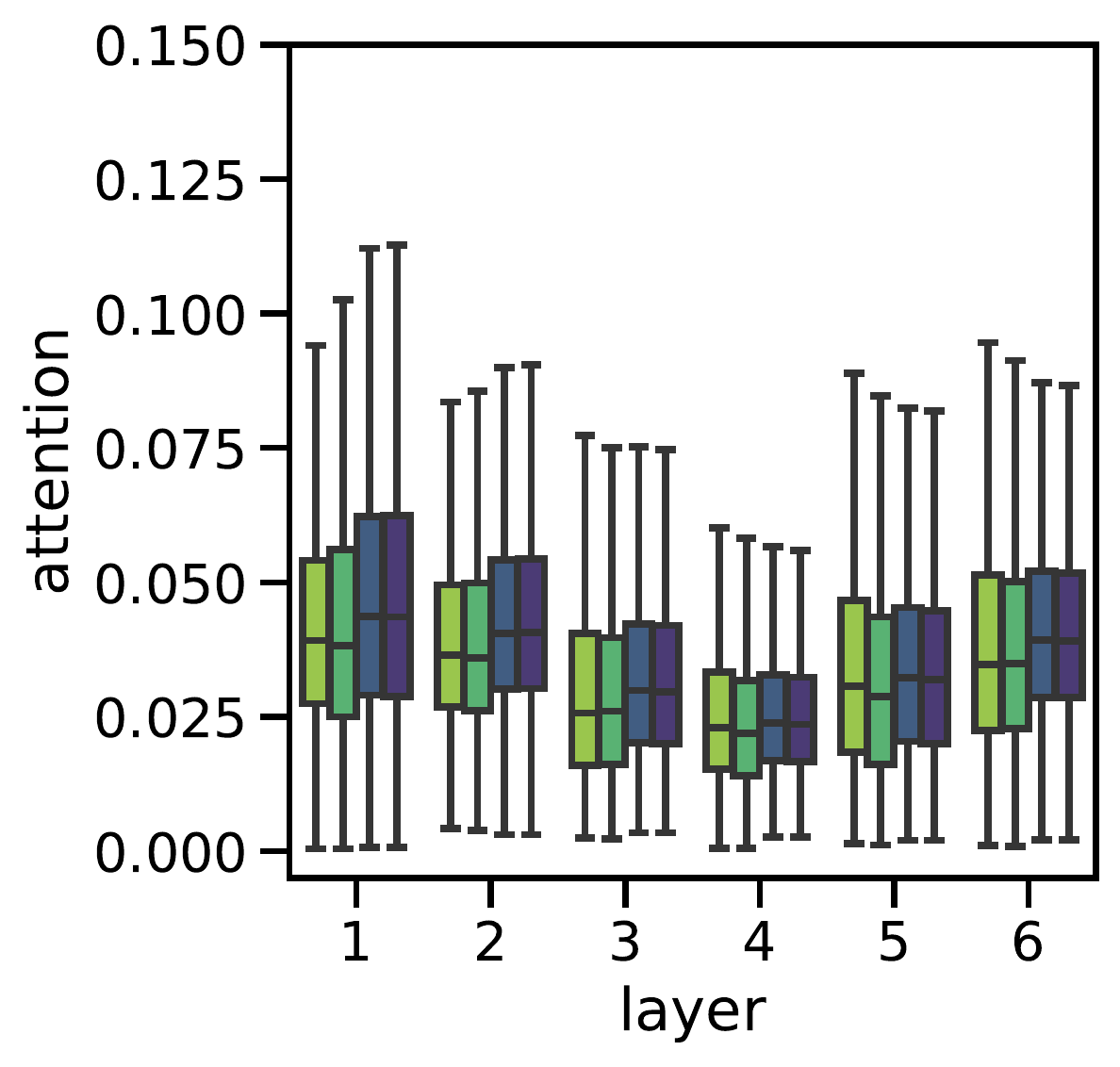}
    \caption{Context to PIE}
    \label{fig:att_con2idi}
\end{subfigure}
\begin{subfigure}[b]{0.52\columnwidth}\centering
    \includegraphics[width=\textwidth]{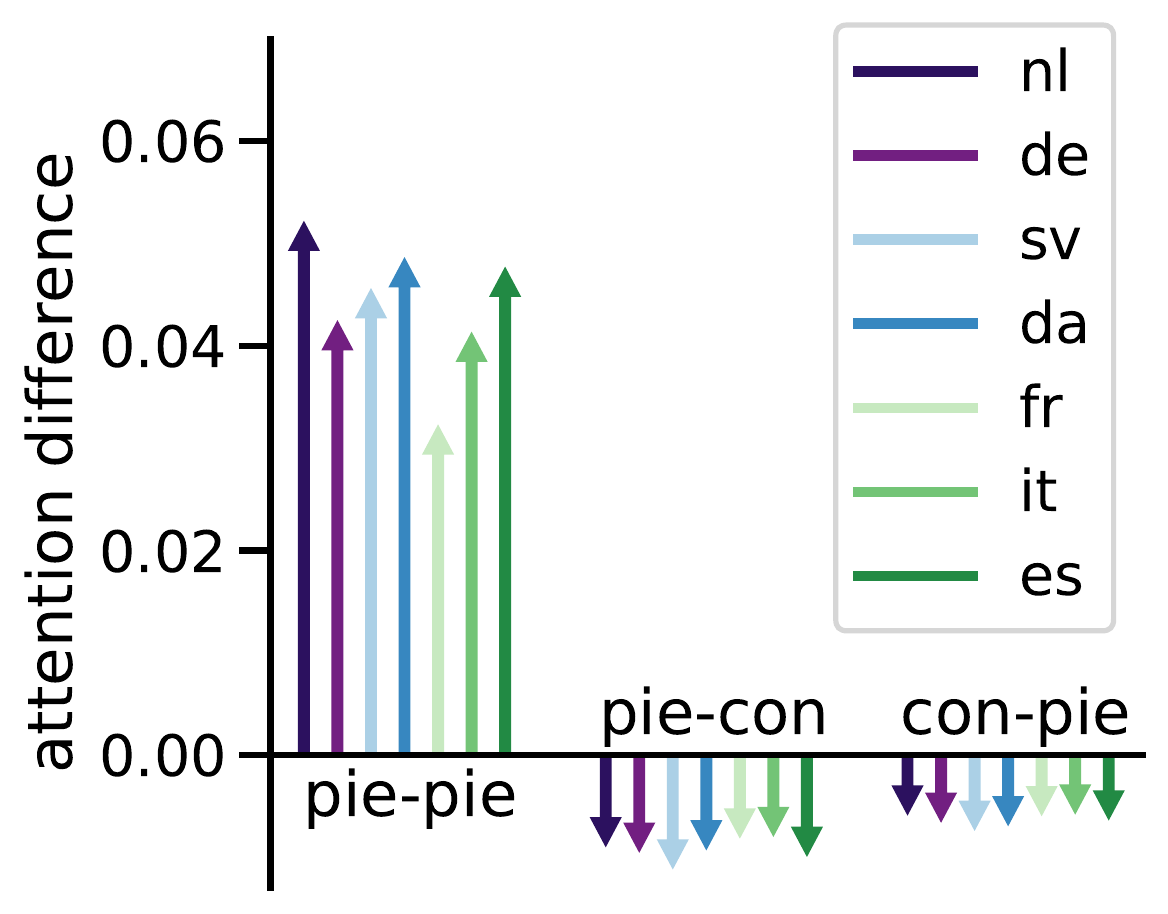}
    \caption{Language comparison}
    \label{fig:att_langs}
\end{subfigure}
\caption{Weight distributions of the encoder's self-attention (a-c), and the mean difference of \textit{fig-par} and \textit{lit-wfw} for all languages (d).
Boxes represent quartiles; whiskers show the distribution, excluding outliers.
}
\label{fig:encoder_attention}
\end{figure}

\begin{figure}
\begin{subfigure}[b]{0.435\columnwidth}\centering
    \includegraphics[width=\textwidth]{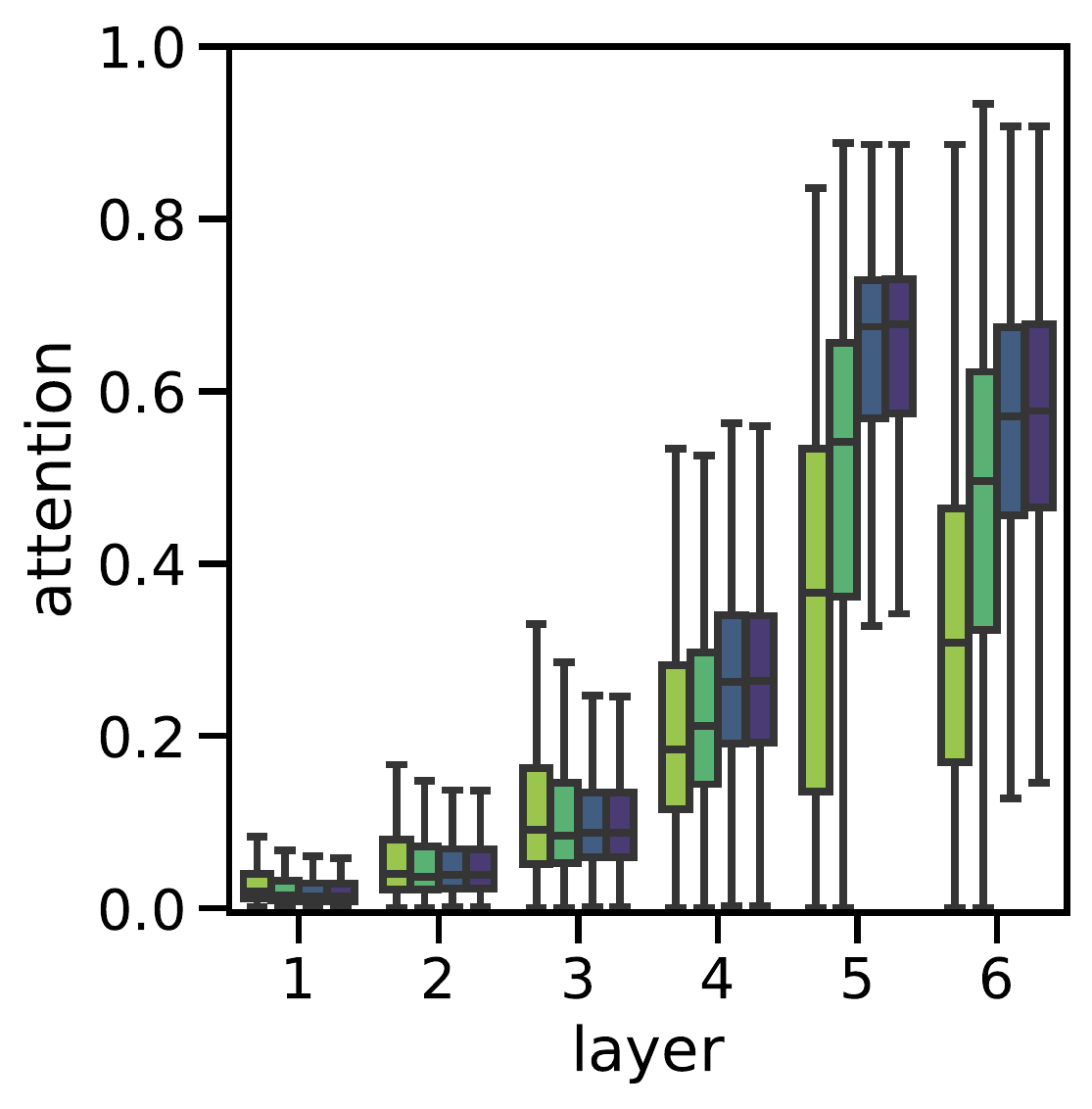}
    \caption{Target to PIE noun}
    \label{fig:cross_att_idi2idi}
\end{subfigure}
\begin{subfigure}[b]{0.465\columnwidth}\centering
    \includegraphics[width=0.95\textwidth]{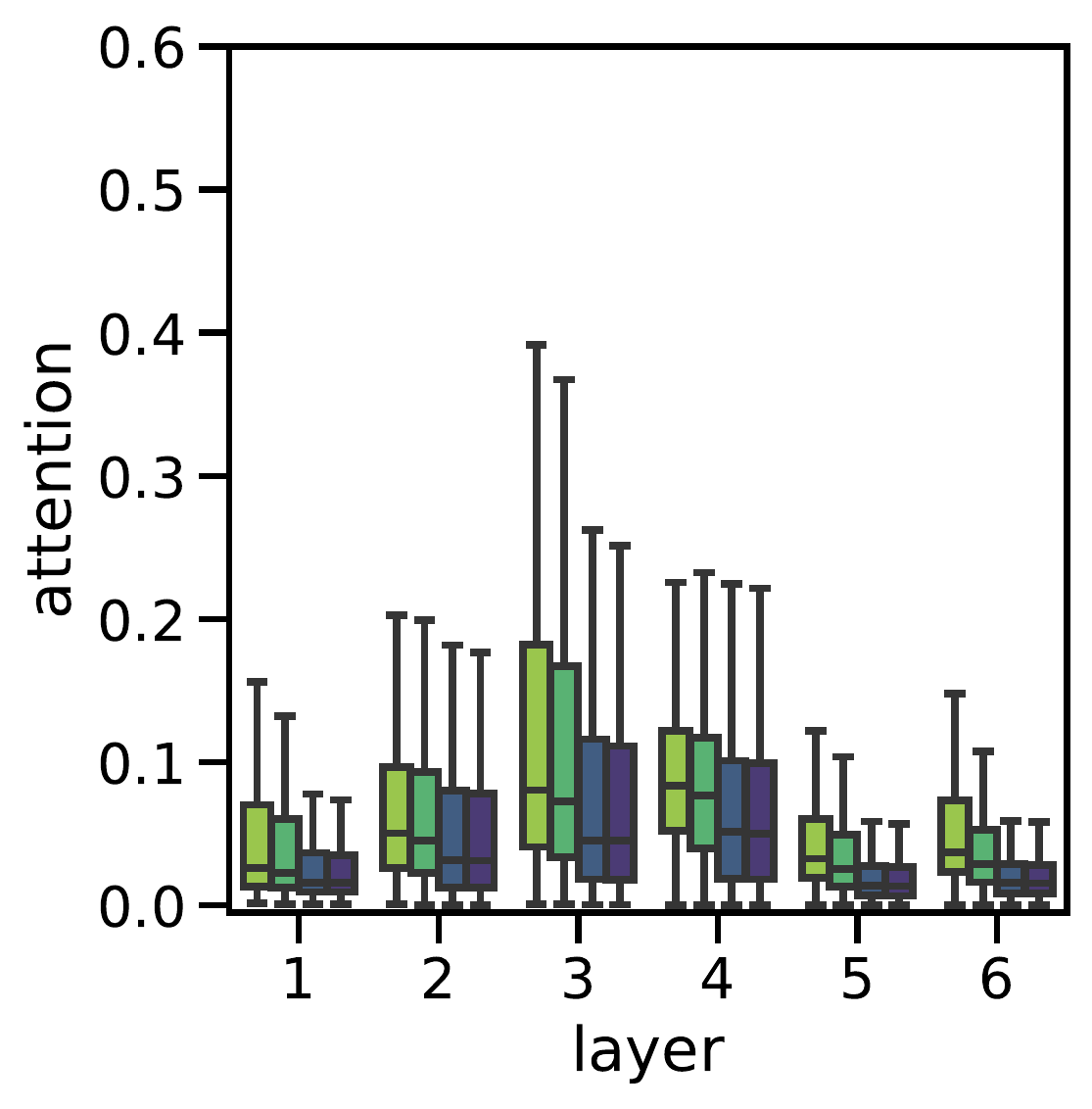}
    \caption{Target to other PIE tokens}
    \label{fig:cross_att_idi2con}
\end{subfigure}
\begin{subfigure}[b]{0.44\columnwidth}\centering
    \includegraphics[width=\textwidth]{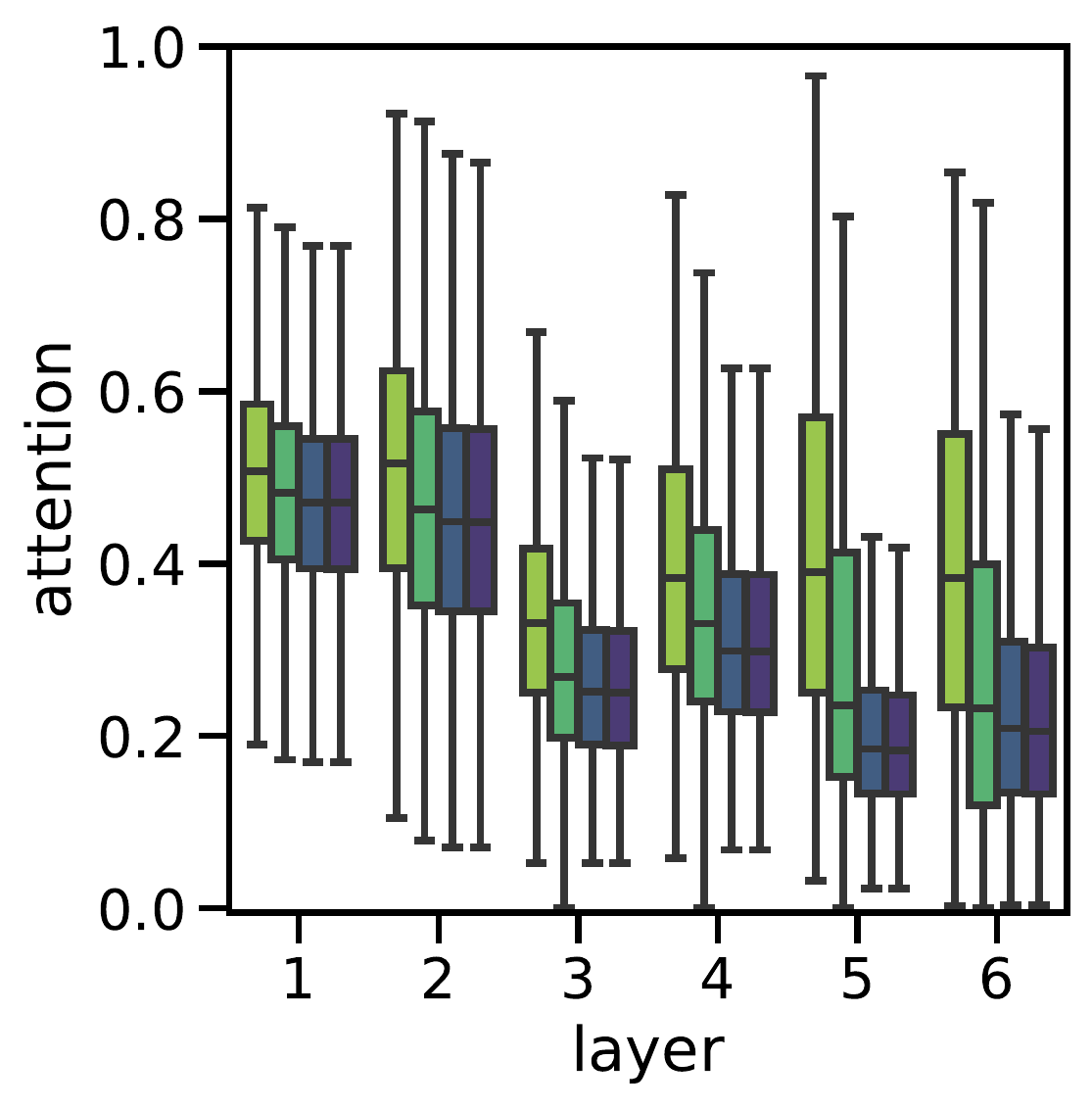}
    \caption{Target to \texttt{</s>}}
    \label{fig:cross_att_idi2eos}
\end{subfigure}
\begin{subfigure}[b]{0.52\columnwidth}\centering
    \includegraphics[width=\textwidth]{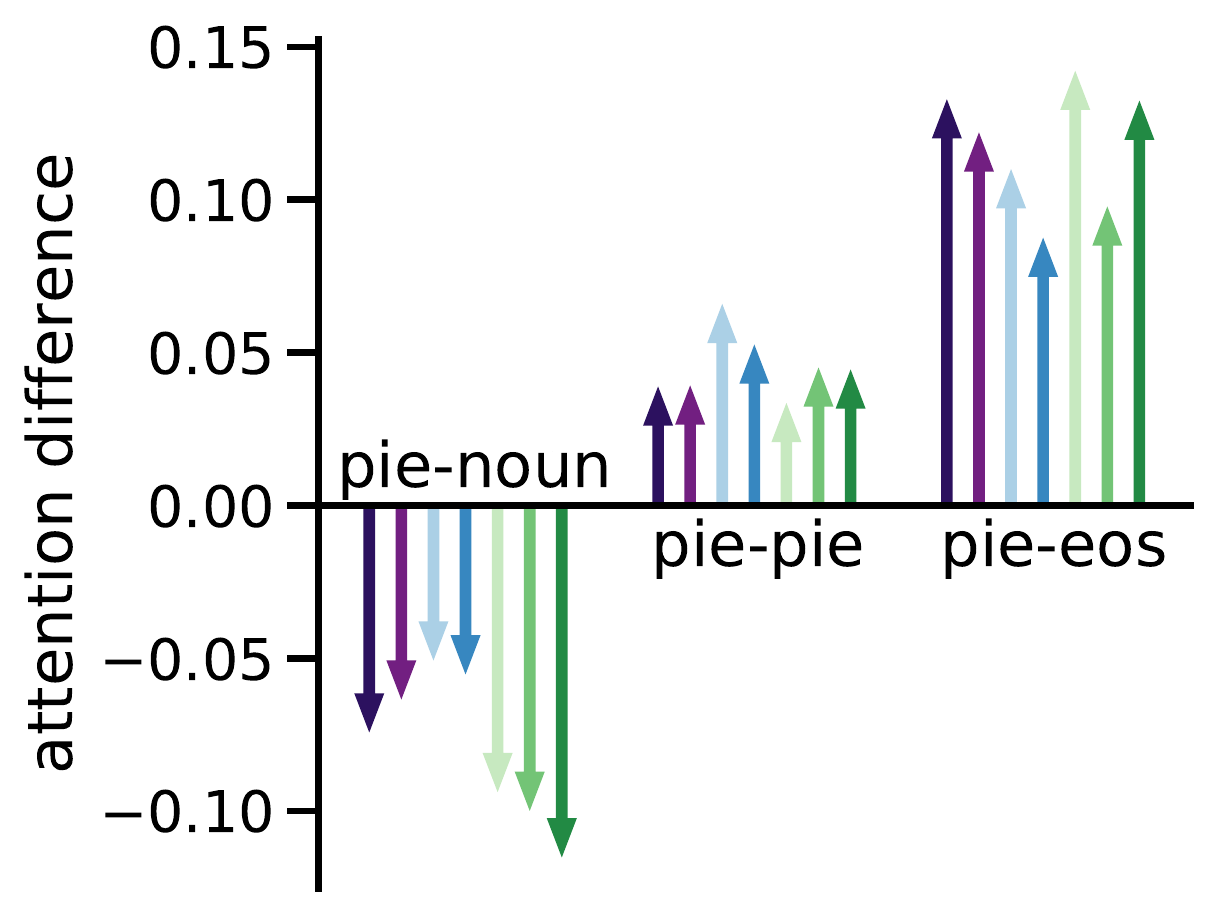}
    \caption{Language comparison}
    \label{fig:cross_att_langs}
\end{subfigure}
\caption{The cross-attention for target-side tokens aligned to PIE nouns (a-c), and the mean difference between \textit{fig-par} and \textit{lit-wfw} for all languages (d).}
\label{fig:cross_attention}
\end{figure}

\paragraph{Attention within the PIE}
For the \texttt{En-Nl} Transformer, Figure~\ref{fig:att_idi2idi} visualises the distribution of attention weights in the encoder's self-attention mechanism for incoming weights to one noun contained in the PIE from the remaining PIE tokens.
Throughout the figures in the paper, we refer to the subset of sentences that have a figurative PIE and a paraphrased translation as `\textit{fig-par}'. The subset of sentences with a literal PIE and a word-for-word translation are indicated by `\textit{lit-wfw}'. We compare those two subsets, as well as all instances of figurative PIEs (`\textit{fig}') to all instances of literal PIEs (`\textit{lit}') using the labels from the MAGPIE dataset.
Overall, there is increased attention in figurative occurrences of PIEs compared to literal instances. This difference is amplified for the subset of figurative PIEs yielding paraphrased translations. This pattern is consistent for all languages, as is displayed in Figure~\ref{fig:att_langs} that presents the difference between the mean attention weights of the figurative, paraphrased instances, and the mean weights of the literal instances translated word for word.\footnote{Appendix~\ref{ap:languages_per_layer} details results per language per layer.}
In other words, figurative PIEs are grouped more strongly than their literal counterparts.

\paragraph{Attention between PIEs and context}
To examine the interaction between a PIE and its context, we obtain the attention weights from tokens within the PIE to nouns in the surrounding context of size 10 (Figure~\ref{fig:att_idi2con}).\footnote{Throughout the paper, a context size of 10 to the left and 10 to the right or smaller is used, as sentence length permits.}
Similarly, the attention from the surrounding context to PIE nouns is measured (Figure~\ref{fig:att_con2idi}).
There is reduced attention from PIEs to context for figurative instances, which mirrors the effect observed in Figure~\ref{fig:att_idi2idi}: increased attention within the PIE is accompanied by reduced attention to the context.
This pattern is consistent across languages (Figure~\ref{fig:att_langs}).
From the context to the PIE, the average weight is slightly higher for literal PIEs, but the effect size is small, indicating only a minor impact of figurativeness on the context's attention weights.
This will be further investigated in \S\ref{sec:hidden_states}.

\paragraph{Cross-attention}
To analyse the encoder-decoder interaction, we decode translations with beam size five, and extract the cross-attention weights for those translations.
Afterwards, alignments are computed for the models' predictions by, together with 1M sentences from the OPUS corpus per target language, aligning them using the \texttt{eflomal} toolkit \citep{ostling2016efficient}
The alignment is used to measure attention from a token aligned to a PIE's noun to that noun on the source side.\footnote{Automated alignments may be less accurate for paraphrases, and, therefore, we inspect the \textit{fig-par} alignments: for all languages $\leq$ 34\% of those sentences has no aligned word for the PIE noun. Those sentences are excluded.
We manually inspect the most frequently aligned words for Dutch, that cover 48\% of the \textit{fig-par} subcategory in Ap.~\ref{ap:alignments}, and are all accurate.}

Figure~\ref{fig:cross_att_idi2idi} presents the attention distribution for the weights that go from the noun's translation to that PIE noun on the source side, for the \texttt{En-Nl} model.
There is a stark difference between figurative and literal PIEs, through reduced attention on the source-side noun for figurative PIEs. This difference is particularly strong for the figurative sentences that are paraphrased during the translation: when paraphrasing the model appears to rely less on the source-side noun than when translating word for word.
Where does the attention flow, instead? To some extent, to the remaining PIE tokens (Figure~\ref{fig:cross_att_idi2con}).
A more pronounced pattern of increased attention on the \texttt{</s>} token is shown in Figure~\ref{fig:cross_att_idi2eos}. Similar behaviour has been observed by \citet{clark2019does} for BERT's \texttt{[SEP]} token, who suggest that this indicates a \textit{no-operation}. In Transformer's cross-attention mechanism, this would mean that the decoder collects little information from the source side.
Figure~\ref{fig:cross_att_langs} compares the mean attention weights of the seven languages for the figurative inputs that are paraphrased to the literal samples that are translated word for word, confirming that these patterns are not specific to \texttt{En-Nl} translation.

\vspace{0.1cm}
\noindent Collectively, the results provide the observations depicted in Figure~\ref{fig:attention_flow}.
When paraphrasing a figurative PIE, the model groups idioms' parts more strongly than it would otherwise -- i.e. it captures the PIE more as one unit. A lack of grouping all figurative PIEs could be a cause of too compositional translations. Increased attention within the PIE is accompanied by reduced interaction with context, indicating that the PIE is translated in a stand-alone manner, contrary to what is expected, namely that contextualisation can resolve the figurative versus literal ambiguity. There is less cross-attention on the source-side PIE and more attention on the \texttt{</s>} token when the model emits the translation of figurative (paraphrased) PIEs. This suggests that even though the encoder cues figurative usage, the decoder retrieves a PIE's paraphrase and generates its translation more as a language model would.

%% file: hidden_states.tex
\section{Hidden representations}
\label{sec:hidden_states}

Within Transformer, the encoder's upper layers have previously been found to encode semantic information \citep[e.g.][]{raganato2018analysis}. PIEs' hidden states are expected to transform over layers due to contextualisation, and become increasingly more indicative of figurativeness.
This section focuses on the impact of PIEs on the hidden states of Transformer's encoder.
We firstly discuss how much these hidden states change between layers.
Secondly, we measure the influence of a token by masking it out in the attention and analysing the degree of change in the hidden representations of its neighbouring tokens. This analysis is performed to consolidate findings from \S\ref{sec:attention}, since the extent to which attention can explain model behaviour is a topic of debate \citep{jain2019attention,wiegreffe2019attention}. 

\subsection{PIE changes over layers}
To compare representations from different layers, we apply \textit{canonical correlation analysis} (CCA) \citep{hotelling1936relations}, using an implementation from \citet{raghu2017svcca}. Assume matrices $A \in \mathcal{R}^{d_A\times N}$ and $B \in \mathcal{R}^{d_B\times N}$, that are representations for $N$ data points, drawn from two different sources with dimensionalities $d_A$ and $d_B$ -- e.g. different layers of one network.
CCA linearly transforms these subspaces $A'=WA$, $B'=VB$ such as to maximise the correlations $\{\rho_1,\dots,\rho_{\text{min}(d_A,d_B)}\}$ of the transformed subspaces.
We perform CCA using ${>}60$k token vectors for a previously unused subset of the MAGPIE corpus -- the subset of sentences that did not contain nouns in the PIEs -- to compute the CCA projection matrices $W$ and $V$. $W$ and $V$ are then used to project new data points before measuring the data points' correlation. The CCA similarity reported in the graphs is the average correlation of projected data points. We do not perform CCA separately per data subset due to the small subset sizes and the impact of vocabulary sizes on CCA correlations for small datasets (see Appendix~\ref{ap:cca}).\footnote{Extensions of CCA have been proposed that limit the number of CCA directions over which the correlation is computed, to only include directions that explain a large portion of the variance \citep{raghu2017svcca, morcos2018insights}. We do not remove directions such as to avoid removing smaller variance components that could still cue figurativeness (the focus of our work).}

\begin{figure}
\begin{subfigure}[b]{0.44\columnwidth}
    \centering
    \includegraphics[width=0.9\textwidth]{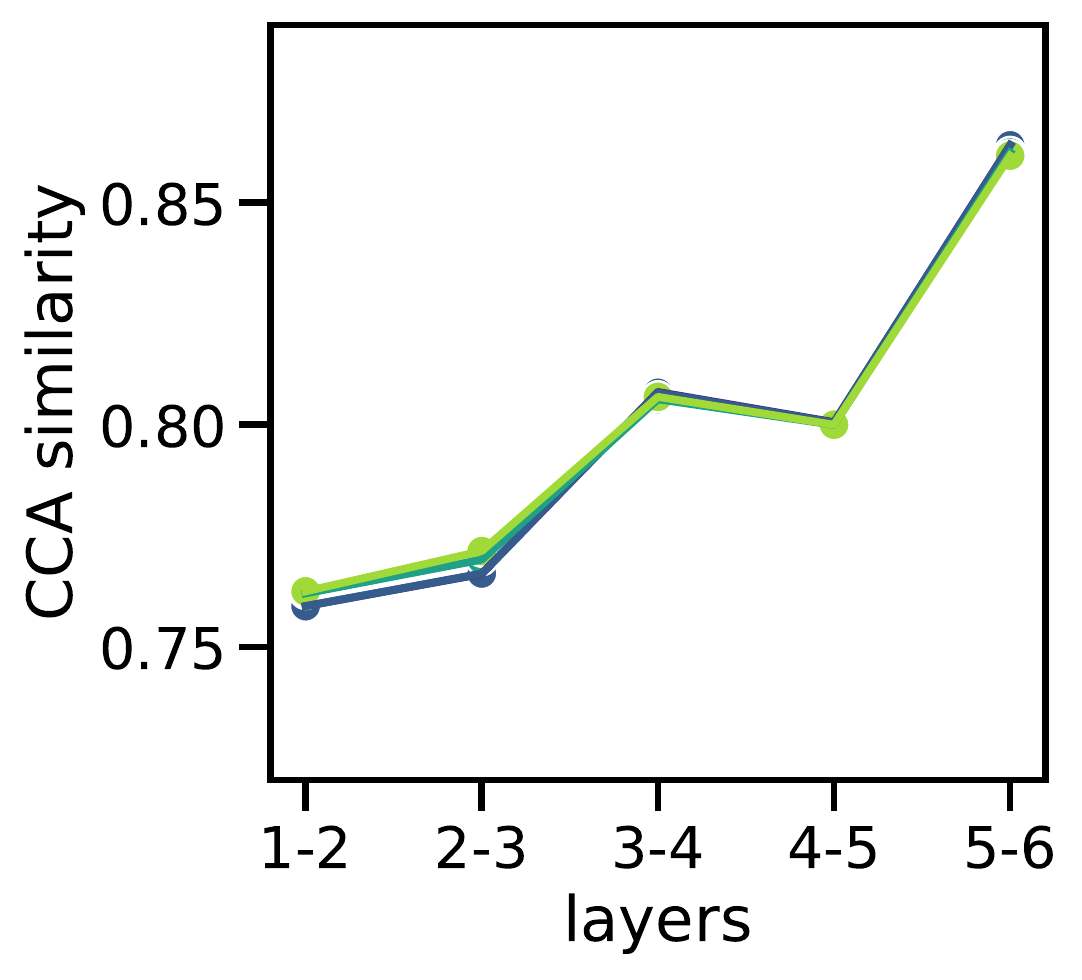}
    \caption{Non-PIE nouns}
    \label{fig:cca_layers_non_pie}
\end{subfigure}
\begin{subfigure}[b]{0.54\columnwidth}
    \centering
    \includegraphics[width=0.9\textwidth]{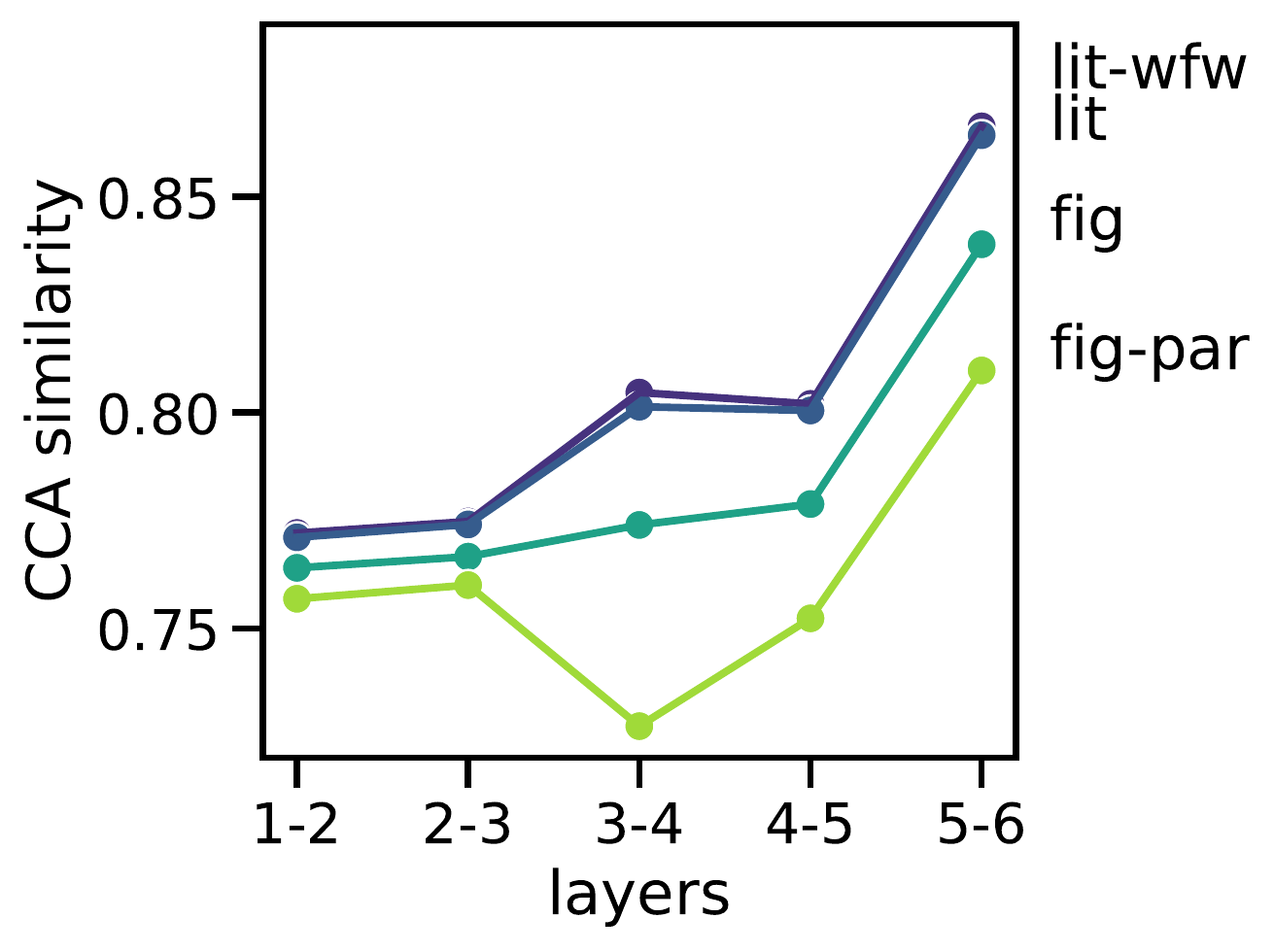}
    \caption{PIE nouns}
    \label{fig:cca_layers_pie}
\end{subfigure}
\begin{subfigure}[b]{\columnwidth}
    \centering
    \includegraphics[width=0.85\textwidth]{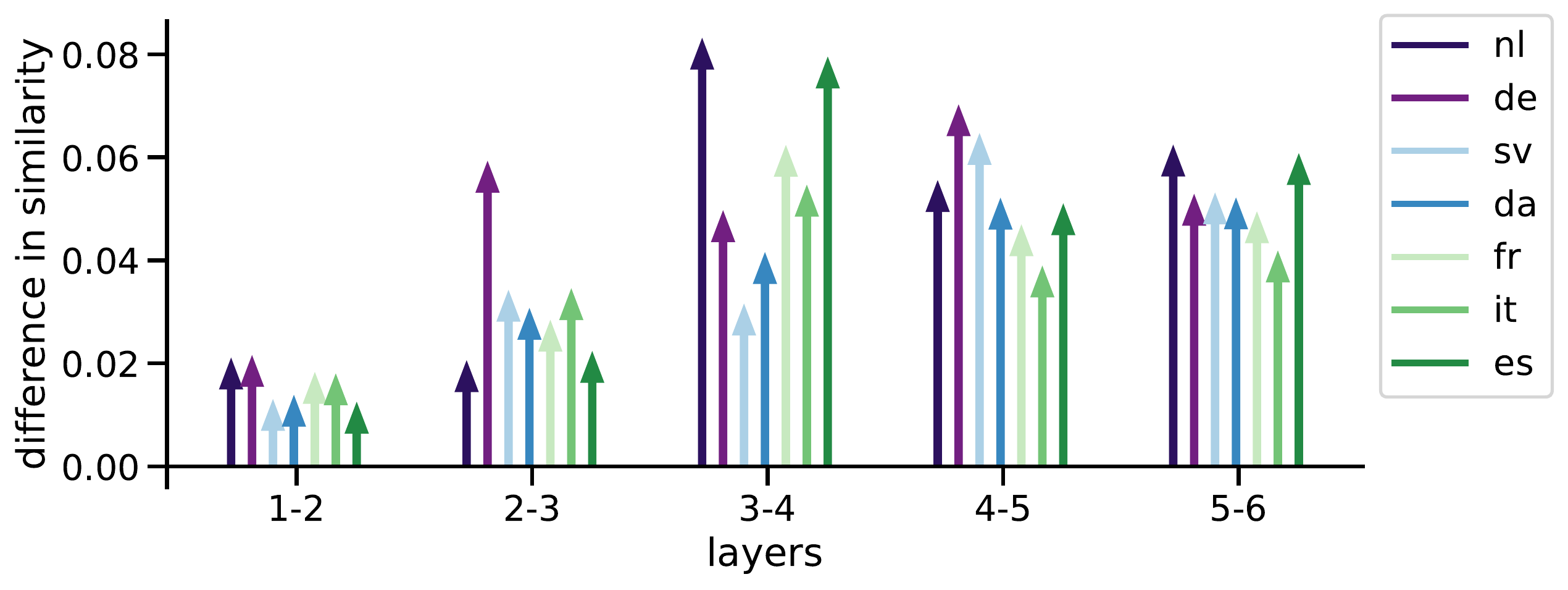}
    \caption{Languages comparison for PIE nouns}
    \label{fig:cca_layers_languages}
\end{subfigure}
\caption{CCA similarity for layer $l$ and layer $l+1$, for PIE and non-PIE nouns. The languages comparison displays the difference in similarity between \textit{lit-wfw} and \textit{fig-par}.}
\label{fig:cca_layers}
\end{figure}

We compute the CCA similarity for hidden states from adjacent layers for PIE and non-PIE nouns.
Figurative PIEs in layer $l$ are typically less similar to their representation in layer $l-1$ compared to literal instances (shown in Figures~\ref{fig:cca_layers_pie} and~\ref{fig:cca_layers_languages}).
The results for non-PIE nouns (Figure~\ref{fig:cca_layers_non_pie} for the \texttt{En-Nl} Transformer) do not differ across data subsets, suggesting that changes observed for figurative PIEs are indeed due to figurativeness.

\begin{figure}[!t]
\begin{subfigure}[b]{0.47\columnwidth}
    \centering
    \includegraphics[width=0.89\textwidth]{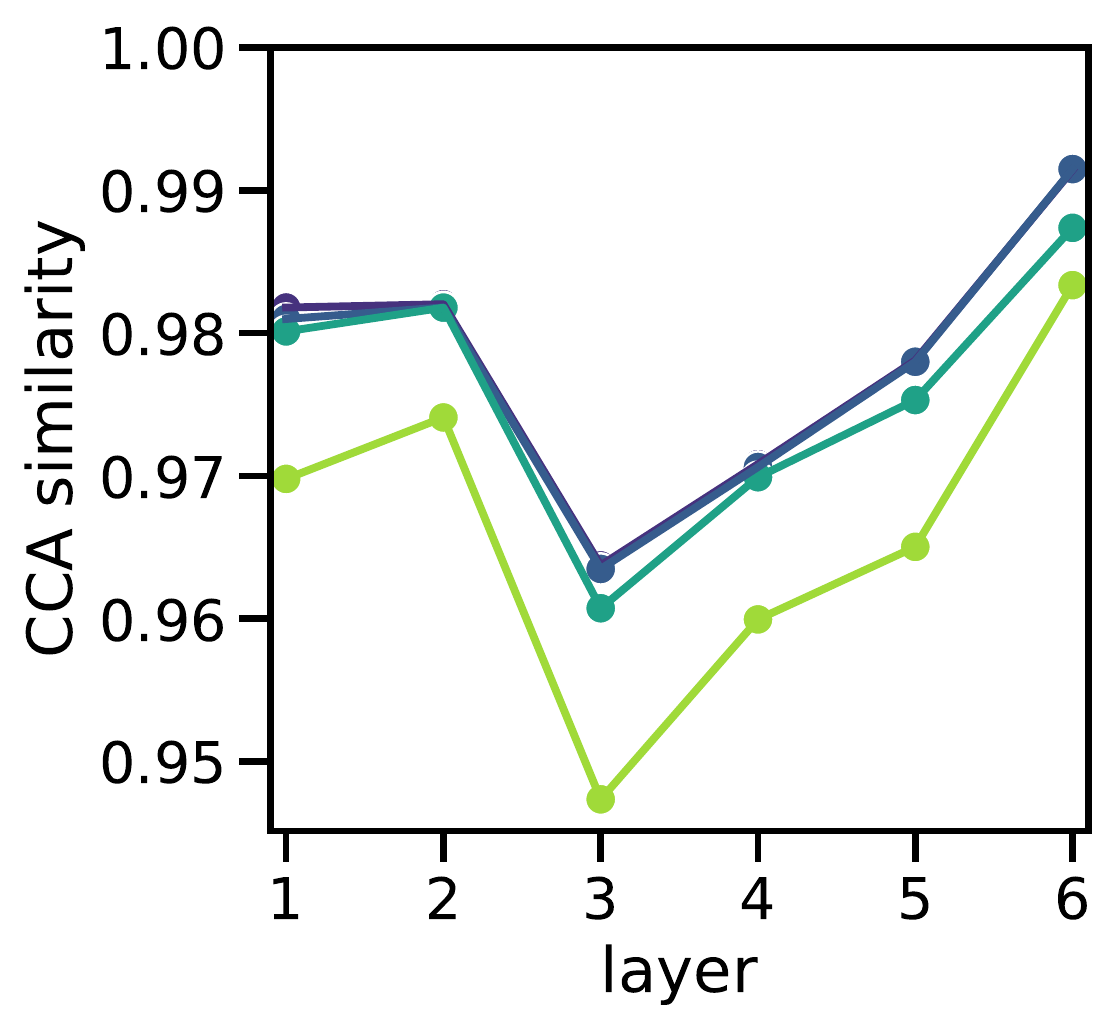}
    \caption{Influence PIE on PIE}
    \label{fig:cca_idiom_on_idiom}
\end{subfigure}
\begin{subfigure}[b]{0.47\columnwidth}
    \centering
    \includegraphics[width=0.89\textwidth]{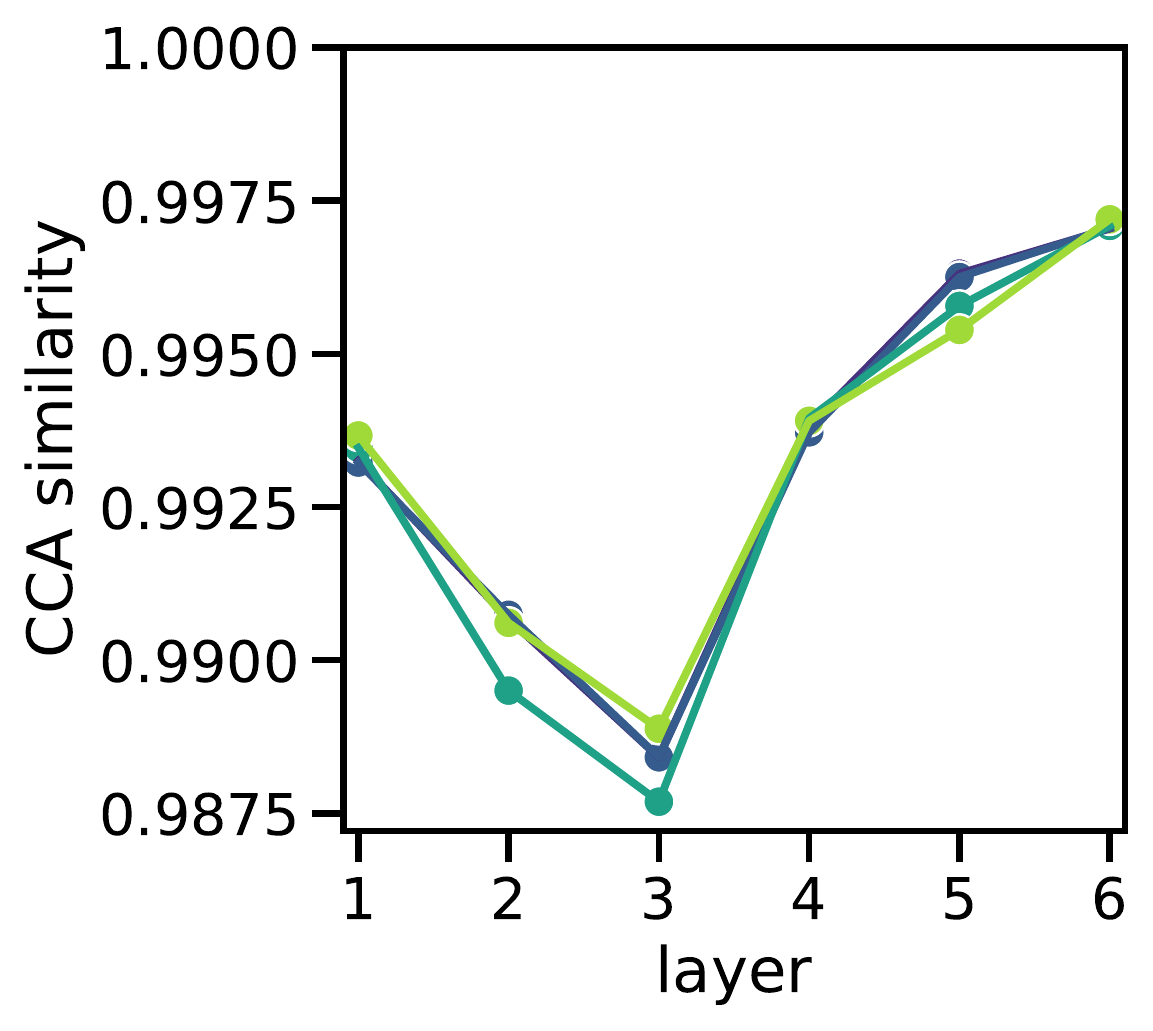}
    \caption{Influence PIE on context}
    \label{fig:cca_idiom_on_context}
\end{subfigure}
\begin{subfigure}[b]{0.47\columnwidth}
    \centering
    \includegraphics[width=0.89\textwidth]{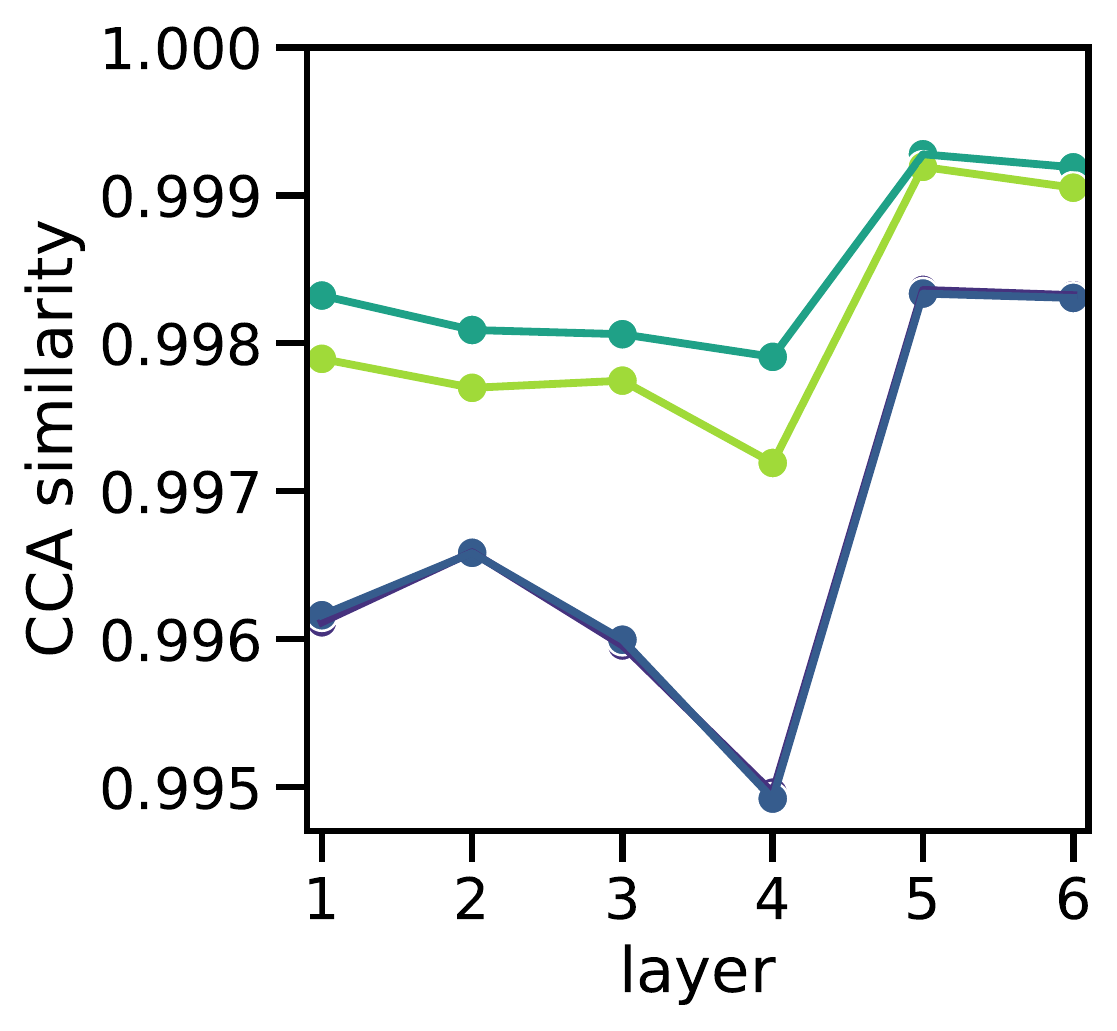}
    \caption{Influence context on PIE}
    \label{fig:cca_context_on_idiom}
\end{subfigure}\hspace{0.3cm}\begin{subfigure}[b]{0.47\columnwidth}
    \centering
    \includegraphics[width=0.89\textwidth]{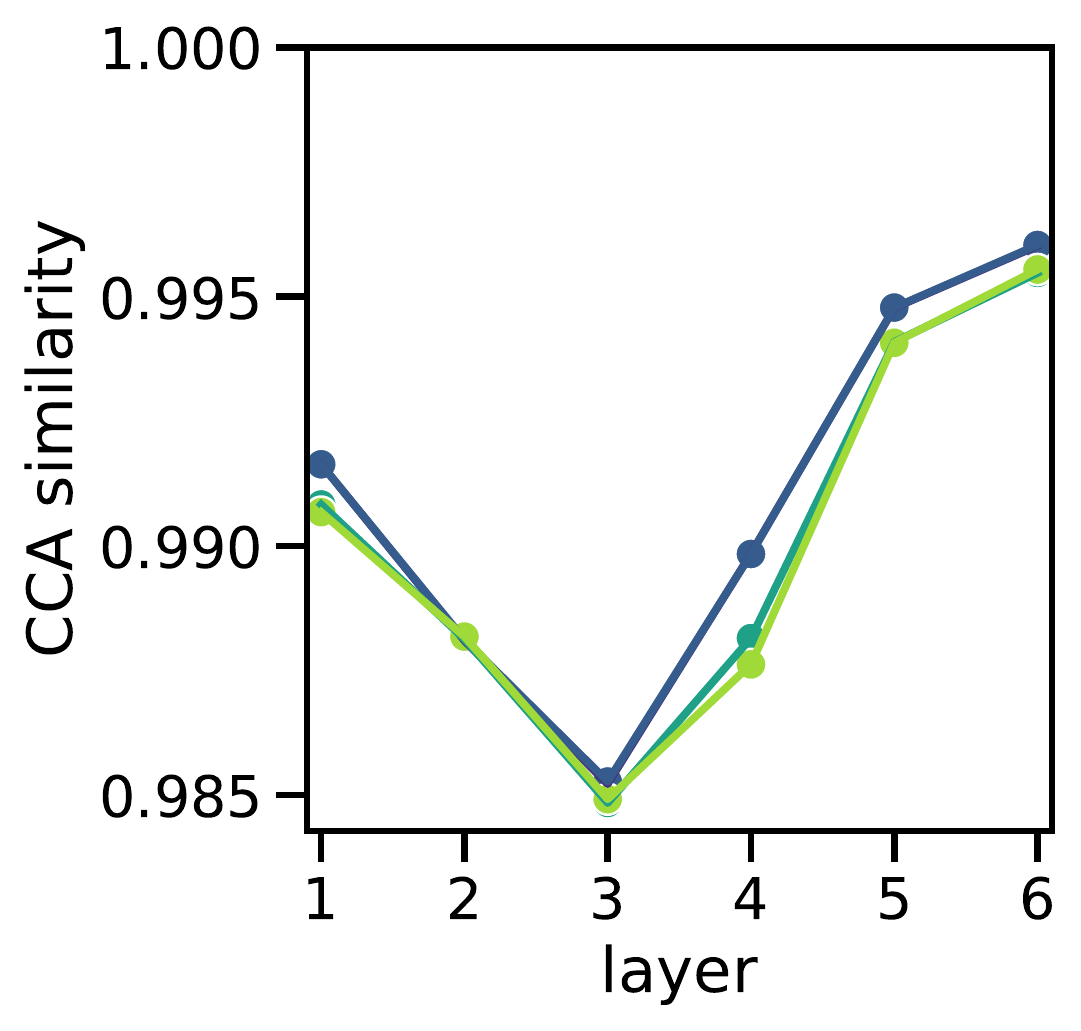}
    \caption{Control setup}
    \label{fig:cca_control}
\end{subfigure}
\begin{subfigure}[b]{\columnwidth}
    \centering
    \includegraphics[width=0.89\textwidth]{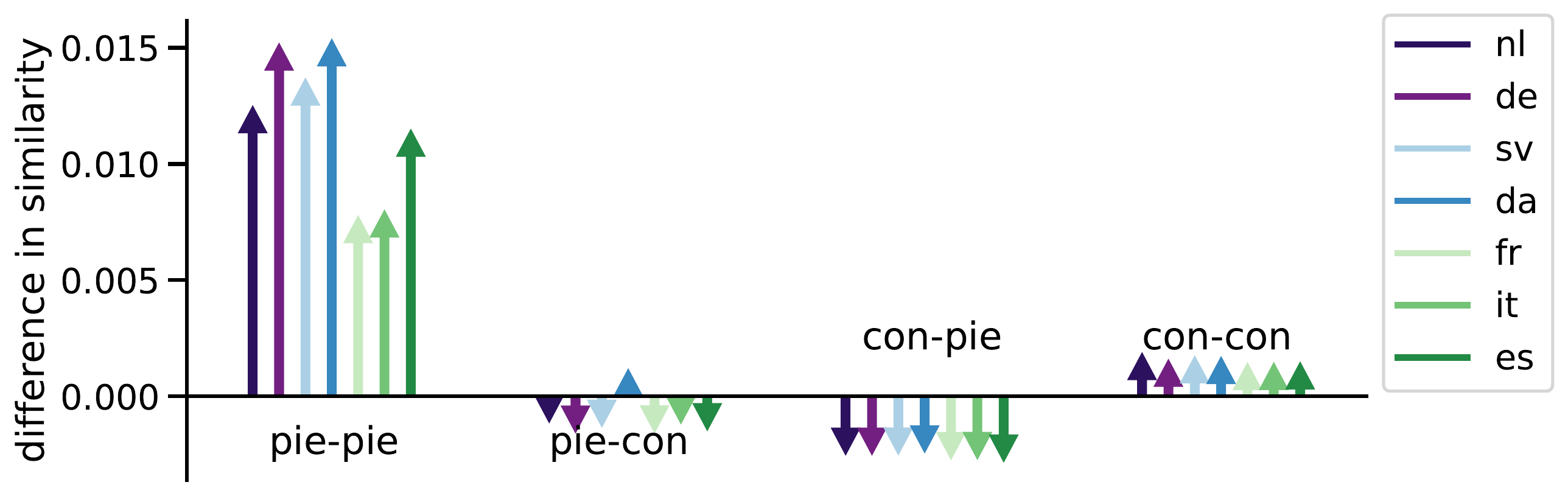}
    \caption{Languages comparison}
    \label{fig:cca_languages}
\end{subfigure}
\caption{Impact of masking a PIE noun in the attention on (a) other PIE tokens, (b) other context tokens.
Impact of masking a non-PIE noun on (c) PIE tokens and (d) other non-PIE tokens.
(e) shows the difference in similarity between \textit{lit-wfw} and \textit{fig-par}.}
\label{fig:cca_influence1}
\end{figure}

\subsection{Intercepting in attention}
We now compute similarities of representations for the model in two setups: with and without one token masked in the attention mechanism, as suggested by \citet{voita2019bottom}.
Masking a token means that other tokens are forbidden to attend to the chosen one.
This can reveal whether the attention patterns discussed in \S\ref{sec:attention} are indicative of the influence tokens have on each other's hidden representations. 
The first representation is the hidden representation from layer $l$ for a token encoded as usual. The second one is the hidden representation of layer $l$ when applying the first $l-1$ layers as usual and masking one token in the $l$th layer. CCA is again performed on separate data, where a non-PIE noun is masked, to provide the projection matrices applied before computing similarities in the remainder of this subsection.

\paragraph{Masking a PIE token}
To estimate the influence of PIE nouns, we first compute the CCA similarity between two representations of tokens from the PIE's context while masking one PIE noun in the attention for one of those representations.
Similarly, we measure the influence on other tokens within the PIE when masking one PIE noun.
Within the PIE, the impact is the largest for figurative instances (see Figure~\ref{fig:cca_idiom_on_idiom} for \texttt{En-Nl} and \ref{fig:cca_languages} for averages over layers for all languages).
This is in line with the attention pattern observed.
However, whether the impact is the largest on context tokens from figurative or literal instances is dependent on the layer (Figure~\ref{fig:cca_idiom_on_context}), suggesting that the slight difference in attention from the context to the PIE observed in \S\ref{sec:attention} need not represent a difference in impact between figurative and literal PIEs.

\paragraph{Masking a context token}
Lastly, we measure the influence of masking a noun in the context of the PIE on PIE tokens and non-PIE tokens. 
Within the PIE, as shown in Figures~\ref{fig:cca_context_on_idiom} and~\ref{fig:cca_languages}, figurative instances are less affected by the masked context noun compared to literal occurrences of PIEs.
Again, this mirrors the patterns observed for attention where there was less attention on the context for figurative PIEs.
When masking a non-PIE noun and measuring the impact on non-PIE tokens, one would hardly expect any differences between data subsets, as is confirmed in Figures~\ref{fig:cca_control} and ~\ref{fig:cca_languages}.

\vspace{0.1cm}
\noindent In summary, these analyses confirm most of the trends noted for attention patterns. Intercepting in the attention through masking indicated that for PIE tokens, there is less interaction with the context.
However, this does not necessarily mean that the context interacts less with figurative PIEs compared to literal PIEs, even if there was a slight difference in attention (see \S\ref{sec:attention}).
The CCA analyses furthermore showed that figurative PIEs are distinct from typical tokens in how they change over layers.

%% file: amnesic_probing.tex
\begin{figure}\centering
\includegraphics[width=\columnwidth]{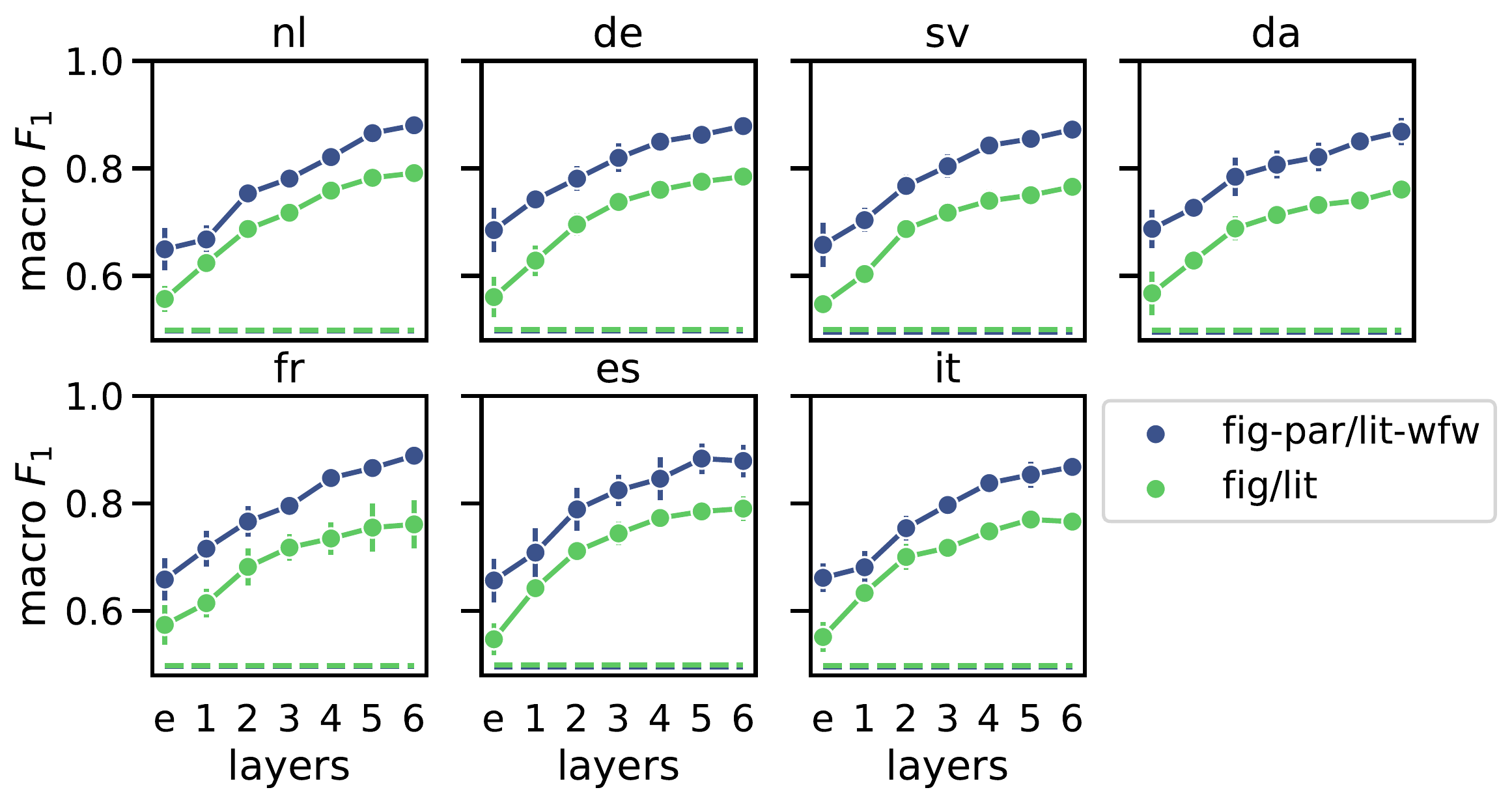}
\caption{Macro $F_1$-score for probes predicting PIEs' labels. Error bars show standard deviations over folds.}
\label{fig:probing}
\end{figure}

\begin{table}[t]
\small\setlength{\tabcolsep}{5pt}
    \resizebox{0.35\columnwidth}{!}{
    \begin{subtable}[b]{.4\columnwidth}\centering\small
    \begin{tabular}{lccccccc}
    \toprule
     & \textbf{Fig.} & \textbf{Freq.} \\ \midrule
    \texttt{nl}  &  36/75 & 34/75 \\
    \texttt{de}  &  33/68 & 33/69 \\
    \texttt{sv}  &  27/77 & 27/77 \\
    \texttt{da}  &  32/77 & 27/78 \\
    \texttt{fr}  &  37/77 & 30/76 \\
    \texttt{it}  &  39/76 & 34/77 \\
    \texttt{es}  &  40/78 & 30/78 \\
    \bottomrule
    \end{tabular}
    \caption{Success rate / BLEU}
    \end{subtable}}\begin{subtable}[b]{.6\columnwidth}
        \includegraphics[width=\textwidth]{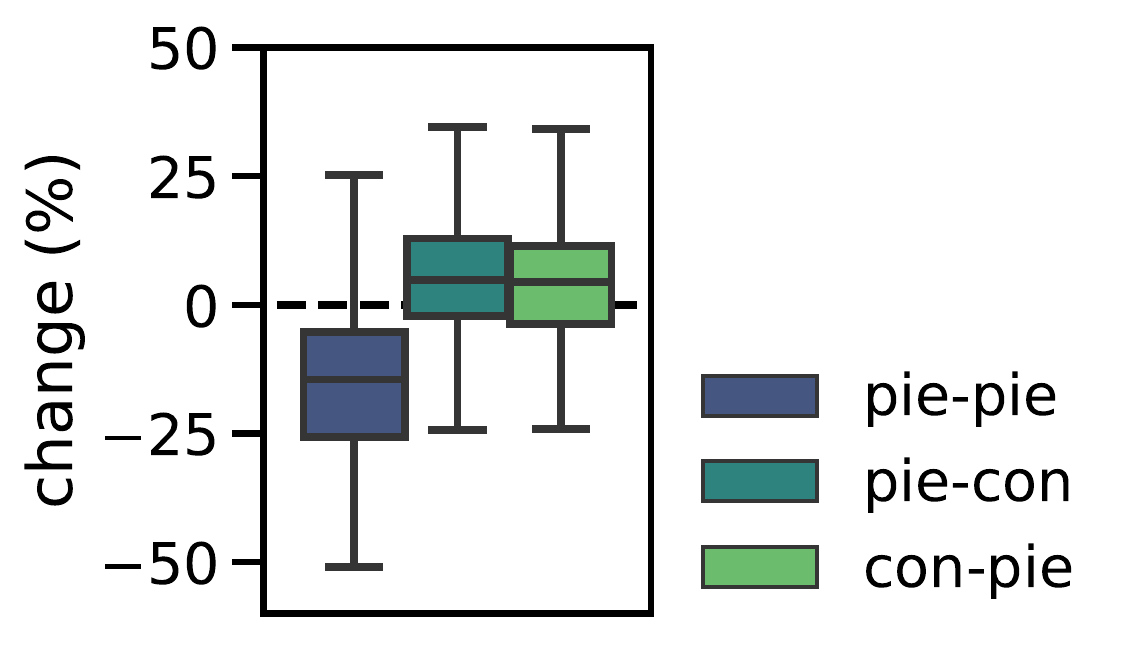}
        \caption{Change in attention (\texttt{nl})}
        \label{fig:inlp_attention}
    \end{subtable}
\caption{Impact of amnesic probing as (a) the success rate per PIE type (\%), and the BLEU score of translations that changed from a paraphrase to a word-for-word translation, and (b) the changes in attention.}
\label{tab:amnesic_probing}
\end{table}

\section{(Amnesic) probing for figurativeness}
\label{sec:amnesic_probing}

The previous analyses compared the hidden states for figurative and literal PIEs, but do not use these labels, otherwise. We now train logistic regression \textit{probing classifiers} \citep{conneau-etal-2018-cram} to predict the label from hidden representations.
The probes' inputs are the hidden states of PIE tokens, and the $F_1$-scores are averaged over five folds.
All samples from one PIE are in the same fold, such that the classifier is evaluated on PIEs that were absent from its training data.
The results (Figure~\ref{fig:probing}) indicate figurativeness can be predicted from these encodings, with
performance increasing until the top layer for all languages. $F_1$-scores for the embeddings already exceed a random baseline, indicating some idioms are recognisable independent of context.

Finally, we use probing classifiers to change models' PIE translations through 
\textit{amnesic probing} \citep{elazar2021amnesic}: removing features from hidden states with \textit{iterative null-space projection} (INLP) \citep{ravfogel-etal-2020-null} and measuring the influence of these interventions.
INLP trains $k$ classifiers to predict a property from vectors. After training probe $i$, parametrised by $W_i$, the vectors are projected onto the nullspace of $W_i$.
The projection matrix of the intersection of all $k$ null spaces can then remove features found by these classifiers.
Using INLP, we train $50$ classifiers to distinguish figurative PIEs that will be paraphrased from those to be translated word for word.
Afterwards, we run the previously paraphrased PIE occurrences through the model while removing information from the PIE's hidden states using INLP -- i.e. information that could be captured by linear classifiers, which need not be the only features relevant to idiomatic translations.
Per idiom, we record the percentage of translations that are no longer paraphrased.
We report the scores for idioms from four folds and BLEU scores comparing translations that changed label before and after INLP. 
A fifth fold is used for parameter estimation (Appendix~\ref{ap:amnesic_probing}).

Table~\ref{tab:amnesic_probing} presents the results.
When intervening in the hidden states for all layers $l\in\{0,1,2,3,4\}$, the average success rate per PIE ranges from 27\% (for Swedish) to 40\% (for Spanish). 
The interventions yield reduced attention within the PIE and increased interaction with the context (see Table~\ref{fig:inlp_attention} for Dutch).
Table~\ref{tab:amnesic_probing} also provides results for a baseline probe predicting whether the half-harmonic mean of the zipf-frequency of PIE tokens is below or above average. This probe is successful too, emphasising how brittle idiomatic translations are: when removing information from the hidden states, the model reverts to compositional translations.

\begin{figure}
    \centering
    \includegraphics[width=\columnwidth]{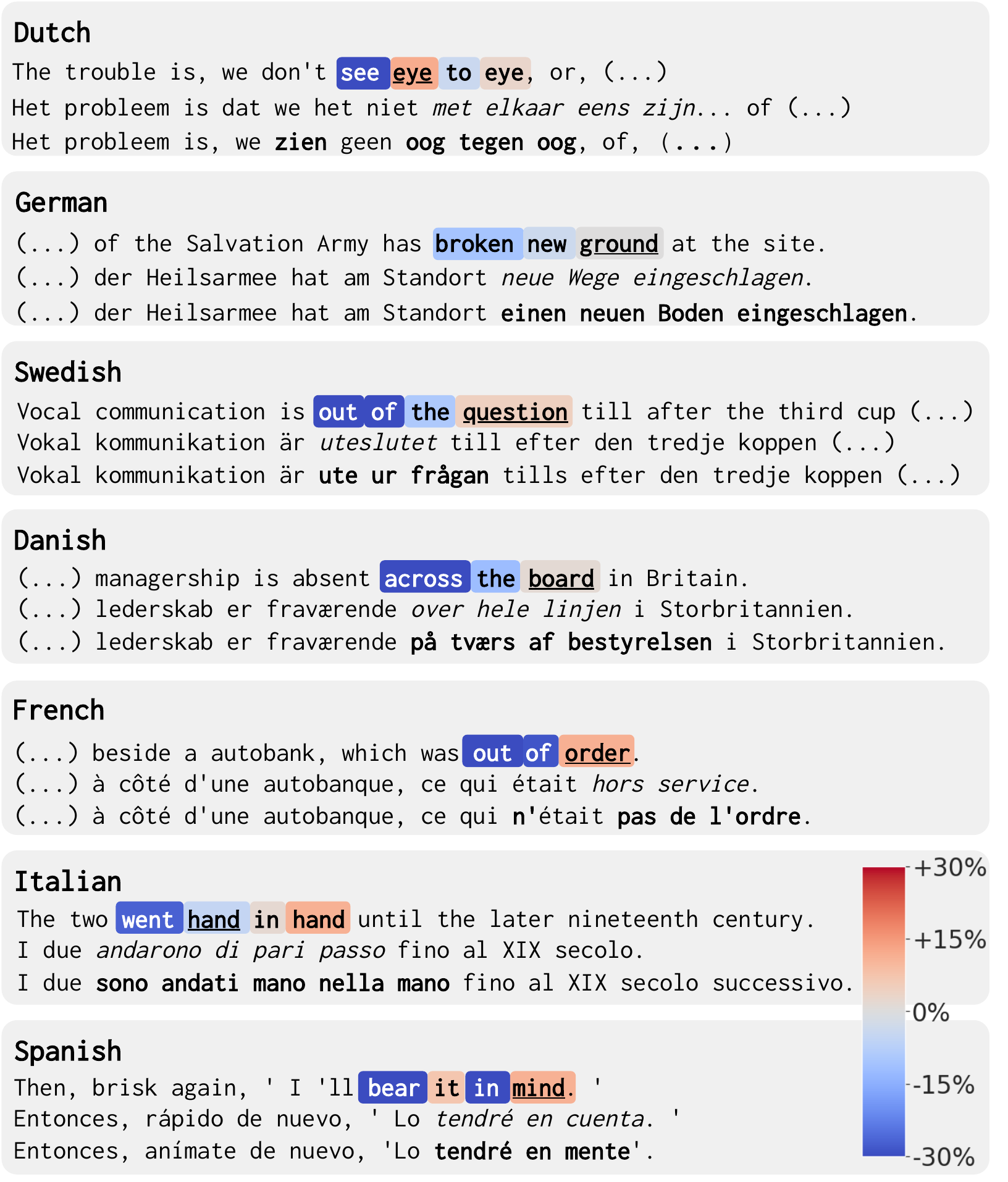}
    \caption{Source sentences and translations before and after INLP. PIEs and word-for-word translations are in bold font; paraphrases in italics. Colours indicate attention changes with respect to the underlined nouns.}
    \label{fig:amnesic_examples}
\end{figure}

Figure~\ref{fig:amnesic_examples} provides example translations before and after the application of INLP, while indicating how the attention on the underlined noun changes.
Generally, the attention on that noun reduces for tokens other than itself.

\vspace{0.2cm}
\noindent In summary, when applying INLP to hidden states, the attention patterns resemble patterns for literal tokens more, confirming a causal connection between the model paraphrasing figurative PIEs and the attention. However, amnesic probing cannot change the paraphrases for all idioms; thus, figurativeness is not merely linearly encoded in the hidden states.
The probing accuracies differed across layers and suggested figurativeness is more easily detectable in higher layers, which is in line with the changes across layers observed in \S\ref{sec:hidden_states}.

%% file: conclusion.tex
\section{Conclusion}
\label{sec:conclusion}

Idioms are challenging for NMT models that often generate overly compositional idiom translations. To understand why this occurs, we analysed idiom processing in Transformer, using an English idiom corpus and heuristically labelled translations in seven target languages.
We compared hidden states and attention patterns for figurative and literal PIEs. In the encoder, figurative PIEs are grouped more strongly as one lexical unit than literal instances and interact less with their context.
The effect is stronger for paraphrased translations, suggesting that capturing idioms as single units and translating them in a stand-alone manner aids idiom processing.
This finding agrees with results from \citet{zaninello2020multiword}, who ascertain that encoding an idiom as one word improves translations. It also agrees with the INLP application causing more compositional translations whilst changing the attention. By relying less on the encoder's output, the decoder determines the meaning of figurative PIEs more independently than for literal ones. To improve idiomatic translations, future work could use these insights to make architectural changes to improve the grouping of idioms as single units by training specific attention heads to capture multi-word expressions or by penalising overly compositional translations in the training objective.

Although we learnt about mechanics involved in idiomatic translations, the vast majority of translations was still word for word, indicating that non-compositional processing does not emerge well (enough) in Transformer.
Paradoxically, a recent trend is to encourage \textit{more} compositional processing in NMT \citep[][i.a.]{chaabouni2021can, li-etal-2021-compositional, raunak2019compositionality}. 
We recommend caution since this inductive bias may harm idiom translations.
It may be beneficial to evaluate the effect of compositionality-favouring techniques on non-compositional phenomena to ensure their effect is not detrimental.

%% file: acknowledgements.tex
\section*{Acknowledgements}

We are grateful to Rico Sennrich for providing feedback on an earlier version of the paper.
Many thanks to Agostina Calabrese, Matthias Lindemann, Gautier Dagan, Irene Winther, Ronald Cardenas, Helena Fabricius-Vieira and Emelie van de Vreken for data annotation and assistance with queries about their native languages.
VD is supported by the UKRI Centre for Doctoral Training in Natural Language Processing, funded by the UKRI (grant EP/S022481/1) and the University of Edinburgh, School of Informatics and School of Philosophy, Psychology \& Language Sciences. IT acknowledges the support of the European Research Council (ERC StG BroadSem 678254) and the Dutch National Science Foundation (NWO Vidi 639.022.518).

%% file: appendix.tex
\clearpage

\begin{appendices}
\section{Survey details}
\label{ap:survey}

\subsection{Crowd-sourcing annotations for Dutch}

In an early phase of the research, the quality of the heuristic annotation method was estimated through a survey conducted using the Qualtrics platform by annotators from Prolific.
The heuristic annotation method labelled a translation as `word for word' if the literal translation of a keyword was present, where the keyword was elicited from MarianMT, and from the translation tool DeepL.
These annotators were native speakers of Dutch, and fluent in English.
To guard the quality of the data collection, participants went through a pre-screening process that consisted of a shorter version of the survey with three practice questions and seven regular questions. Participants were selected for the full study if they correctly answered practice questions, used all three of the labels (paraphrase, word for word, copy), and did not choose `copy' if the keyword was clearly absent from the translation.
The main survey consisted of three parts: (1) An explanation of what an idiom is, of potential literal and figurative usage of PIEs, the meaning of the three labels, and the format to be used in the study. (2) One practice exercise where three potential translations of one sentence had to be connected to the correct label. (3) Lastly, 38 questions were filled out: 12 instances that were figurative and were paraphrased by the model, 4 literal instances paraphrased by the model, 8 literal instances that were translated word for word, 8 figurative instances that were translated word for word, 6 copies (3 figurative, 3 literal).

If the participant indicated that it was a word-by-word translation, the follow-up question would be asked, where the participant indicated the literal translation of the keyword. We repeated the instruction of what constitutes a word-by-word translation since participants would often select the (conventionalised) idiomatic translation in the pre-screening phase -- e.g. `handbereik' for `fingertips', for which a literal translation would be `vingertoppen'.

\begin{table}[!t]
    \centering\small\setlength{\tabcolsep}{4pt}
    \begin{tabular}{l|ccccccccc}
    \toprule
    \textbf{MAGPIE} & \multicolumn{9}{c}{\textbf{Predicted Translations}} \\
    & \multicolumn{3}{c}{\textit{Paraphrase}} & \multicolumn{3}{c}{\textit{Word for word*}} & \multicolumn{3}{c}{\textit{Copy*}} \\
    & \# & \% & agr & \# & \% & agr & \# & \% & agr \\ \midrule
    Figurative  & 96 & 86 & 84 & 64 & 84 & 77 & 24 & 83 & 58 \\
    Literal     & 32 & 73 & 59 & 64 & 91 & 80 & 24 & 69 & 88 \\
    \bottomrule
    \end{tabular}
    \caption{Survey statistics: the number of sentence pairs used (\#), the \% of labels the algorithm and annotators agreed on, and inter-annotator agreement. Agreement means an average of 4 annotators agreed on the label unanimously.
    *Categories merged in the main paper.}
    \label{tab:study_stats_old}
    \vspace{-0.3cm}
\end{table}

Table~\ref{tab:study_stats_old} summarises the survey outcomes. The annotators and the heuristic method agreed in 83\% of the cases.
For 77\% of the samples, the annotations agreed on the label unanimously.

\subsection{Collecting annotations for 7 languages}

Later on, the analyses were applied to heuristically annotated data for all seven languages.
The procedure to elicit the translations of keywords from MarianMT and an online translation tool were adapted to improve the recall of keywords for languages other than Dutch.
Afterwards, postgraduate students from the local university were invited to annotate the data in exchange for payment, where one annotator annotated all 350 samples for a language.
To reduce the cognitive load of the experiment, only sentences with $\leq$ 40 tokens were shown to the participants.
The annotators were native in the target language and fluent in English, with the exception of the Swedish speaker, that was native in Norwegian and Finnish, and fluent in Swedish and English.
The annotators participated in a similar pre-screening test with language-specific explanations and examples, and seven practice questions.
If the annotators' answers differed from what was expected, the instructions were discussed with the annotator before they proceeded with the full survey, and they filled out the remainder of the survey without intermediate help or instructions.
Table~\ref{tab:study_question_format} shows an example question for Dutch.

\begin{table}[!t]
\centering\small
\resizebox{\columnwidth}{!}{
\begin{tabular}{l}
\toprule
\textbf{Question} \\
The following sentence contains "at your \textcolor{red}{\underline{fingertips}}": \\
"Using the latest in audio visual technology, the wonders of \\ 
these six fascinating ‘worlds’ are at your fingertips." \\ \\
Now categorise the translation of the red word from \\
above in this sentence: \\
"Met behulp van de nieuwste audio visuele technologie, \\
zijn de wonderen van deze zes fascinerende \\
werelden binnen handbereik." \\
$\circ$ paraphrase \\
$\circ$ word-by-word \\
$\circ$ copy \\ \midrule
\textbf{Follow-up question} \\
\textit{If you did not select 'word for word', leave blank.} \\
What is the translation of the red keyword in
"at your \textcolor{red}{\underline{fingertips}}" \\
in the sentence below: \\
(\dots insert sentence\dots) \\
(\dots free text response box\dots)\\
\bottomrule
\end{tabular}}
\caption{Format of the questions shown to participants via the Qualtrics platform.}
\label{tab:study_question_format}
\vspace{-0.3cm}
\end{table}

\subsection{Ethical considerations}
\label{sec:ethical_considerations}
The surveys referred to in \S\ref{subsec:data} were both approved through to the university's research ethics process, where an independent committee assessed the setup of the survey, the research's potential harmful impacts and the compensation for the participants.
In collecting data annotations, participants were shown data from the MAGPIE corpus, available under the CC-BY-4.0 License.
All other information shown to them was either collected from the computational model, or written up by the authors.
Any identifiable information about the participants was stored separately from the participants' annotations, for the purposes of compensation.
Participants were able to provide informed consent to data collection and anonymised data being used in academic publications.
They were given the opportunity to withdraw at any time.
Participants were compensated above the minimum hourly wage of the country in which they were a resident at the time of participating in the study.

\vspace{0cm}
\noindent\begin{minipage}[b]{\textwidth}
    \centering\small
    \begin{tabular}{lll}
    \toprule
    \textbf{PIE} & \textbf{Dutch paraphrase (literal backtranslation)} & \textbf{Aligned tokens} \\\midrule\midrule
    across the board    & over hele linie (\textit{over the whole line})   & board $\rightarrow$ linie \\
    behind the scenes   & achter de schermen (\textit{behind the screens}) & scenes $\rightarrow$ schermen \\
    break new ground    & nieuwe weg inslaan (\textit{take a new road})    & ground  $\rightarrow$ weg \\
    by heart            & uit het hoofd (\textit{from the head})           & heart  $\rightarrow$  hoofd \\
    by the same token   & op dezelfde manier (\textit{in the same way})    & token $\rightarrow$ manier \\
    come to mind        & in me opkomen (\textit{come up in me})          & mind $\rightarrow$ me \\
    come of age         & volwassen worden (\textit{become an adult})      & age  $\rightarrow$ volwassen \\
    face to face        & oog in oog (\textit{eye in eye})                 & face  $\rightarrow$  oog \\
    follow suit         & het voorbeeld volgen van (\textit{follow the example of}) & suit  $\rightarrow$  voorbeeld \\
    for good measure    & in goede mate* (\textit{in good measure})        & measure $\rightarrow$ mate \\
    from scratch        & vanaf nul (\textit{from zero})                   & scratch $\rightarrow$ nul \\
    from the word go    & vanaf het begin (\textit{from the start})        & word $\rightarrow$ begin \\
    get a move on       & schiet op (\textit{hurry})                       & move $\rightarrow$ schiet \\
    get the picture     & een completer beeld krijgen (\textit{get a more complete vision}) & picture $\rightarrow$ beeld \\
    get to grips with   & (aan)pakken (\textit{take on})                   & grips $\rightarrow$ pakken \\
    give someone the creeps & kriebels krijgen (\textit{getting tickles})  & creep $\rightarrow$ (krie)bel \\
    in broad daylight   &  op klaarlichte dag (\textit{on a luminous day}) & day(light) $\rightarrow$ dag \\
    in full swing       & in volle gang (\textit{in full progress})        & sw(ing) $\rightarrow$ gang \\
    in the flesh        & in levende lijve (\textit{in the living body})   & flesh $\rightarrow$ lij(ve) \\
    in the long run     & op de lange termijn (\textit{on the long term})  & run $\rightarrow$ termijn \\
    in the short run    & op de korte termijn (\textit{on the short term}) & run $\rightarrow$ termijn \\
    keep a low profile  & zich gedeisd houden (\textit{to lay low})        & profile $\rightarrow$ (gede)is(d) \\
    off the record      & onofficieel (\textit{unofficial})                & record $\rightarrow$ (onoffici)eel \\
    on someone's mind   & iets aan je hoofd hebben (\textit{have something on your head}) & mind $\rightarrow$ hoofd \\
    once in a while     & af en toe (\textit{on and off})                  & while $\rightarrow$ toe \\
    out of the blue     & uit het niets (\textit{out of nothing})          & blue $\rightarrow$ niets \\
    out of the question & uit de boze (\textit{from the bad})              & question $\rightarrow$ boze \\
    set eyes on         & zien / zag (\textit{see / saw})                  & eyes $\rightarrow$ zag \\
    small print         & in de kleine lettertjes (\textit{in the little letters}) & print $\rightarrow$ (letter)tjes \\
    take a back seat    & op de achterbank (\textit{on the back bench})   & seat $\rightarrow$ bank \\
    take stock          & de balans opmaken (\textit{make up the balance}) & stock $\rightarrow$ balans \\
    to all intents and purposes & in alle opzichten (\textit{in all aspects}) & intent $\rightarrow$ opzichten \\
    to boot             & opstarten (\textit{to start})                    & boot $\rightarrow$ (op)starten \\
    to the tune of      & voor het bedrag van (\textit{for the amount of}) & tune $\rightarrow$ bedrag \\
    with a view to      & met het oog op (\textit{with the eye on})        & view $\rightarrow$ oog \\
    \bottomrule
    \end{tabular}
    \captionof{table}{PIEs for which the word most commonly aligned to the keyword occurs $>20$ times. Together, these keywords determine 48\% of all the alignments used to perform the cross-attention analysis for \textit{fig-par} in the English-Dutch model. Subwords shown in brackets are due to the subtokens used in Marian-MT: \texttt{eflomal} aligns the parts outside of the brackets to one another.\\
    $^{\star}$Example of a PIE for which the heuristic annotation missed out on a potential literal translation of `measure'.}
    \label{tab:eflomal}
\end{minipage}
\section{Aligning PIEs and paraphrases}
\label{ap:alignments}
When automatically aligning sentences with PIEs to translations that are labelled as a paraphrase by the heuristic, how does the automated aligner (the \texttt{eflomal} toolkit of \citeauthor{ostling2016efficient}, \citeyear{ostling2016efficient}) handle paraphrases?
For many PIEs ($\leq$ 34\% of the \textit{fig-par} sentences for all languages), the paraphrases do not have a word in the translation aligned to the PIE keyword on the source side using \texttt{eflomal}. These examples are excluded. However, for a subset that appears more well-known, there are common paraphrases that the PIE keyword aligns to. We provide examples for Dutch in Table~\ref{tab:eflomal}. The examples provided in the table together cover 48\% of all aligned sentences used in the cross-attention analysis for the \textit{fig-par} category, and all are reasonable alignments.

\clearpage
\section{Attention for data subsets}
\label{ap:attention}

The attention weight distributions in the main paper included all data. To further investigate whether the differences in attention patterns observed are due to factors other than figurativeness, we recompute the attention patterns for three additional data subsets.

\begin{figure}[!b]
    \centering
    \begin{subfigure}[b]{0.45\columnwidth}\centering
    \includegraphics[width=0.99\textwidth]{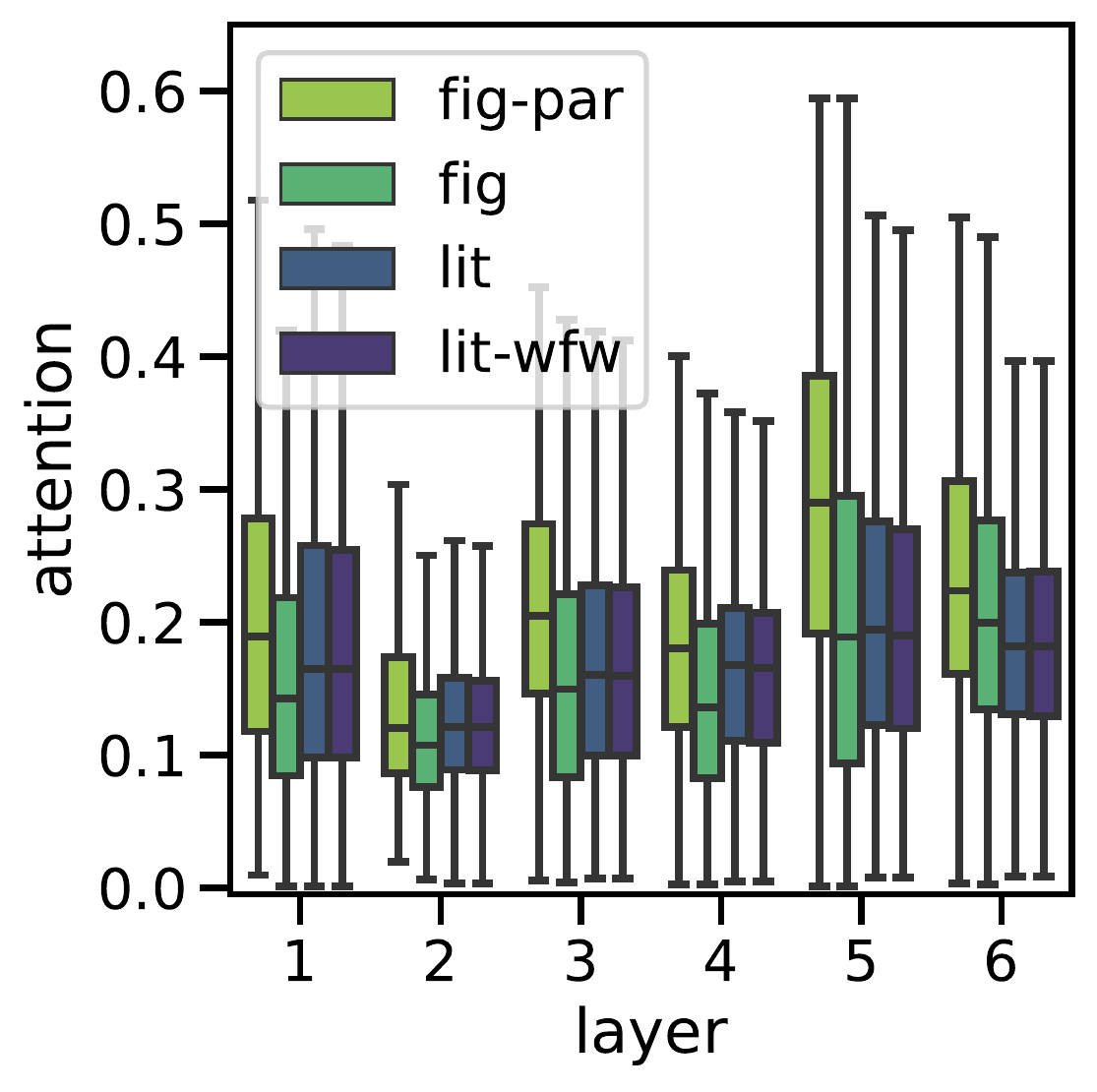}
        \caption{PIE-PIE}
        \label{fig:ap_pie_pie}
    \end{subfigure}
    \begin{subfigure}[b]{0.47\columnwidth}\centering
    \includegraphics[width=\textwidth]{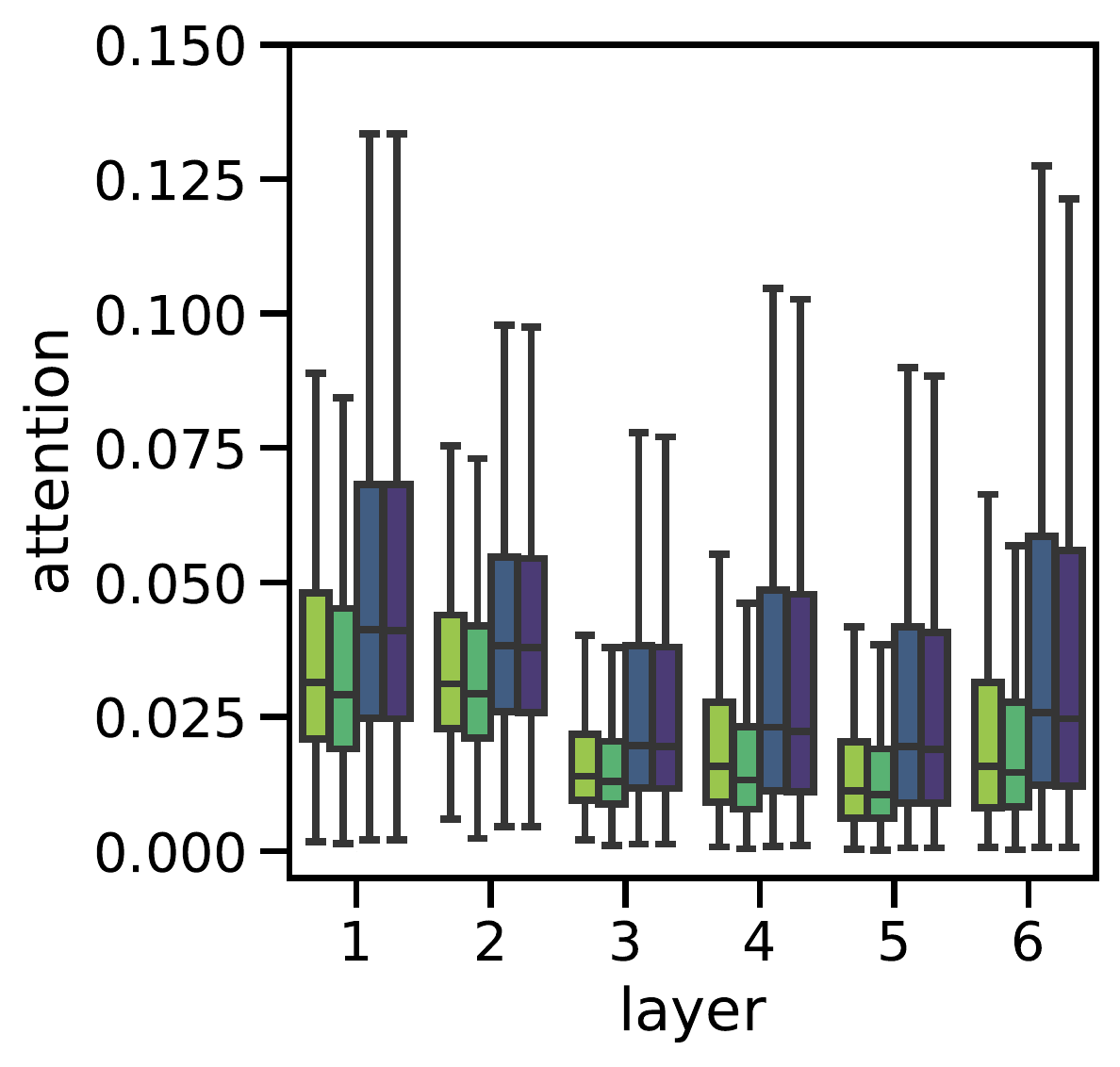}
        \caption{PIE-context}
    \end{subfigure}
    \begin{subfigure}[b]{0.47\columnwidth}\centering
    \includegraphics[width=\textwidth]{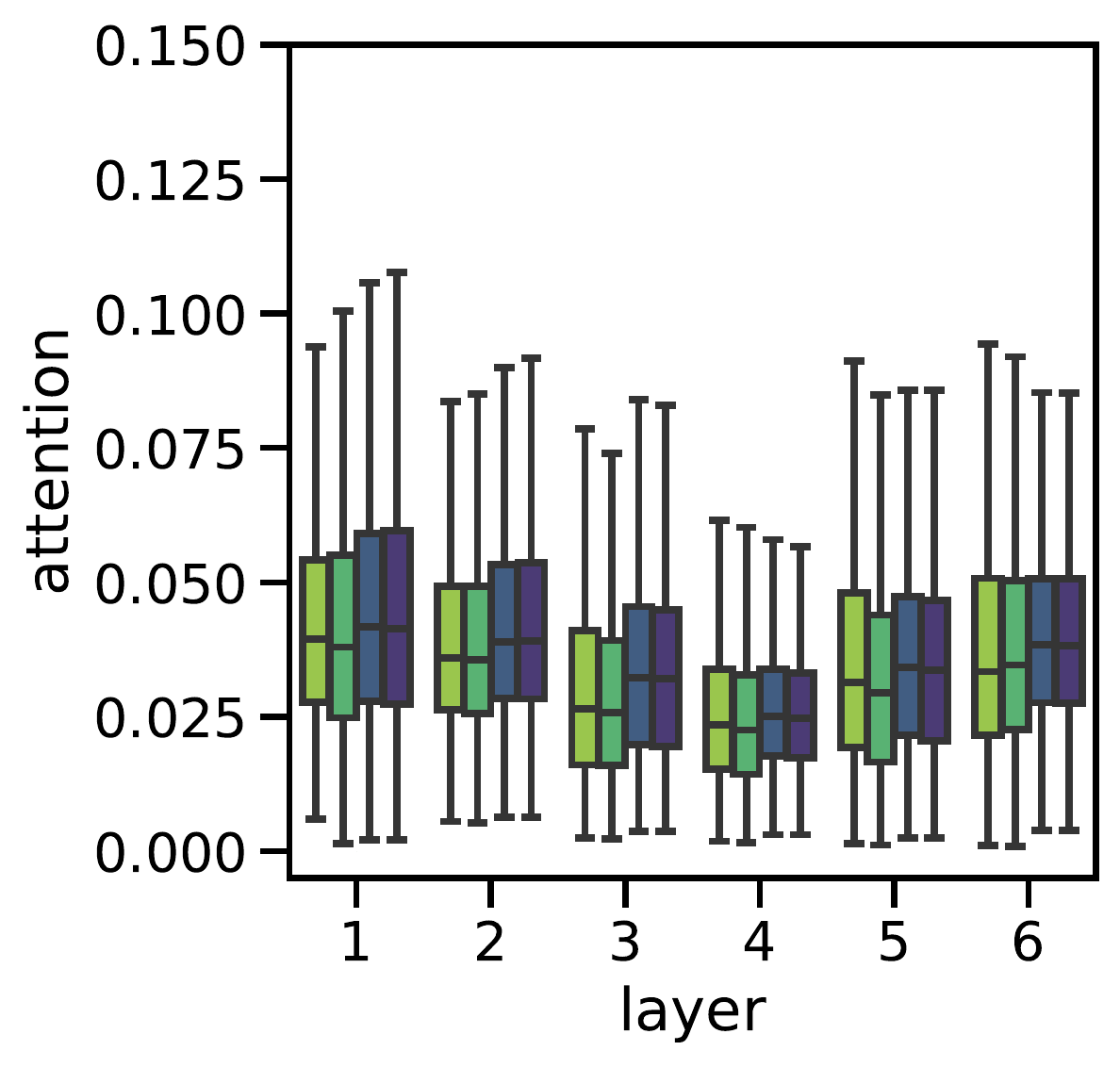}
        \caption{Context-PIE}
    \end{subfigure}
    \begin{subfigure}[b]{0.515\columnwidth}\centering
    \includegraphics[width=\textwidth]{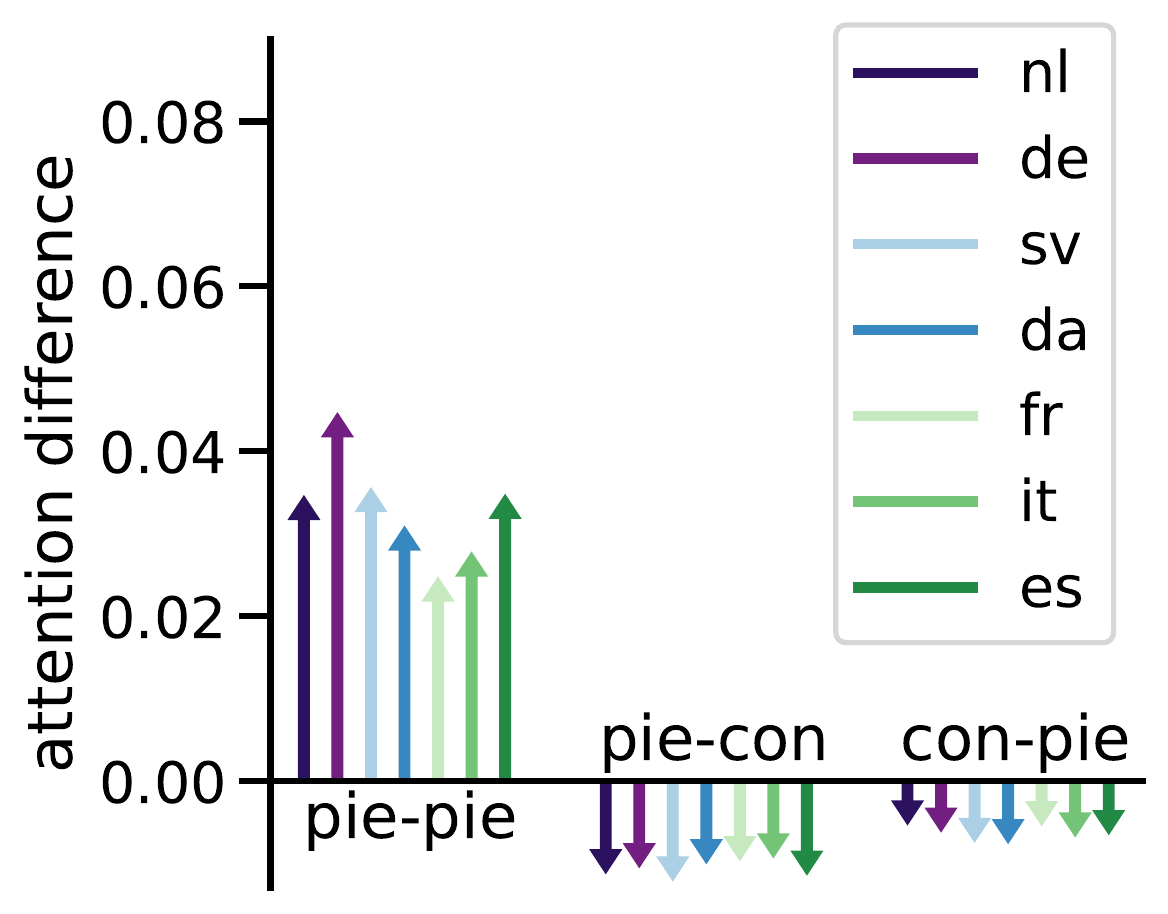}
        \caption{Language comparison}
    \end{subfigure}
\caption{Encoder self-attention distributions, illustrating attention within the PIE and the interaction between the PIE and its context, for the identical data subset.}
\label{fig:ap_encoder_attention_identical}
\end{figure}
\begin{figure}[!b]
    \centering\small
    \begin{subfigure}[b]{0.48\columnwidth}\centering
    \includegraphics[width=0.94\textwidth]{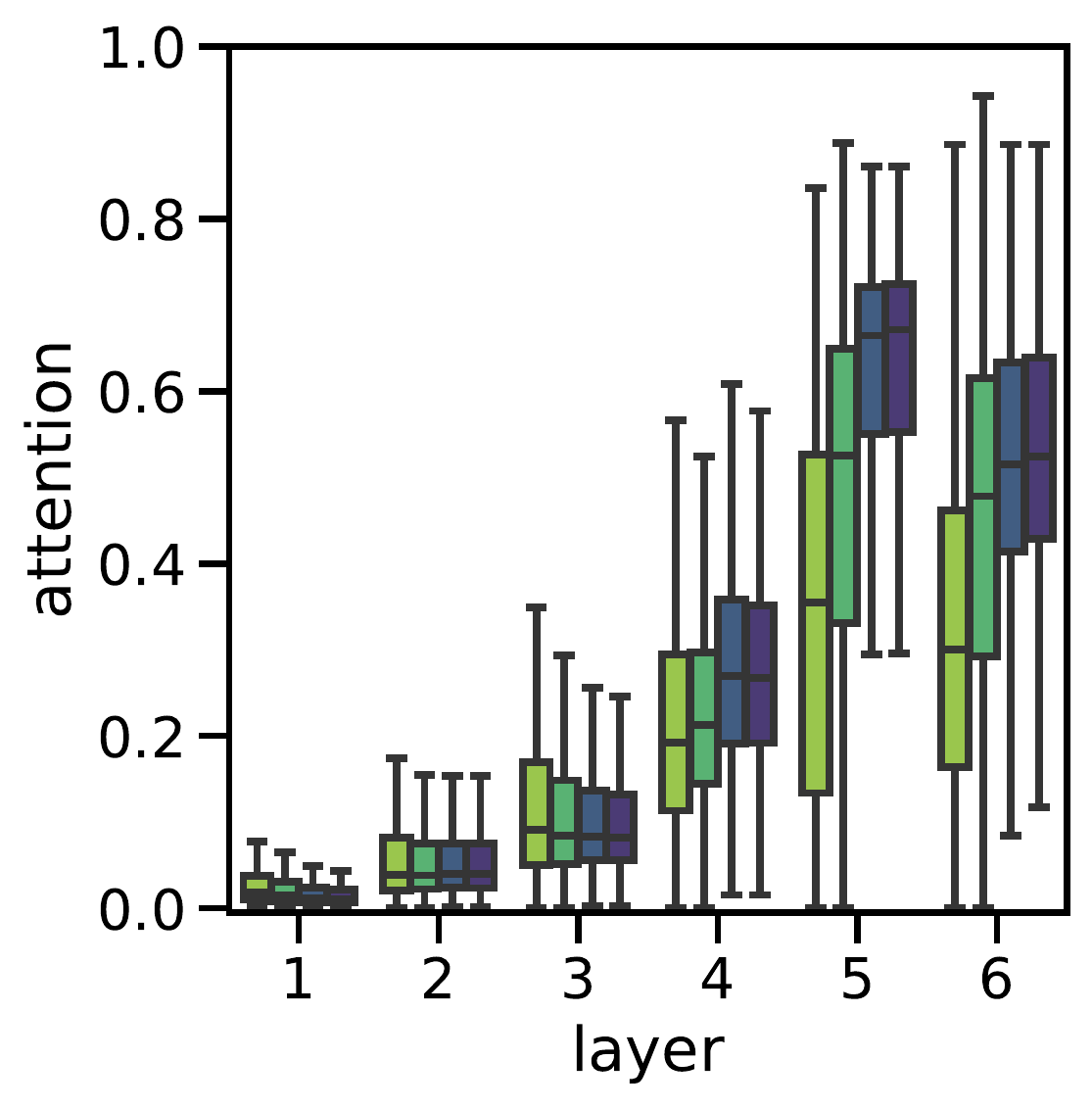}
    \caption{Target - PIE noun}
    \end{subfigure}
    \begin{subfigure}[b]{0.48\columnwidth}\centering
    \includegraphics[width=0.94\textwidth]{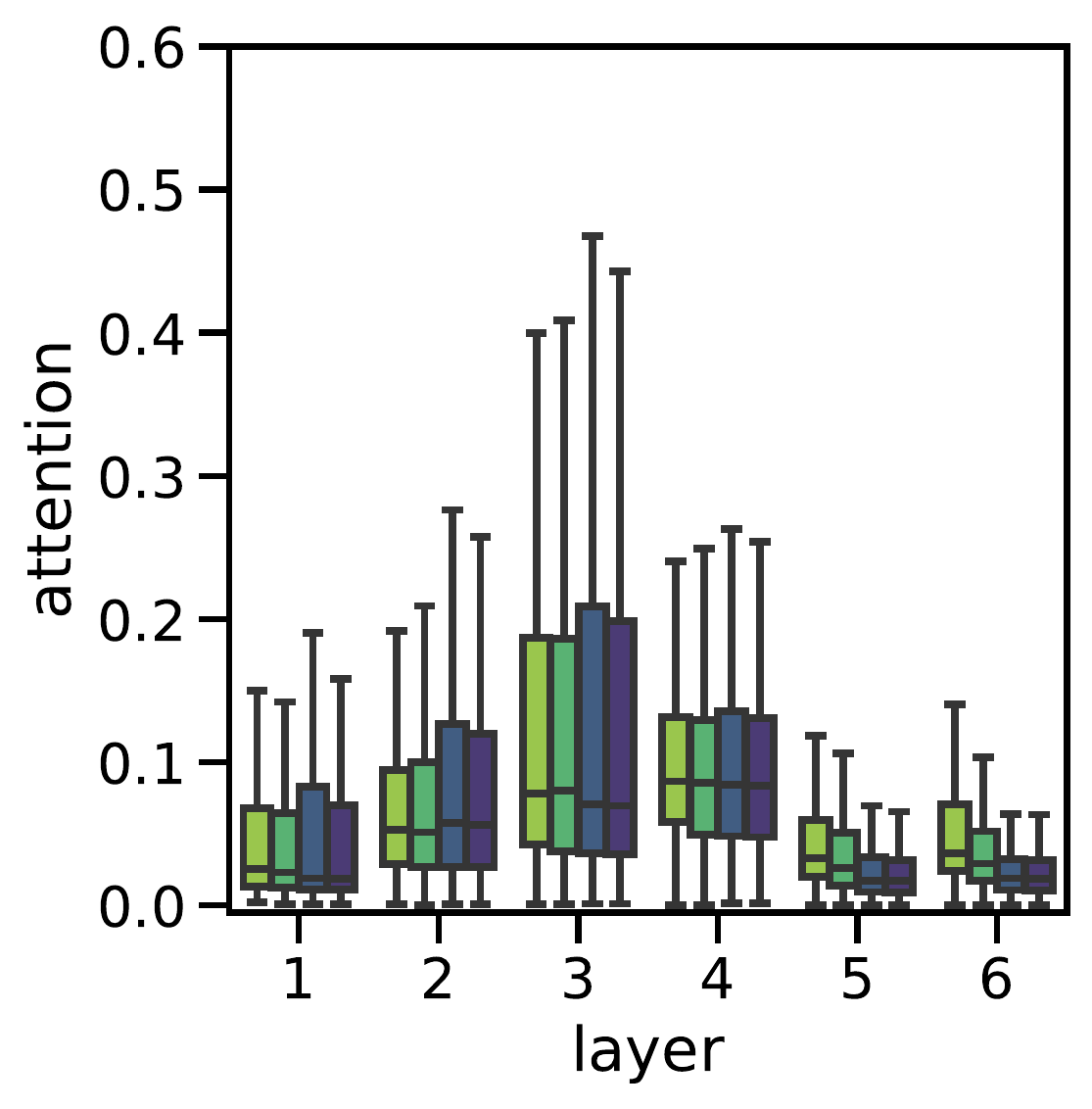}
    \caption{Target - PIE other}
    \end{subfigure}
    \begin{subfigure}[b]{0.48\columnwidth}\centering
    \includegraphics[width=0.94\textwidth]{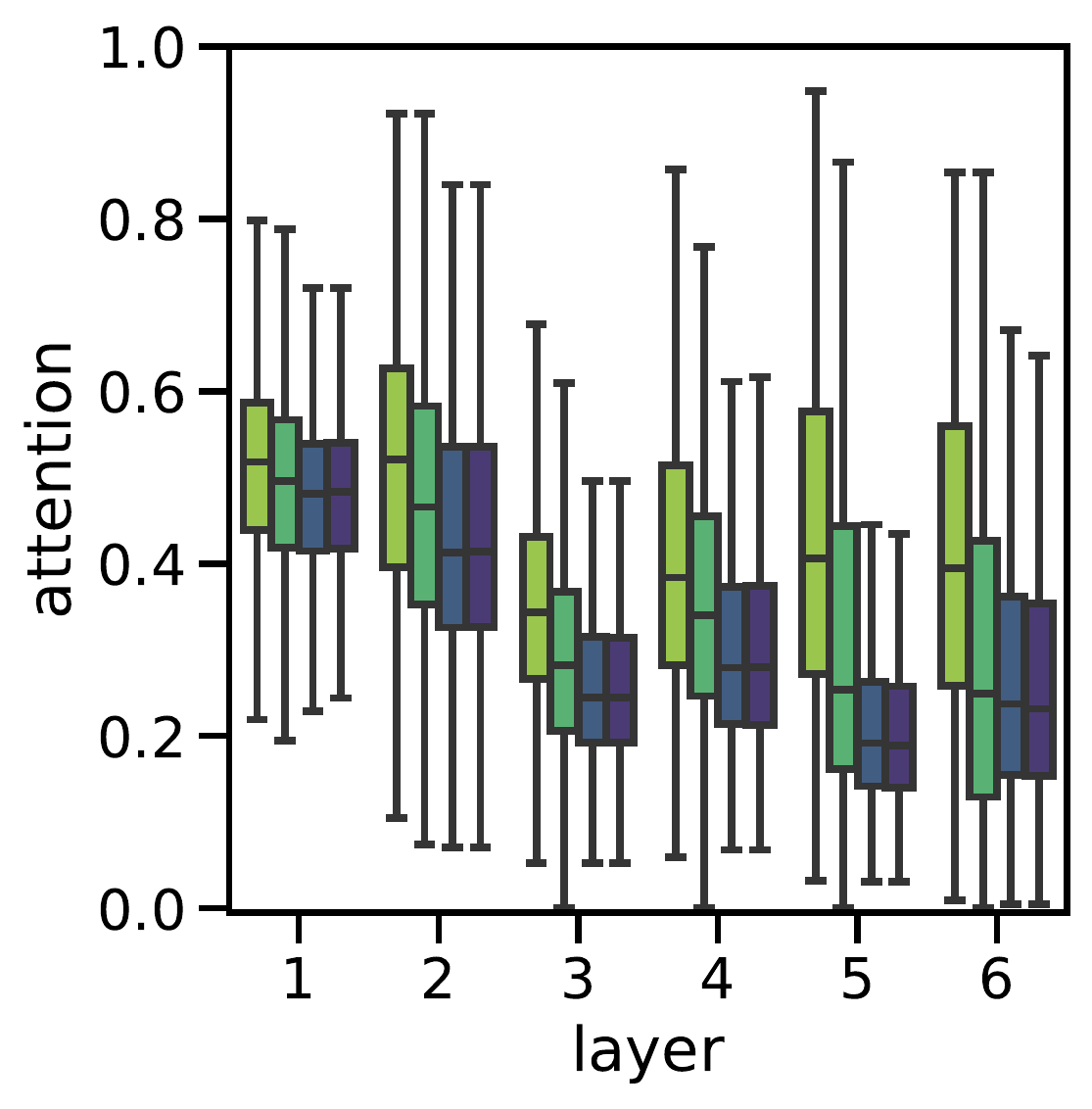}
    \caption{Target - \texttt{</s>}}
    \end{subfigure}
    \begin{subfigure}[b]{0.51\columnwidth}\centering
    \includegraphics[width=\textwidth]{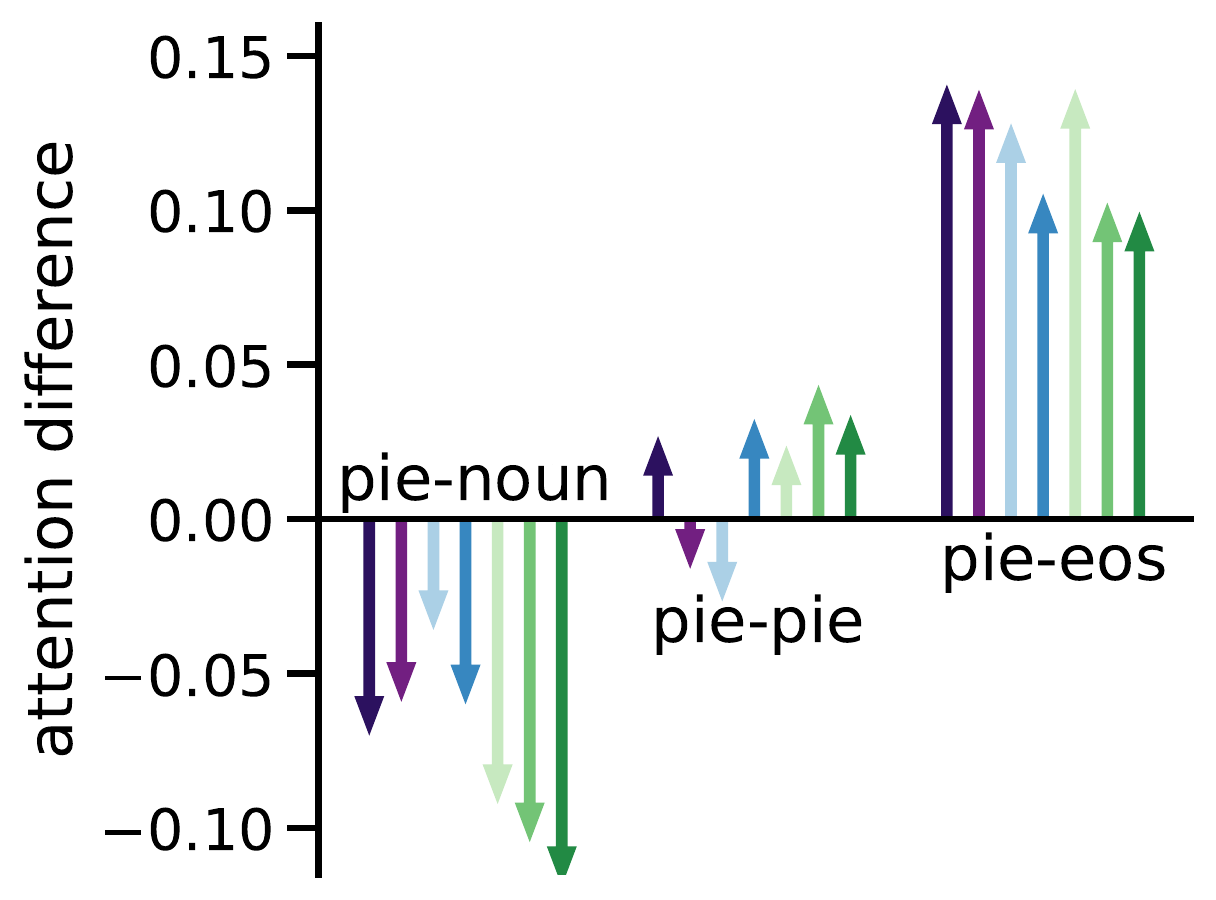}
    \caption{Language comparison}
    \end{subfigure}
\caption{Cross-attention distributions from the translation of one PIE noun on the target side to that noun on the source side, for the identical data subset.}
\label{fig:ap_cross_attention_identical}
\end{figure}
\paragraph{PIE identical matches}
We first use a subset that only includes samples for which MAGPIE reports an \textbf{identical match} between the PIE and the English sentence, that includes 17k samples.
This subset excludes sentences with modifications to the typical surface form of a PIE, such as upper-cased tokens or insertions of a token into the PIE (e.g. ``That gossip of a man spilled \textit{all} of the beans.'').

Figure~\ref{fig:ap_encoder_attention_identical} shows the three attention patterns previously discussed for the encoder's self-attention -- i.e. attention from the PIE to the PIE, attention from the PIE to the context, and from the context to the PIE. Overall, the patterns resemble those discussed in the main text, apart from Figure~\ref{fig:ap_pie_pie}, where figurative instances do not display consistently higher attention weights compared to literal instances, although the \textit{fig-par} subset does.

This procedure is repeated for the cross-attention distributions. Figure~\ref{fig:ap_cross_attention_identical} depicts the three patterns discussed in the main paper -- i.e. from the aligned target-side tokens to the PIE noun, to another PIE token, and to \texttt{</s>} -- for this data subset, providing the same qualitative findings.

\begin{figure}[!b]
    \centering
    \begin{subfigure}[b]{0.455\columnwidth}\centering
    \includegraphics[width=0.97\textwidth]{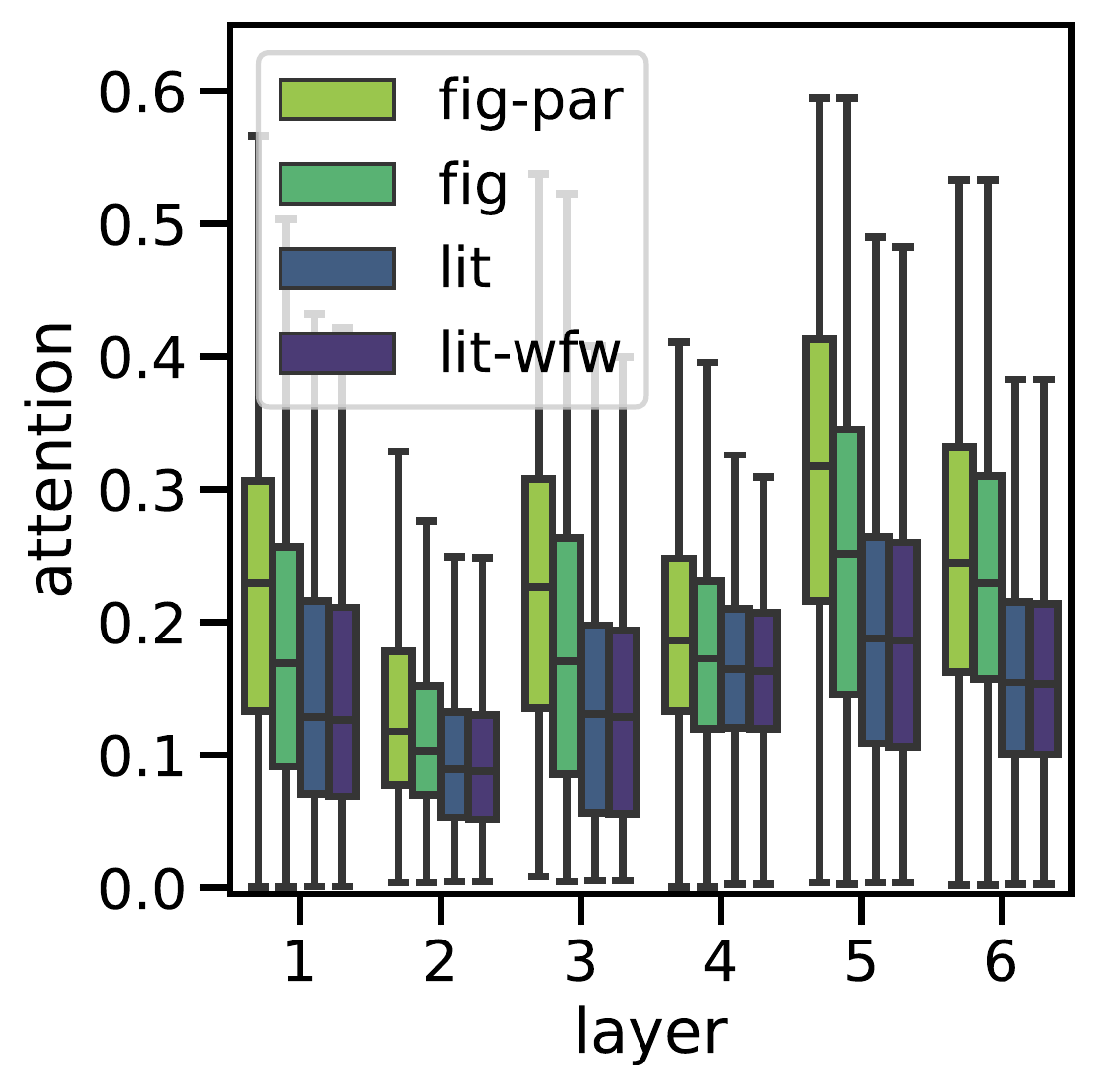}
        \caption{PIE-PIE}
    \end{subfigure}
    \begin{subfigure}[b]{0.47\columnwidth}\centering
    \includegraphics[width=\textwidth]{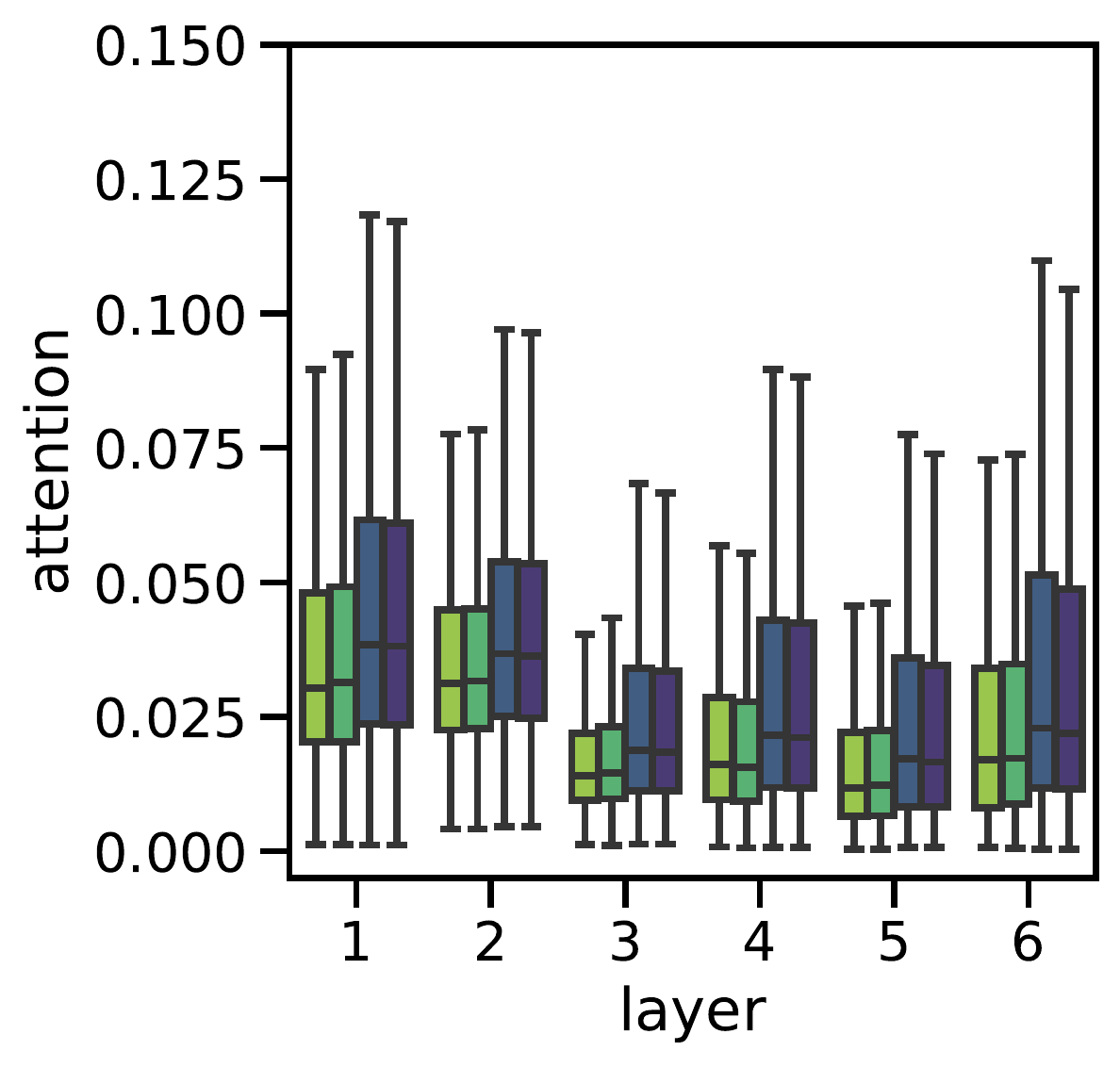}
        \caption{PIE-context}
    \end{subfigure}
    \begin{subfigure}[b]{0.47\columnwidth}\centering
    \includegraphics[width=\textwidth]{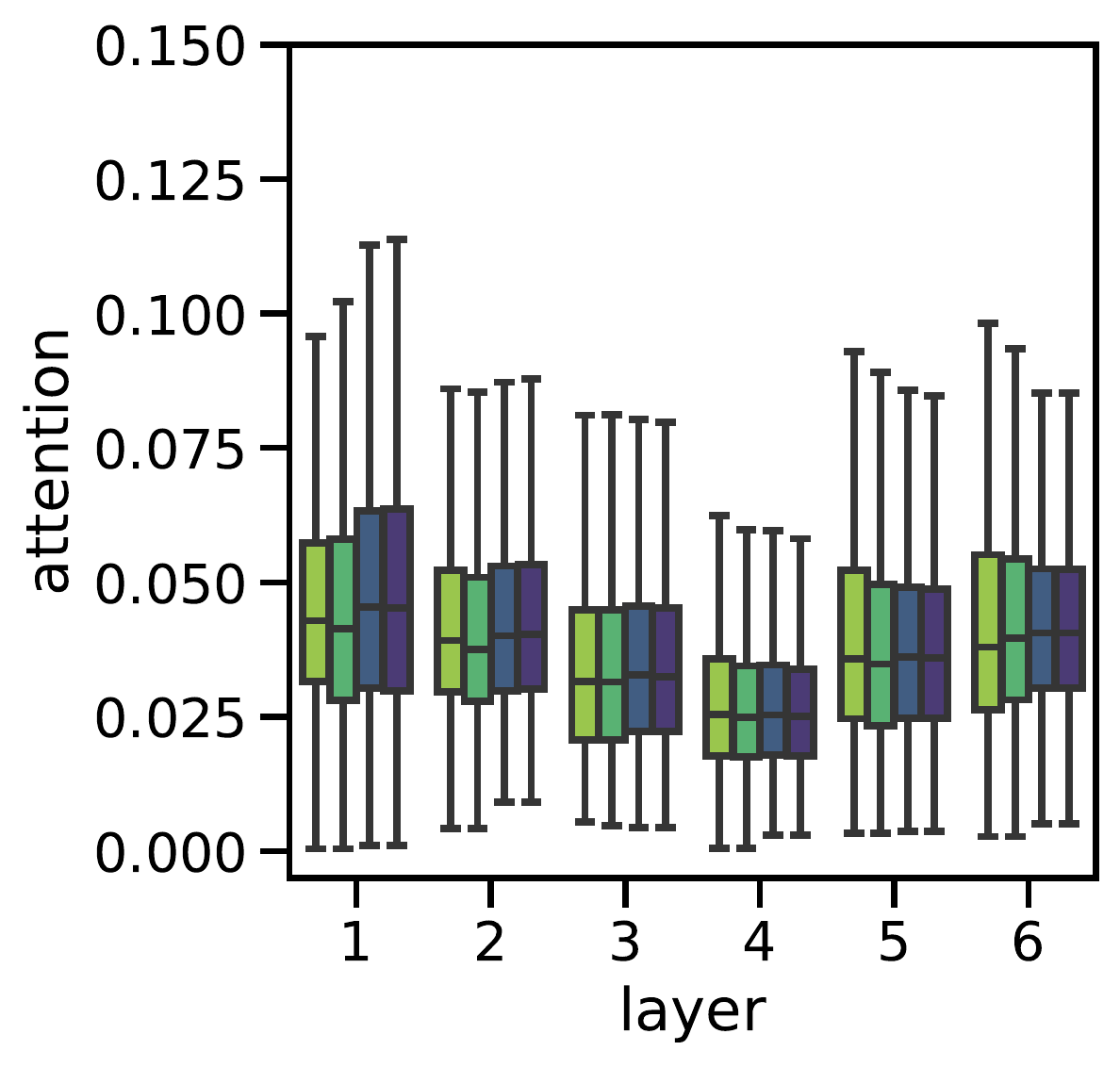}
        \caption{Context-PIE}
    \end{subfigure}
    \begin{subfigure}[b]{0.515\columnwidth}\centering
    \includegraphics[width=\textwidth]{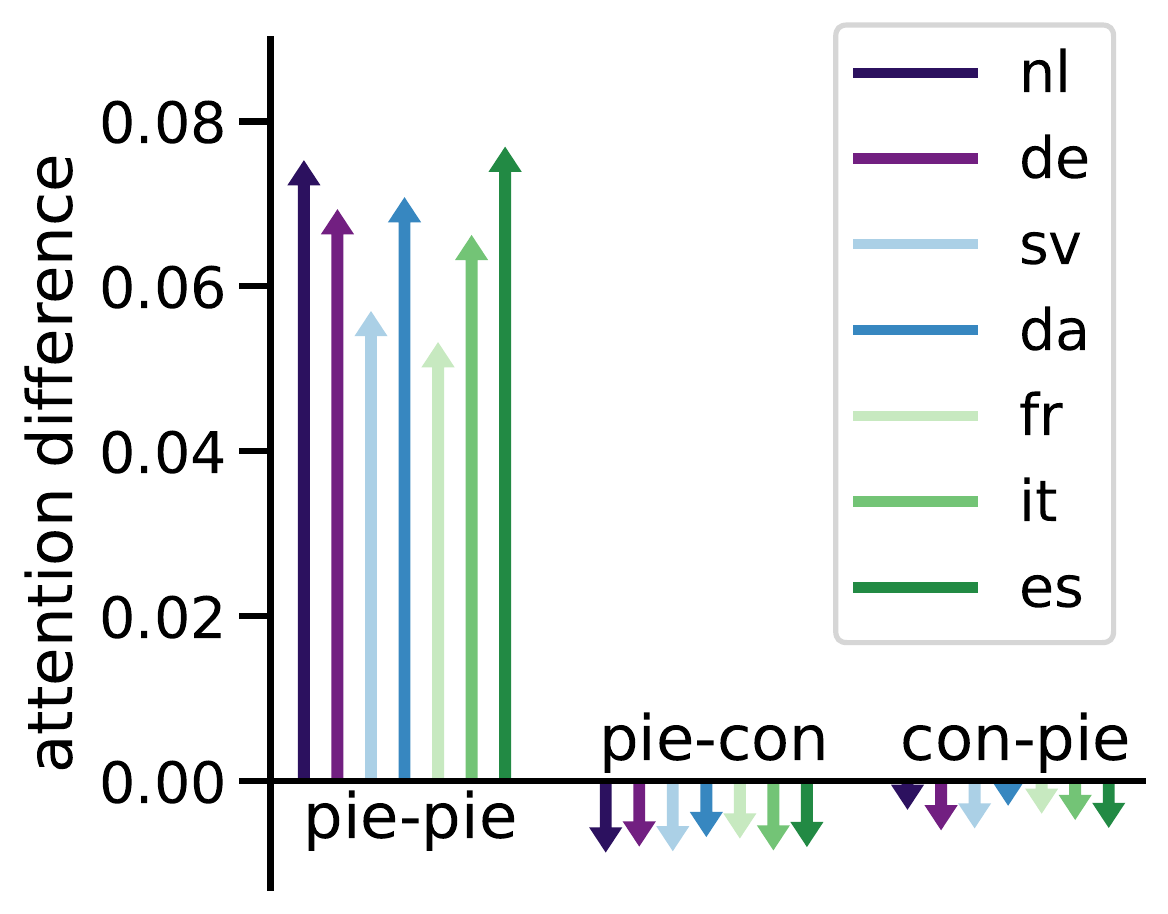}
        \caption{Language comparison}
    \end{subfigure}
\caption{Encoder self-attention distributions, showing attention within the PIE and the interaction between the PIE and its context, for the intersection data subset.}
\label{fig:ap_encoder_attention_intersection}
\end{figure}

\begin{figure}[!t]
    \centering\small
    \begin{subfigure}[b]{0.48\columnwidth}\centering
    \includegraphics[width=0.94\textwidth]{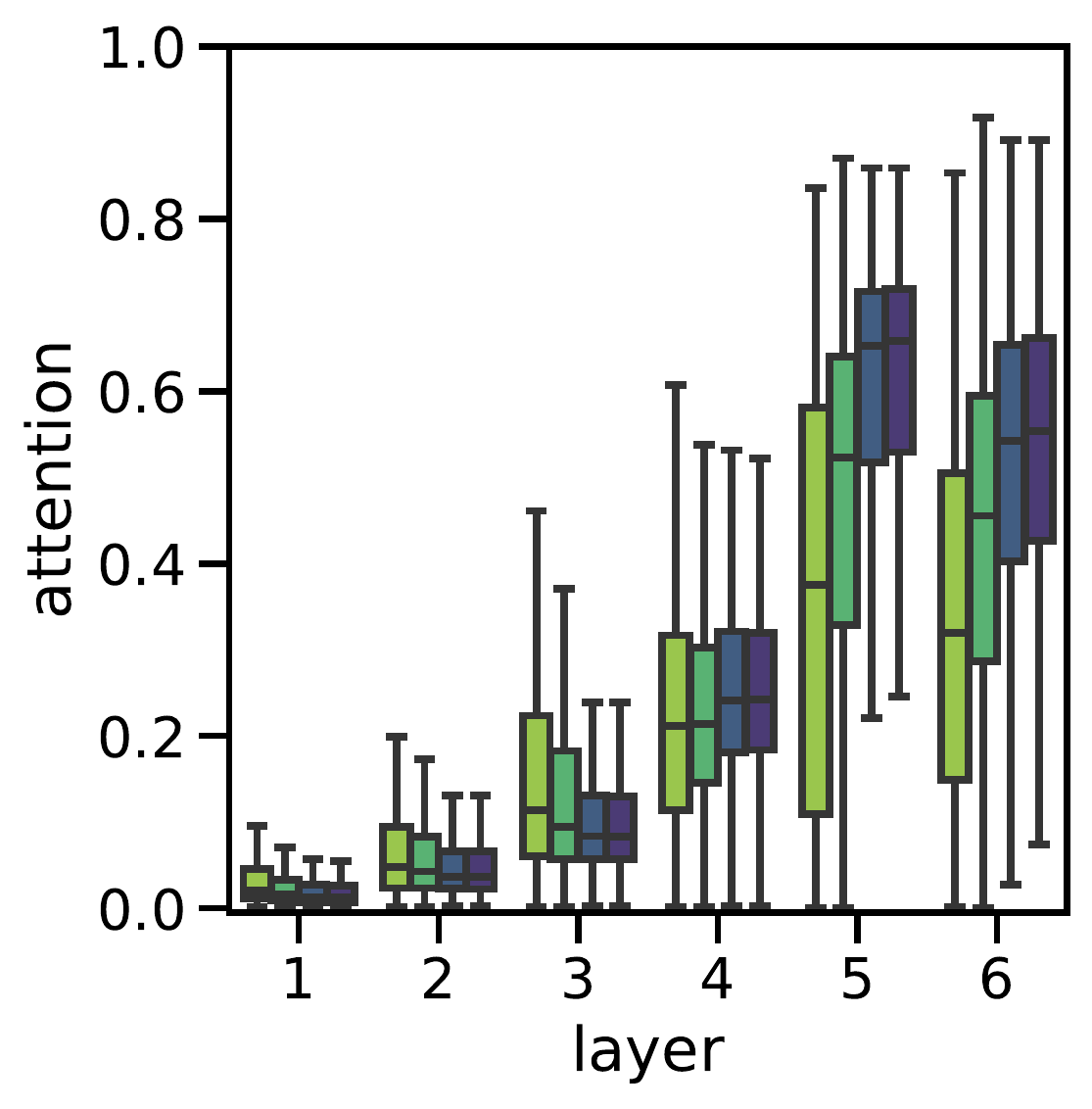}
    \caption{Target - PIE noun}
    \end{subfigure}
    \begin{subfigure}[b]{0.48\columnwidth}\centering
    \includegraphics[width=0.94\textwidth]{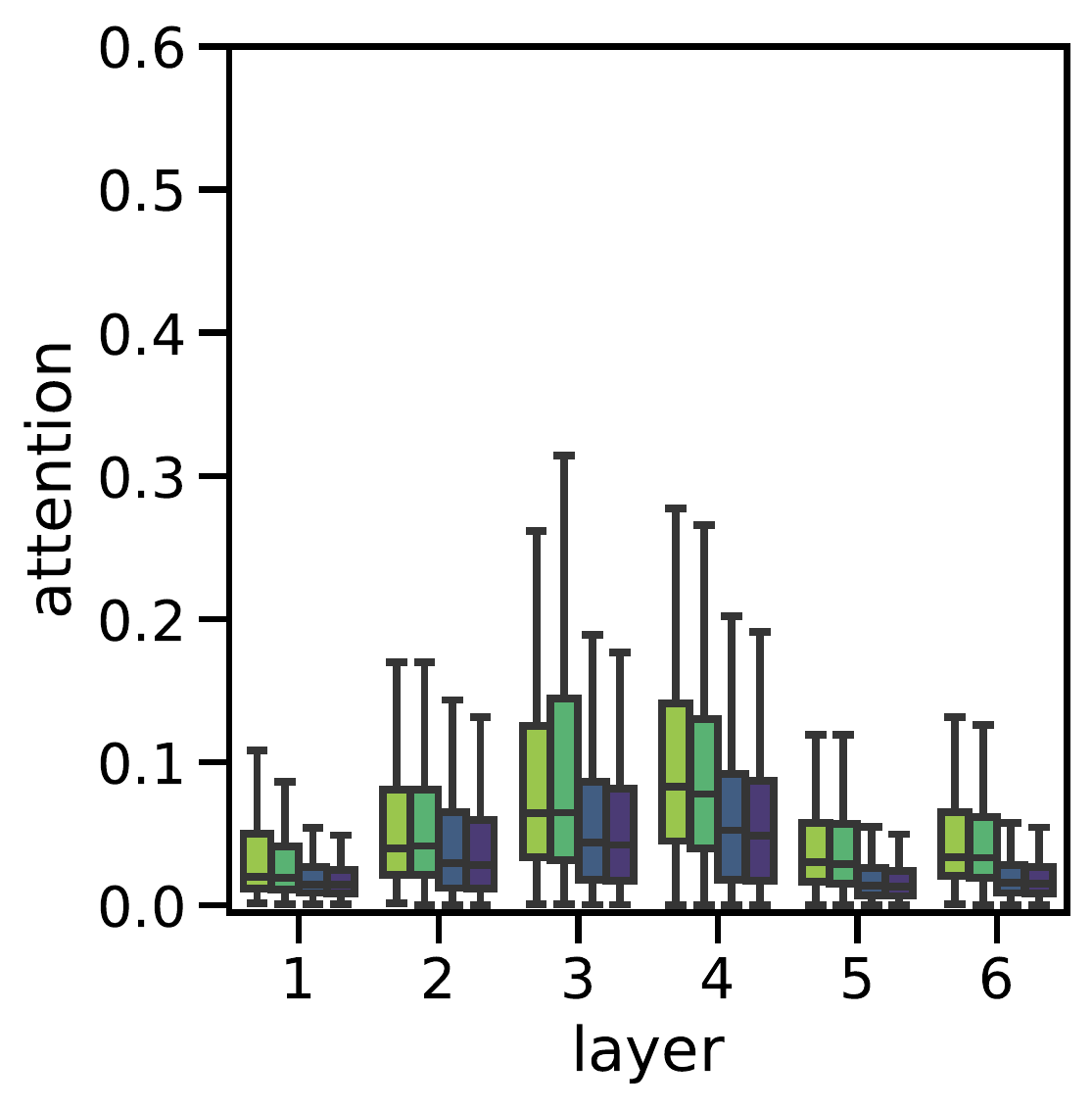}
    \caption{Target - PIE other}
    \end{subfigure}
    \begin{subfigure}[b]{0.48\columnwidth}\centering
    \includegraphics[width=0.94\textwidth]{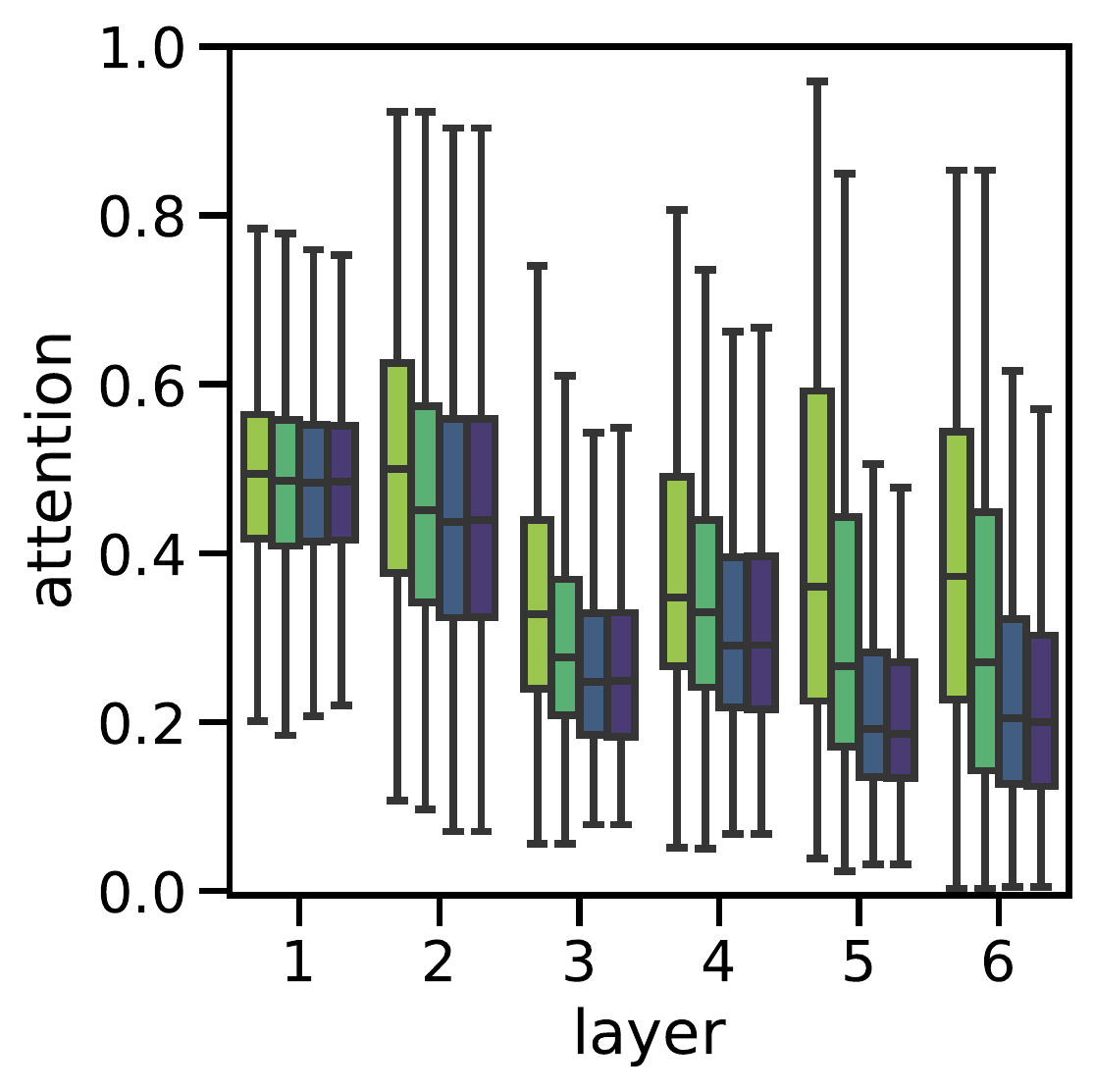}
    \caption{Target - \texttt{</s>}}
    \end{subfigure}
    \begin{subfigure}[b]{0.51\columnwidth}\centering
    \includegraphics[width=\textwidth]{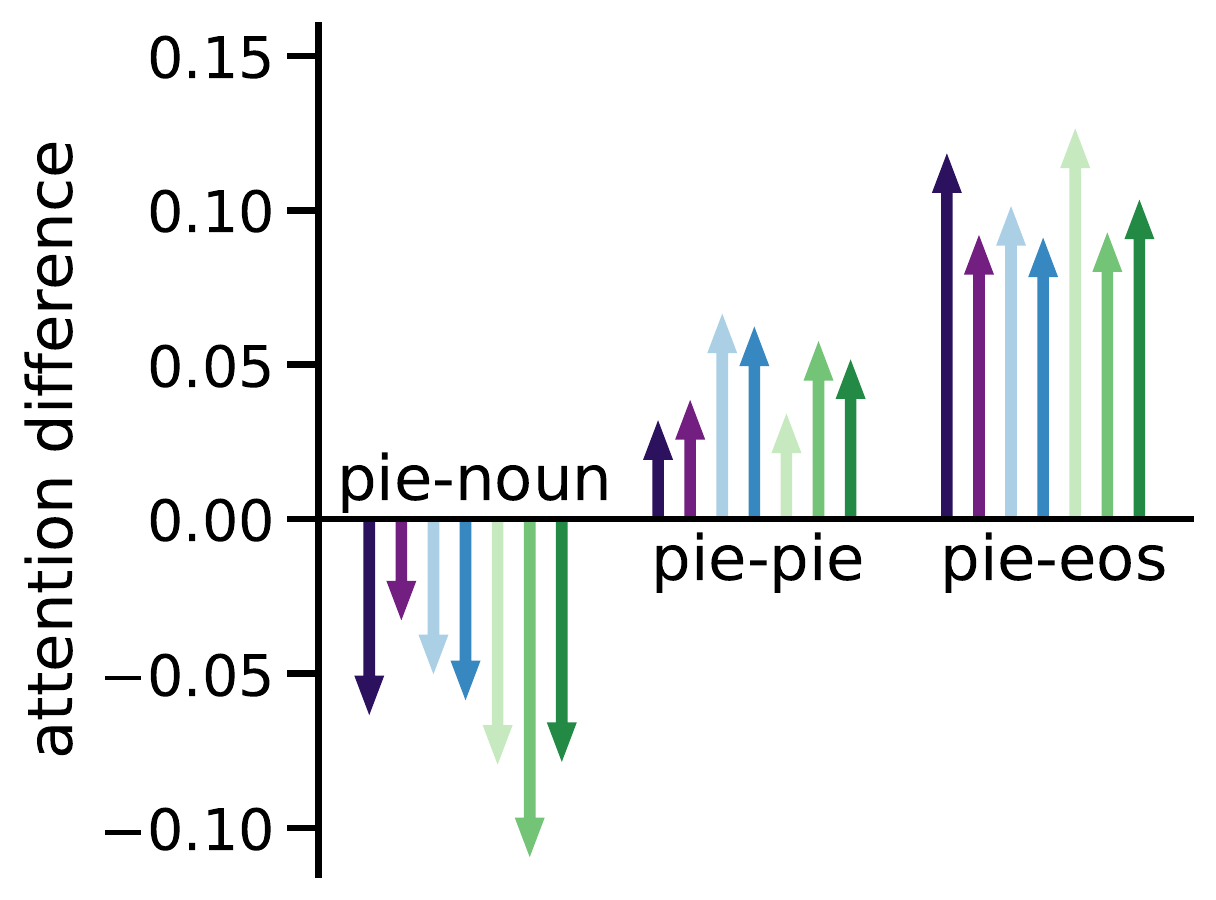}
    \caption{Language comparison}
    \end{subfigure}
\caption{Cross-attention distributions from the translation of one PIE noun on the target side to that noun on the source side, for the intersection data subset.}
\label{fig:ap_cross_attention_intersection}
\vspace{-0.3cm}
\end{figure}

\paragraph{Intersection of PIEs}
The second subset (referred to as \textbf{intersection}) considered is one that only contains idioms that are among all of the subsets of figurative, literal, paraphrased and word for word instances, covering 11k examples from the dataset.
The results for the encoder's self-attention patterns are shown in Figure~\ref{fig:ap_encoder_attention_intersection}. Figure~\ref{fig:ap_cross_attention_intersection} summarises the results for the cross-attention mechanisms.
These results lead to the same qualitative findings as mentioned in the main paper, and, in the encoder, the PIE to PIE attention patterns for figurative and literal PIEs are even more distinct.

\paragraph{Controlling PIE length}
To investigate the impact of the length of a PIE and the length of its context on the results, we now report additional measures over sentences, namely:
    \begin{itemize}[noitemsep]
        \item the \textbf{average number of MarianMT tokens} labelled as being part of the PIE (in MAGPIE words like prepositions and determiners are not counted as part of the PIE, so the annotation can be discontinuous);
        \item the \textbf{distance between the first and the last token} of the PIE (two tokens right next to each other have a distance of 1);
        \item the \textbf{relative position} of the tokens that are annotated as belonging to the PIE, which impacts the potential context size, but could also impact how a PIE `behaves';
        \item the average distance of the first position of the PIE's context tokens (PIE - 10) to the last position (PIE + 10) (\textbf{context length}).
    \end{itemize}

\begin{figure}[!h]
    \begin{subfigure}[b]{0.2\columnwidth}
        \includegraphics[width=\textwidth]{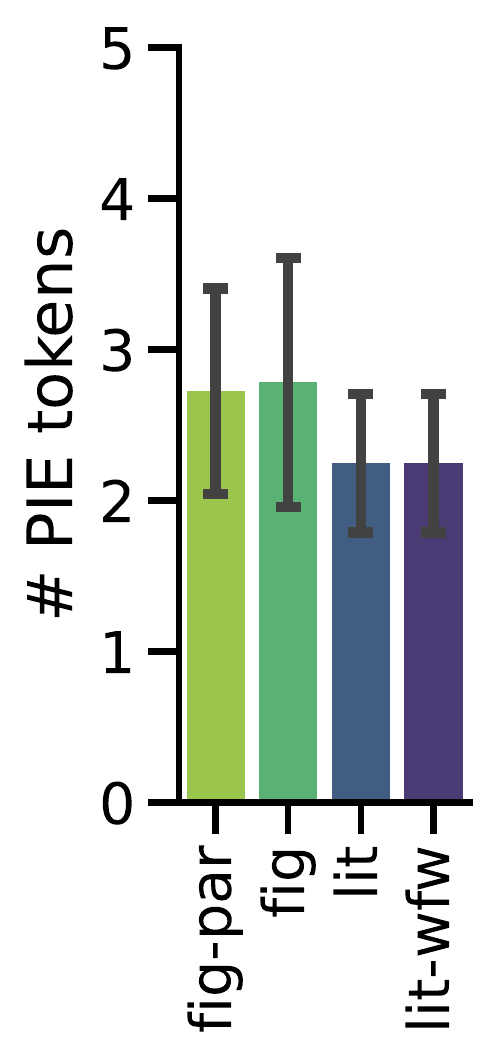}
    \end{subfigure}
    \begin{subfigure}[b]{0.2\columnwidth}
        \includegraphics[width=\textwidth]{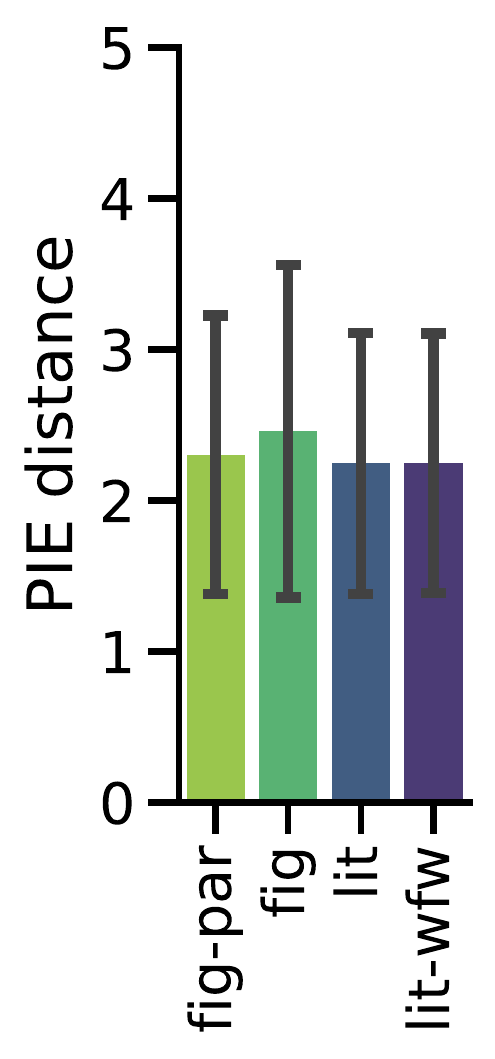}
    \end{subfigure}
    \begin{subfigure}[b]{0.225\columnwidth}
        \includegraphics[width=\textwidth]{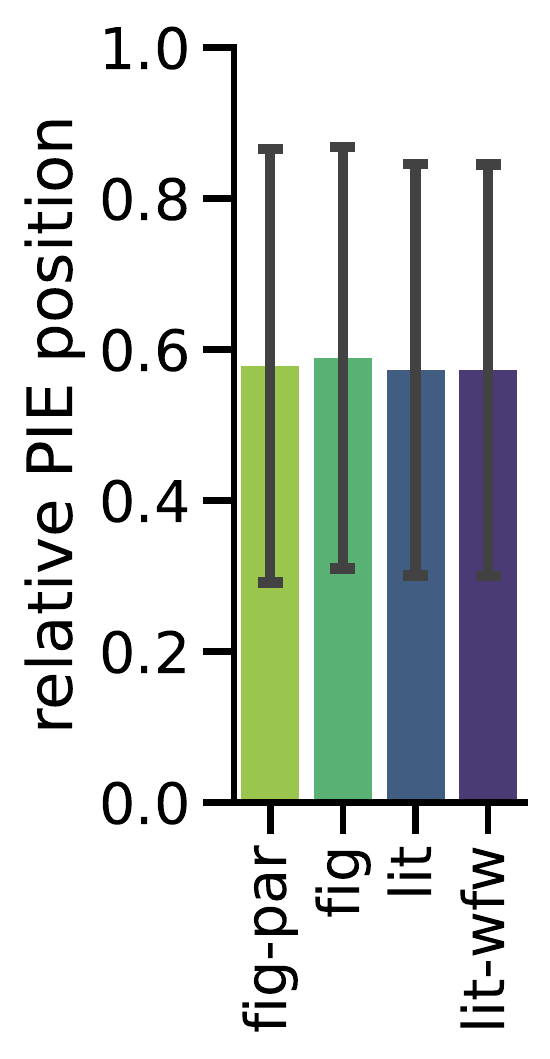}
    \end{subfigure}
    \begin{subfigure}[b]{0.22\columnwidth}
        \includegraphics[width=\textwidth]{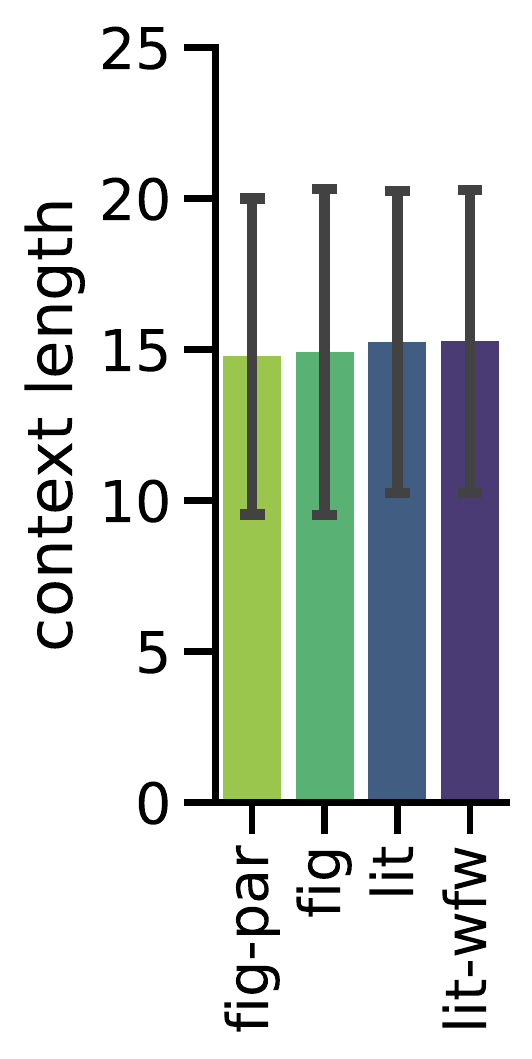}
    \end{subfigure}
    \centering
    \caption{Length statistics for the four categories of PIEs (\textit{fig-par}, \textit{lit-wfw}, \textit{fig}, \textit{lit}). Error bars indicate standard deviations over sentences.}
    \label{fig:extra_stats}
\end{figure}

\begin{figure}[!h]
    \centering
    \begin{subfigure}[b]{0.455\columnwidth}\centering
    \includegraphics[width=\textwidth]{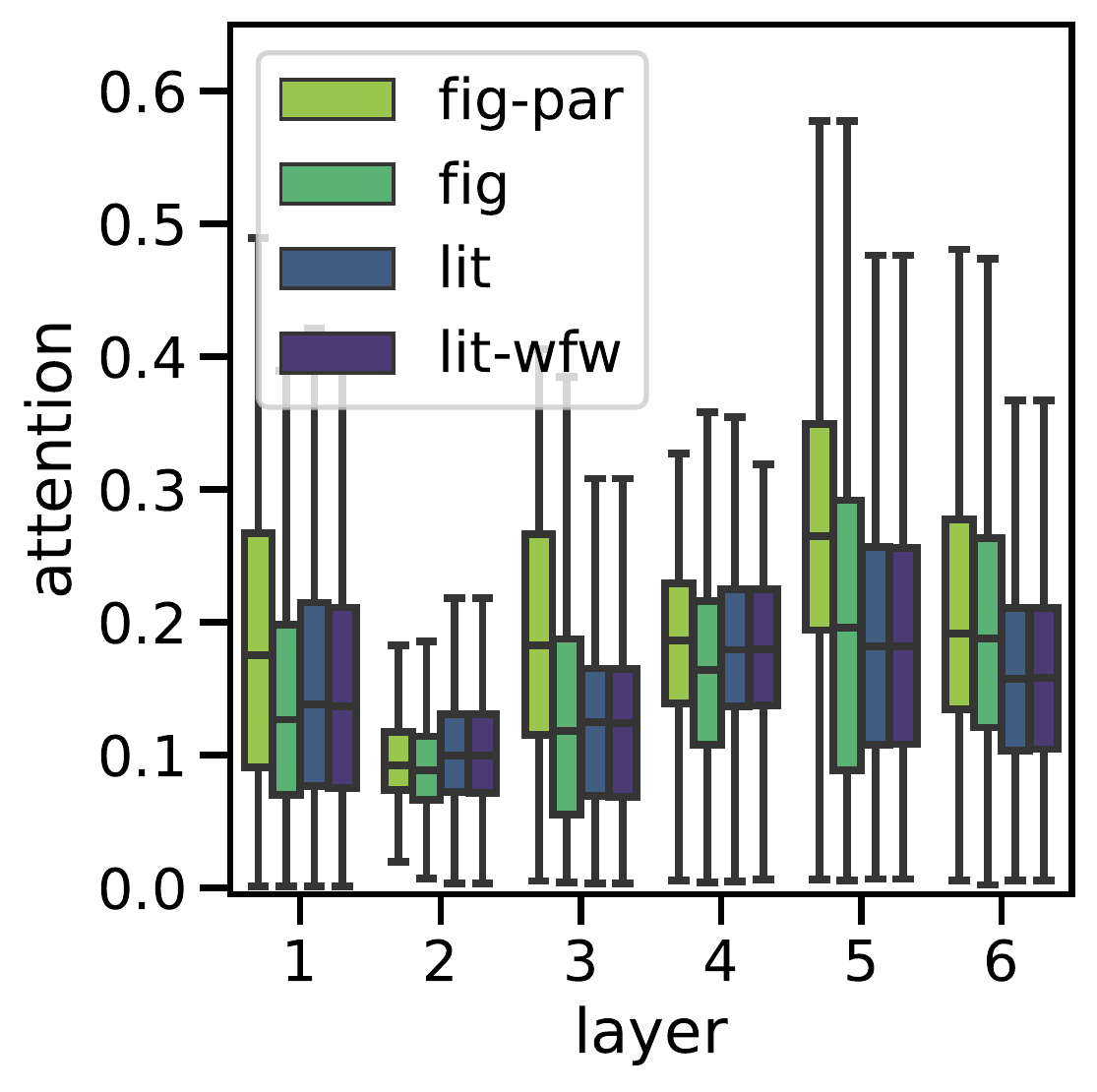}
        \caption{PIE-PIE}
    \end{subfigure}
    \begin{subfigure}[b]{0.47\columnwidth}\centering
    \includegraphics[width=\textwidth]{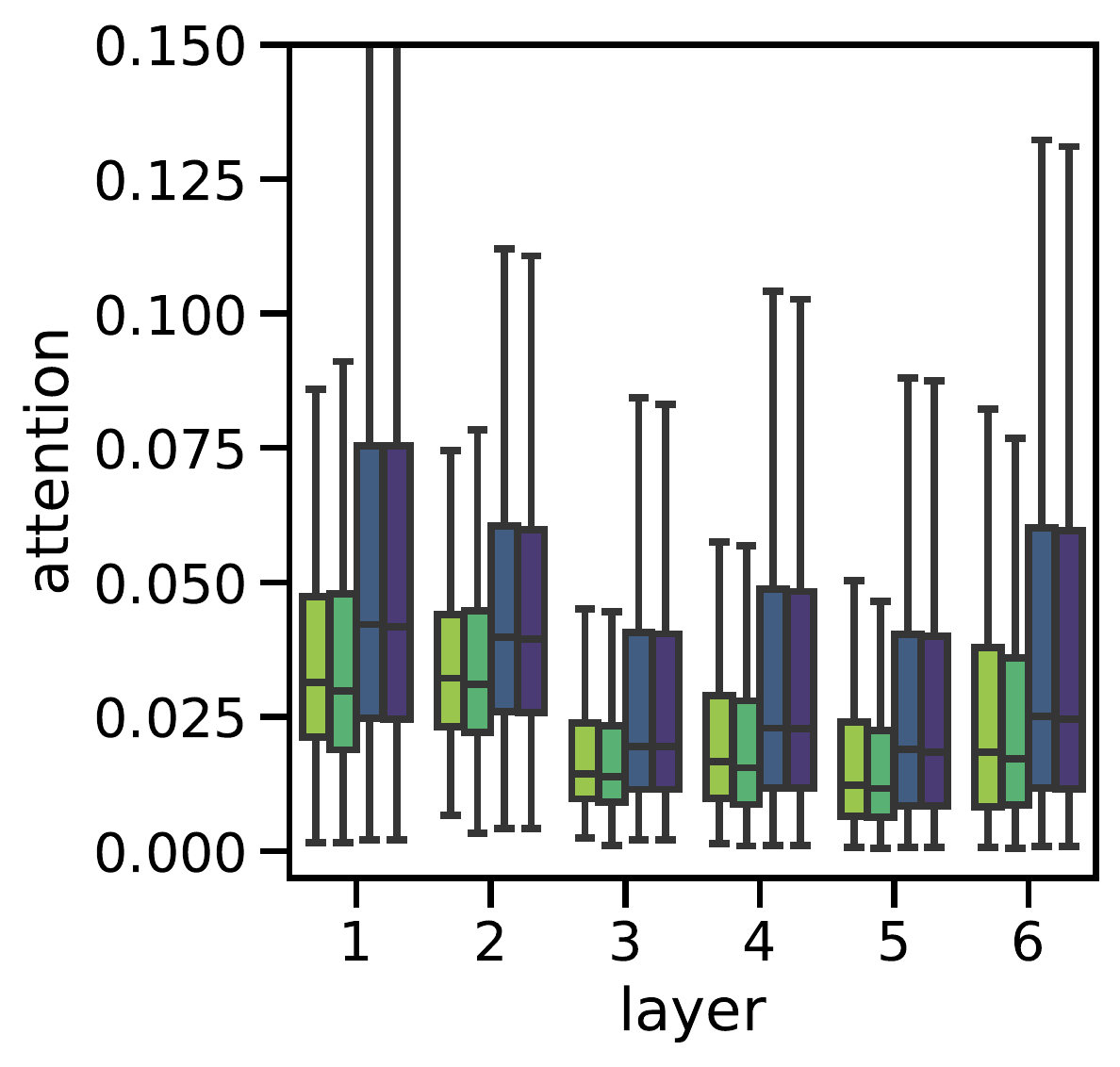}
        \caption{PIE-context}
    \end{subfigure}
    \begin{subfigure}[b]{0.47\columnwidth}\centering
    \includegraphics[width=\textwidth]{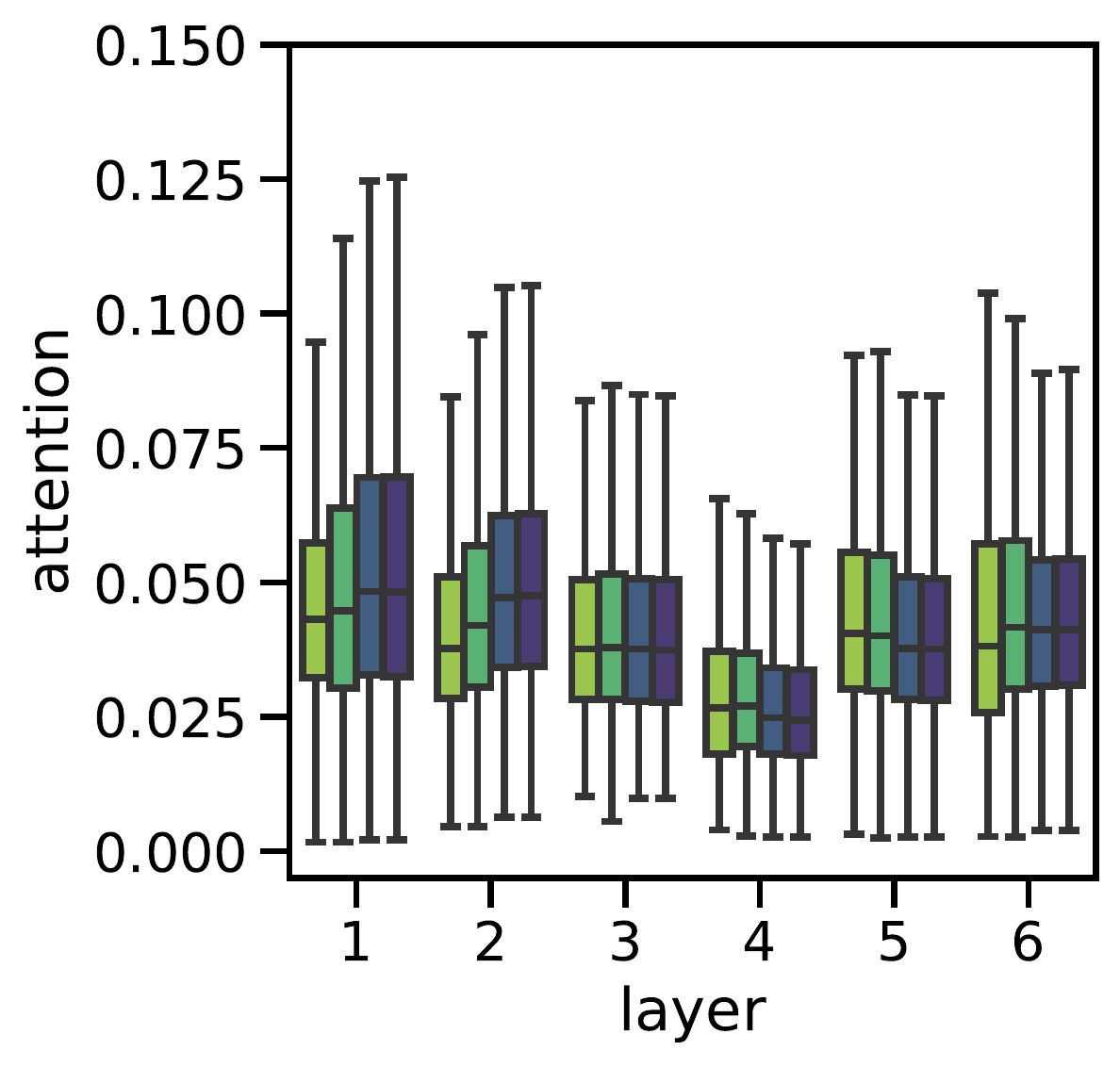}
        \caption{Context-PIE}
    \end{subfigure}
    \begin{subfigure}[b]{0.515\columnwidth}\centering
    \includegraphics[width=\textwidth]{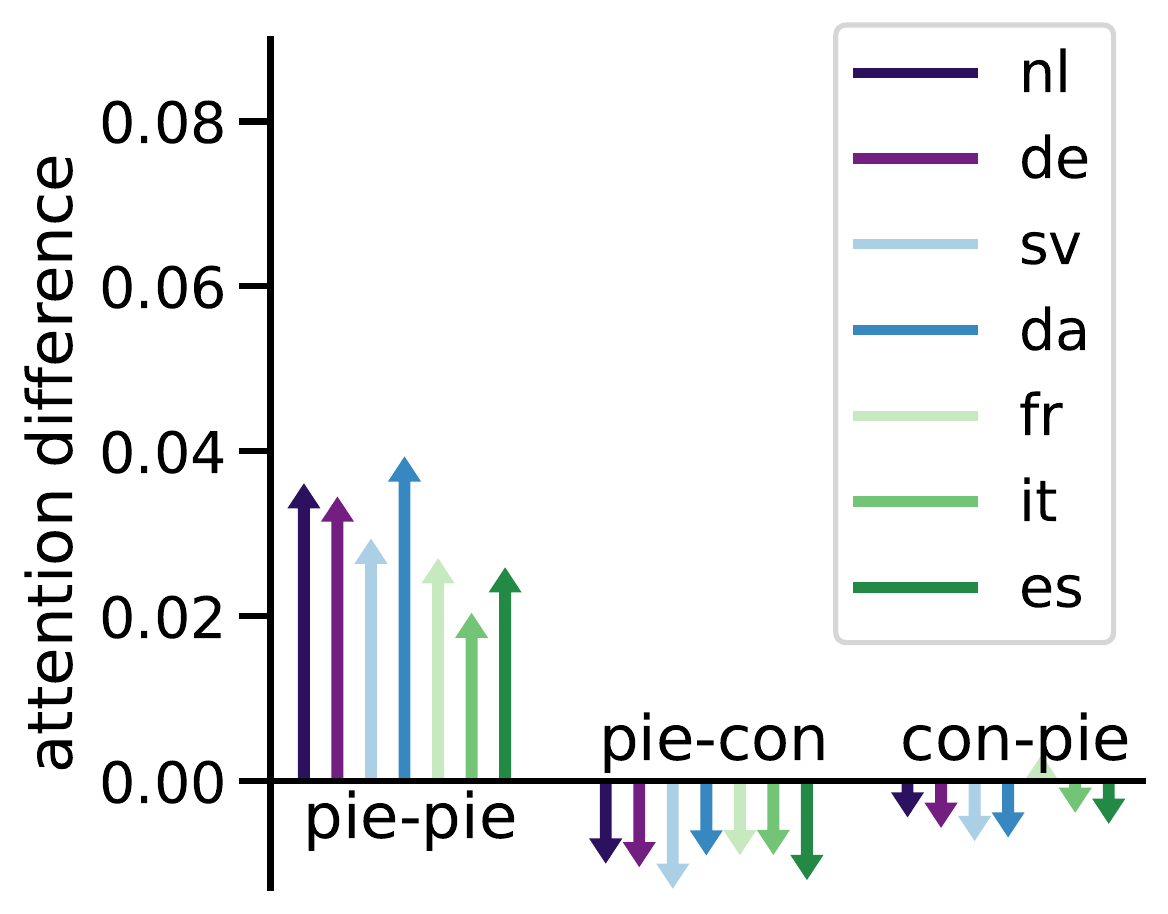}
        \caption{Language comparison}
    \end{subfigure}
\caption{Encoder self-attention distributions, showing attention within the PIE and the interaction between the PIE and its context, for the length controlled subset.}
\label{fig:ap_encoder_attention_short}
\vspace{-.3cm}
\end{figure}

Figure~\ref{fig:extra_stats} summarises these statistics for the MAGPIE PIEs. The last two metrics are very stable across categories, with an average relative position of 0.57 for PIEs (0.58 for figurative, 0.56 for literal), and context lengths of 17.0 (17.0 for figurative, 17.1 for literal). The first two metrics indicate that figurative PIEs are a bit longer than literal PIEs (0.69 words), and that the distance between the first and the last word is slightly larger (0.46 positions).

\begin{figure}[!t]
    \centering\small
    \begin{subfigure}[b]{0.47\columnwidth}\centering
    \includegraphics[width=0.96\textwidth]{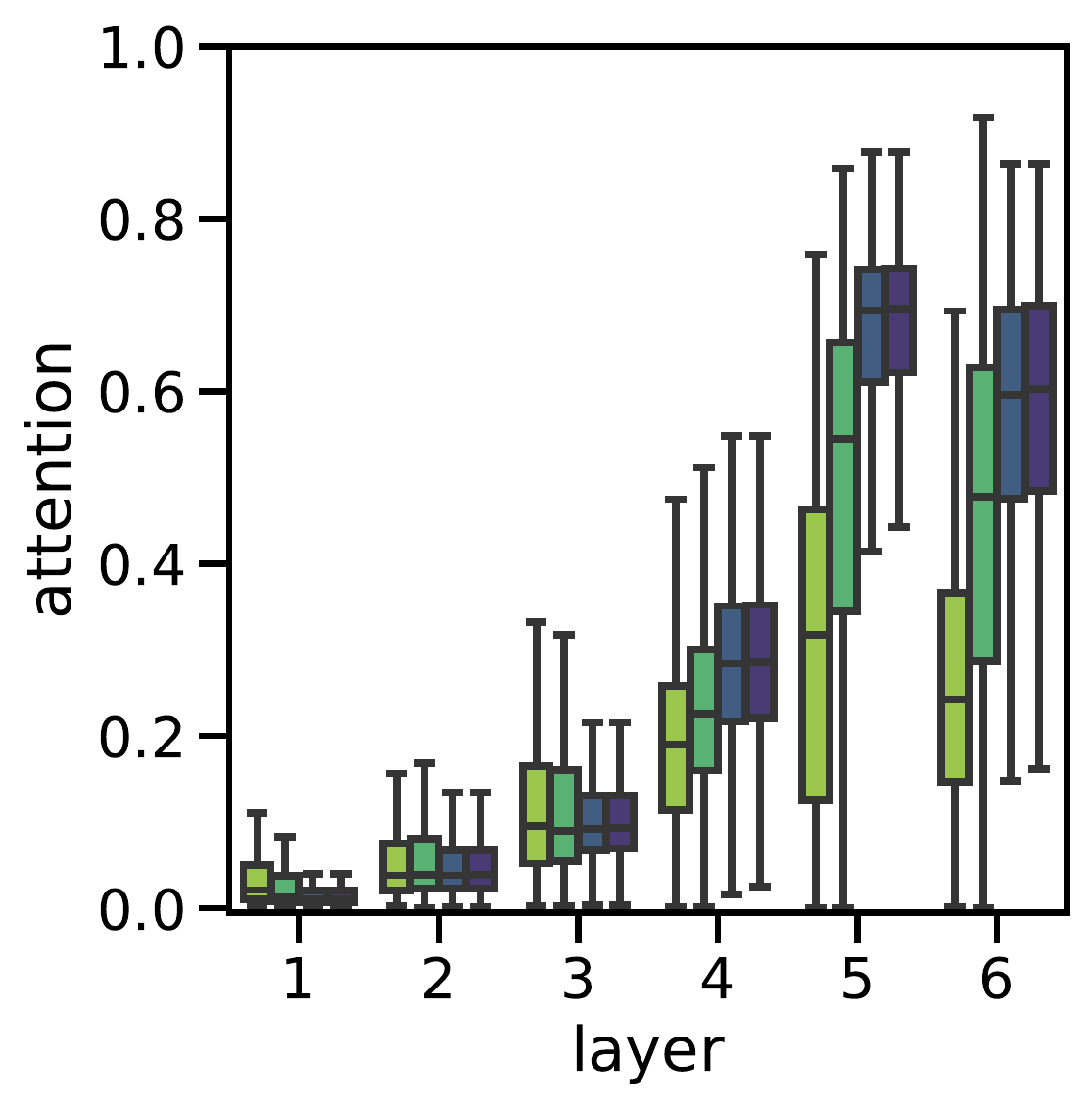}
    \caption{Target - PIE noun}
    \end{subfigure}
    \begin{subfigure}[b]{0.47\columnwidth}\centering
    \includegraphics[width=0.96\textwidth]{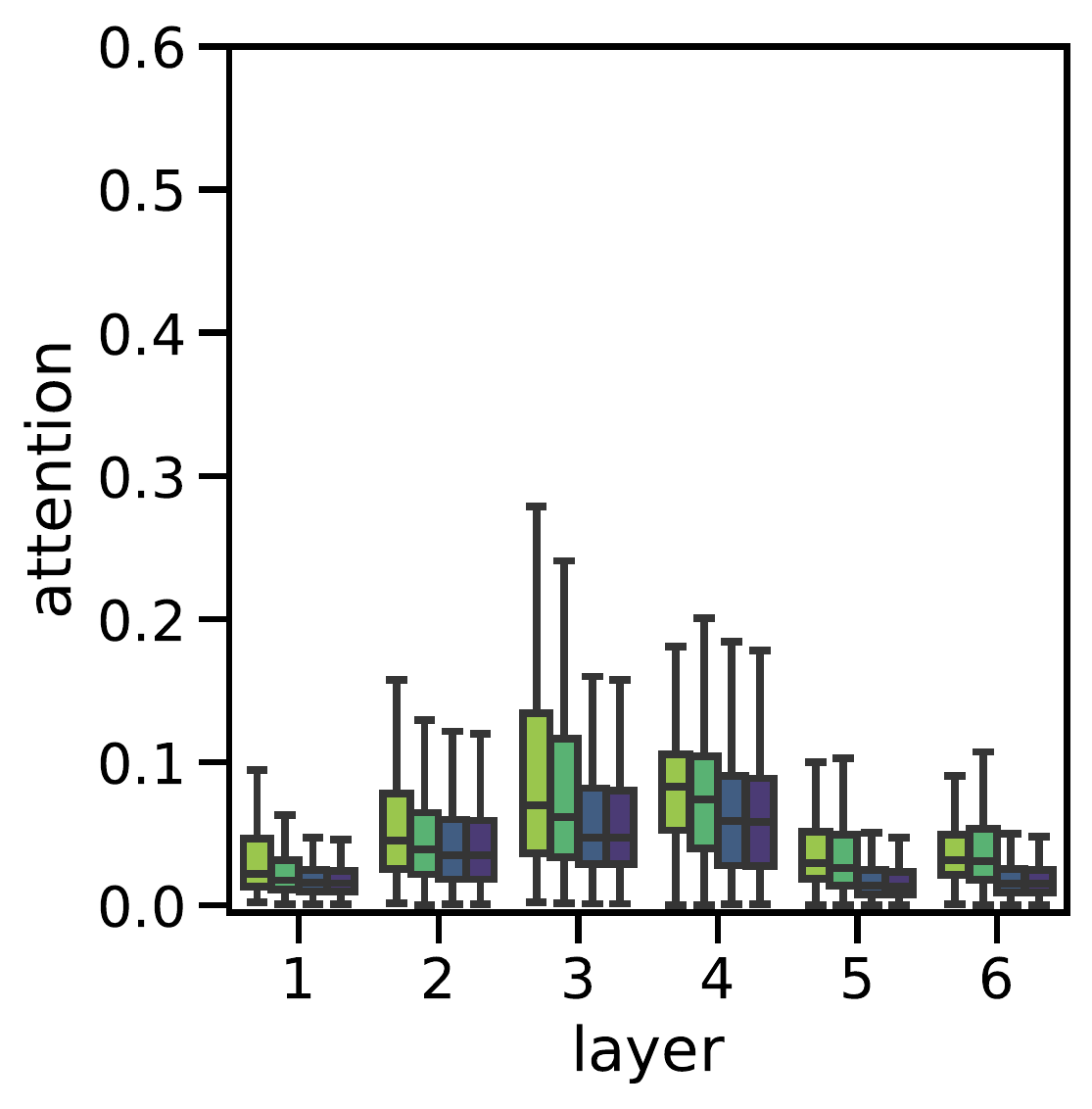}
    \caption{Target - PIE other}
    \end{subfigure}
    \begin{subfigure}[b]{0.47\columnwidth}\centering
    \includegraphics[width=0.96\textwidth]{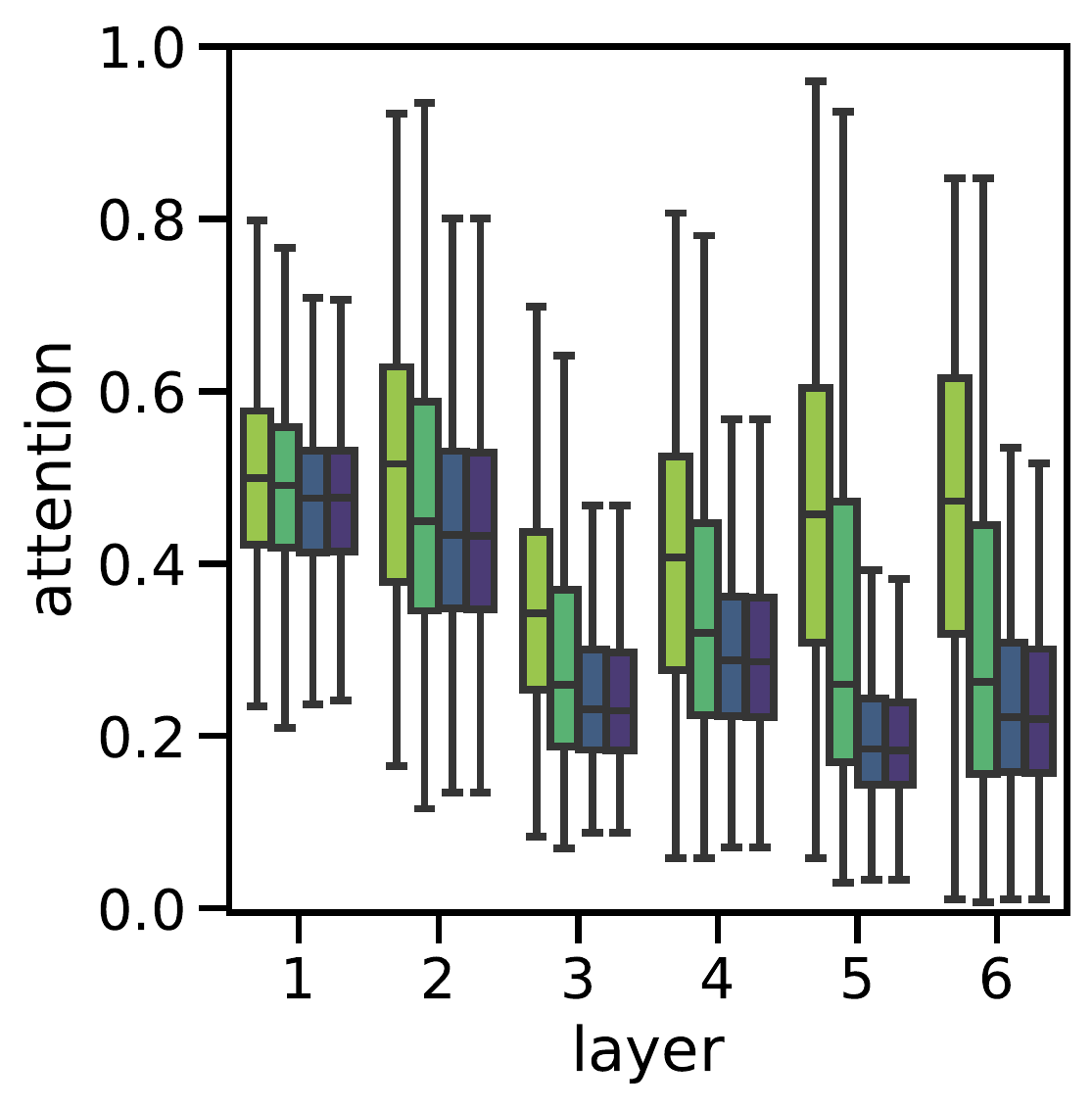}
    \caption{Target - \texttt{</s>}}
    \end{subfigure}
    \begin{subfigure}[b]{0.515\columnwidth}\centering
    \includegraphics[width=\textwidth]{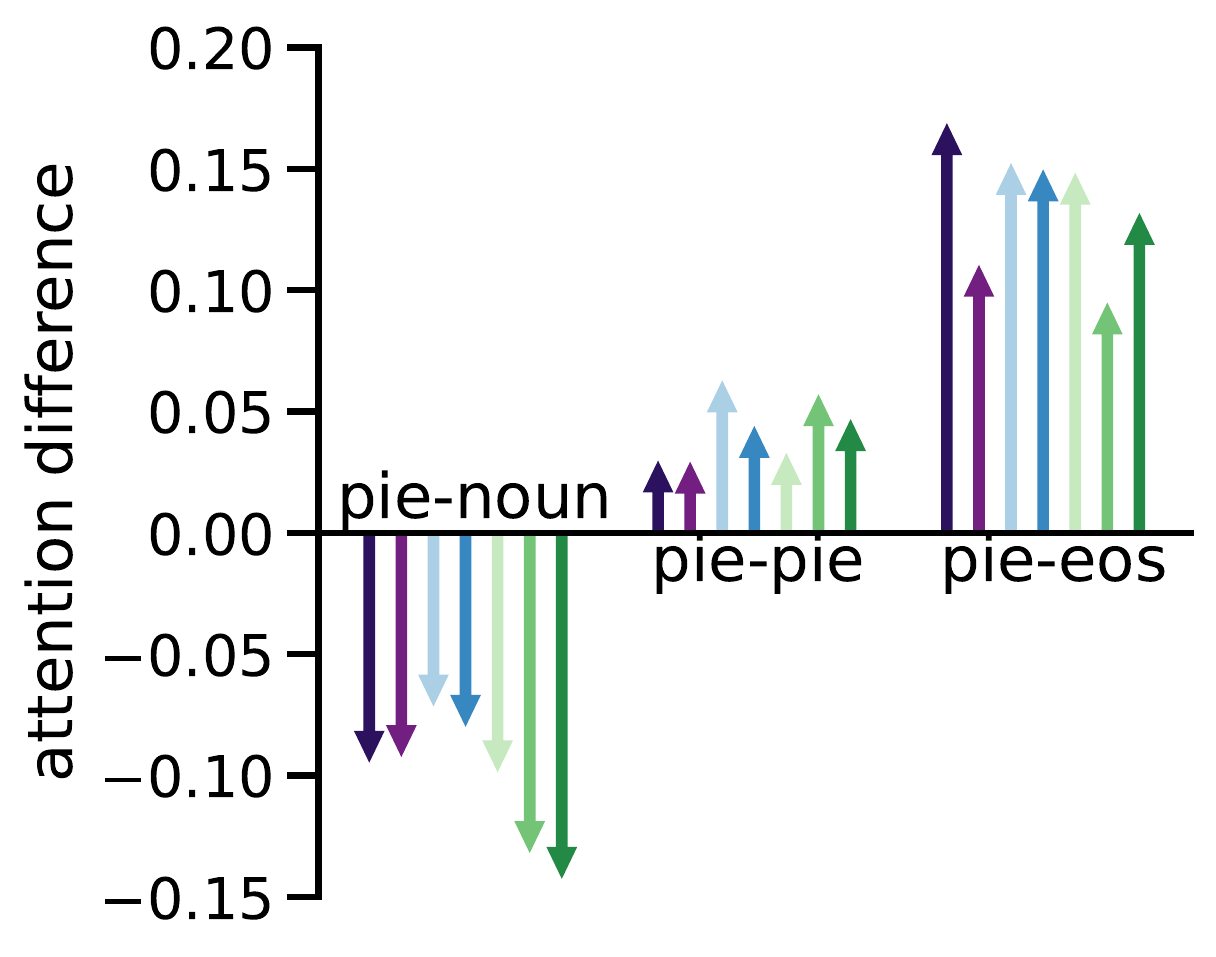}
    \caption{Language comparison}
    \end{subfigure}
\caption{Cross-attention distributions from the translation of one PIE noun on the target side to that noun on the source side, for the length controlled subset.}
\label{fig:ap_cross_attention_short}
\vspace{-.3cm}
\end{figure}

To assert that these differences do not substantially impact our qualitative findings, we compute the attention analyses over a data subset that only uses sentences where there are three tokens annotated for the PIE, for which the start and end are three positions apart.
This covers a subset of approximately 7k samples, with small variations between languages due to slightly different tokenisation of the English words. Figures~\ref{fig:ap_encoder_attention_short} and~\ref{fig:ap_cross_attention_short} present the results for the encoder's self-attention and the encoder-decoder cross-attention analyses, respectively.
Qualitatively, our findings for this subset do not differ from the previous findings.

\section{Results for 7 languages, per layer}
\label{ap:languages_per_layer}

\begin{figure}[!h]\centering
    \begin{subfigure}[b]{\columnwidth}\centering
    \includegraphics[width=0.8\columnwidth]{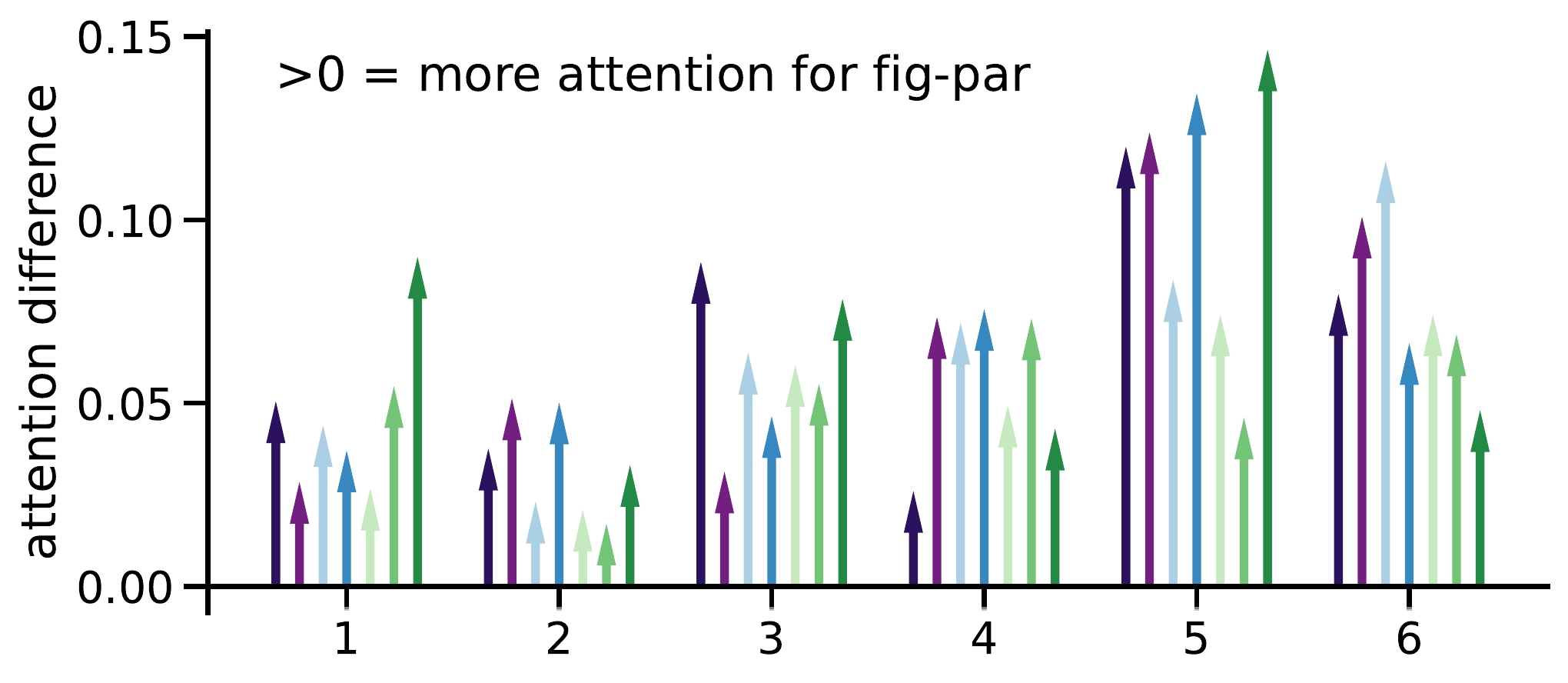}
    \caption{Attention from PIE tokens to PIE noun}
    \end{subfigure}
    \begin{subfigure}[b]{\columnwidth}\centering
    \includegraphics[width=0.8\columnwidth]{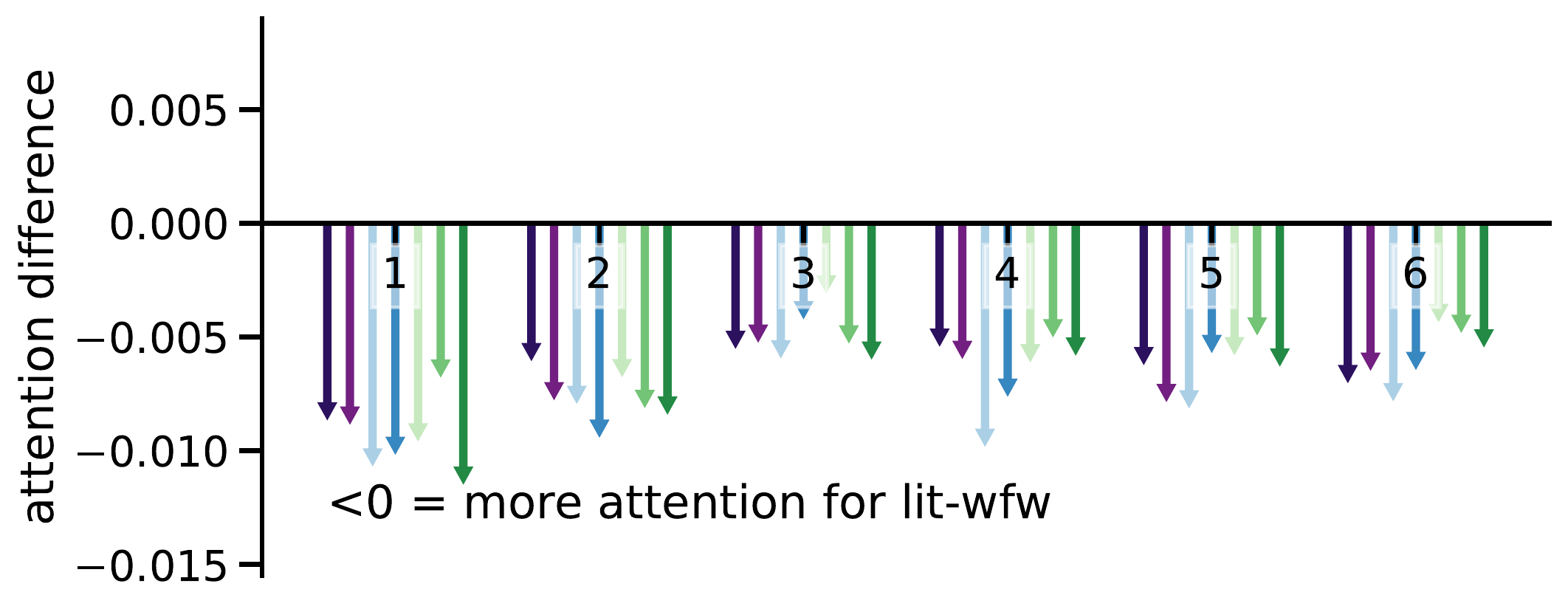}
    \caption{Attention from PIE tokens to context}
    \end{subfigure}
    \begin{subfigure}[b]{\columnwidth}\centering
    \includegraphics[width=0.8\columnwidth]{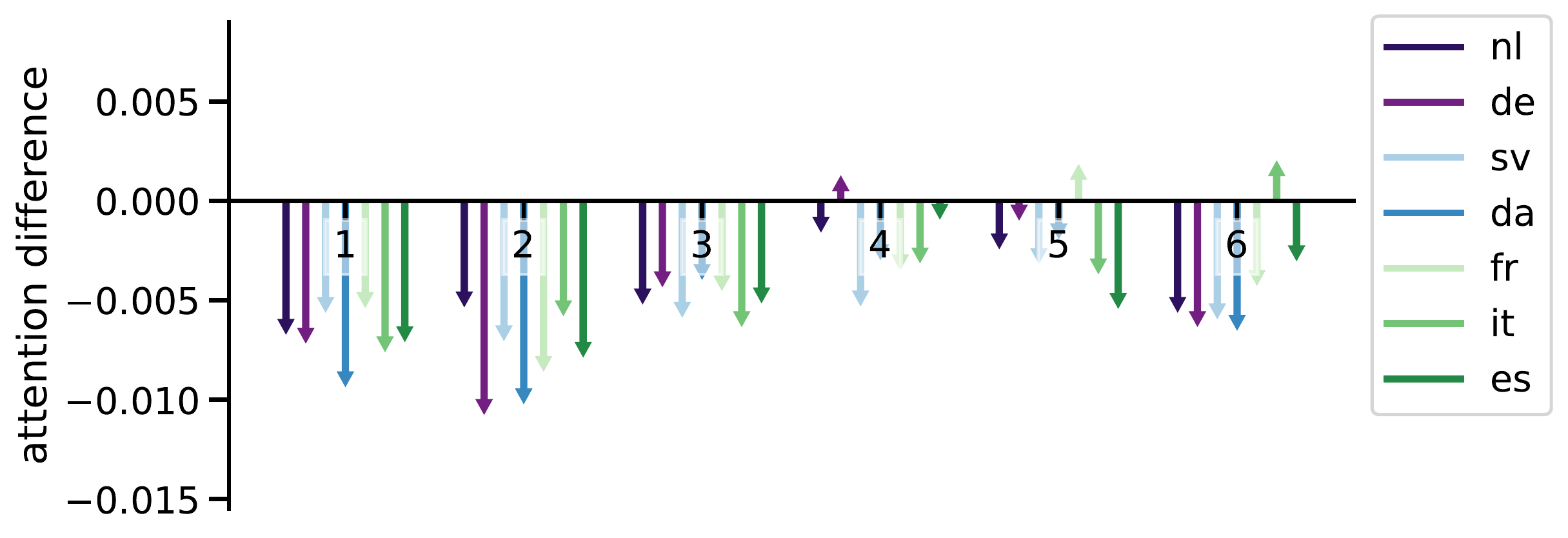}
    \caption{Attention from context to PIE}
    \end{subfigure}
    \centering
    \caption{The differences in attention between \textit{fig-par} and \textit{lit-wfw} visualised per layer, per language.}
    \label{fig:ap_languages_attention}
\end{figure}

\begin{figure}[!h]\centering
    \begin{subfigure}[b]{\columnwidth}\centering
    \includegraphics[width=0.8\columnwidth]{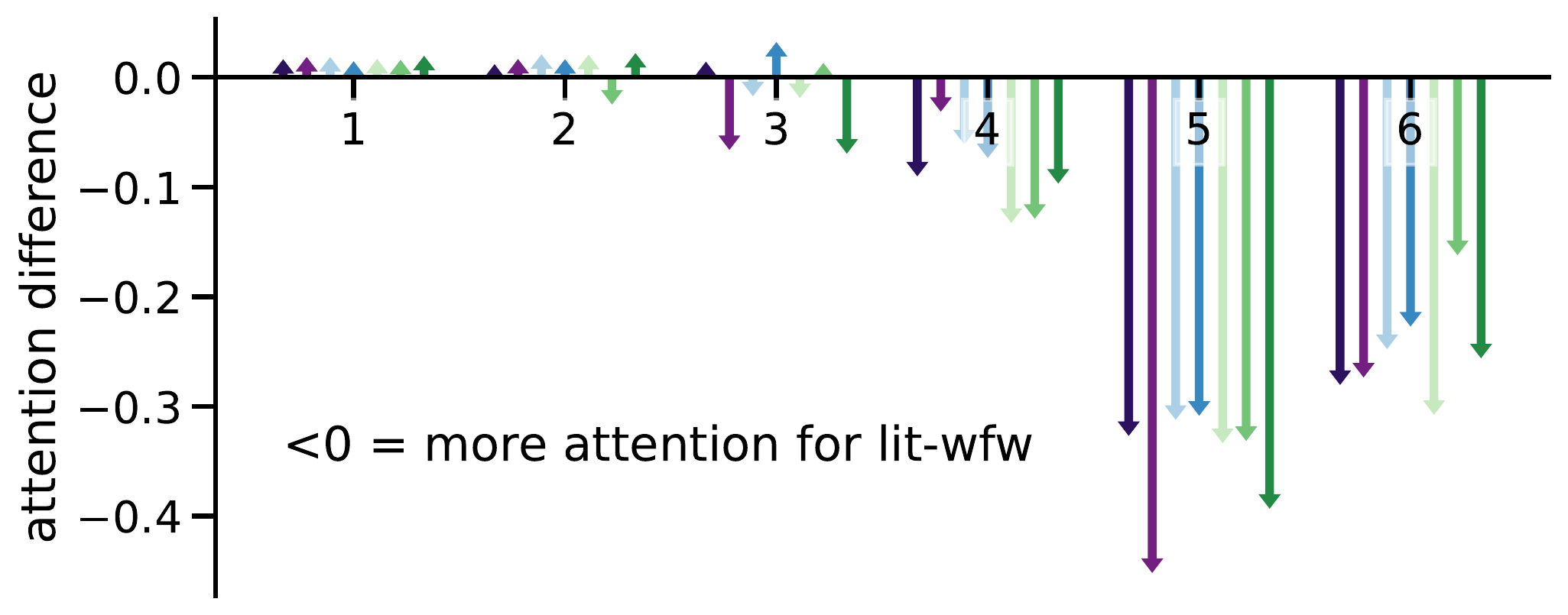}
    \caption{Cross-attention from target PIE noun to source PIE noun}
    \end{subfigure}
    \begin{subfigure}[b]{\columnwidth}\centering
    \includegraphics[width=0.8\columnwidth]{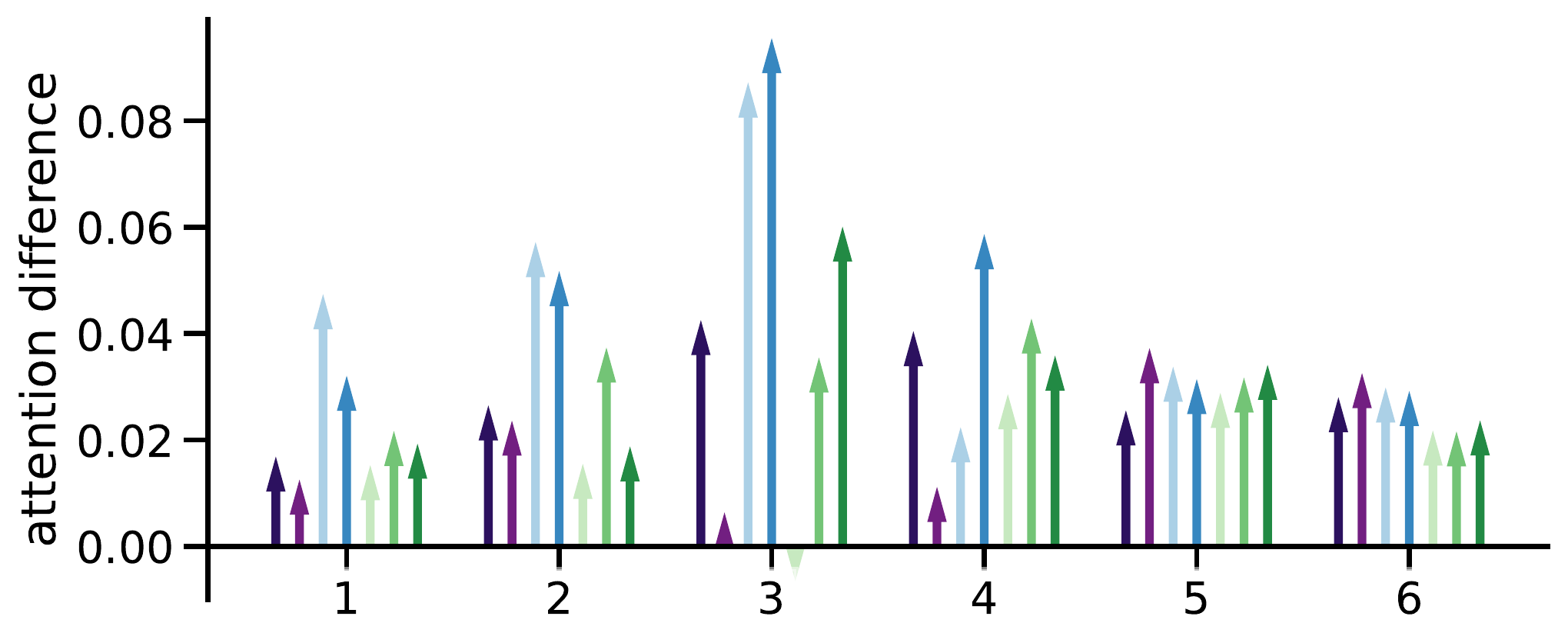}
    \caption{Cross-attention from target PIE noun to rest source PIE}
    \end{subfigure}
    \begin{subfigure}[b]{\columnwidth}\centering
    \includegraphics[width=0.8\columnwidth]{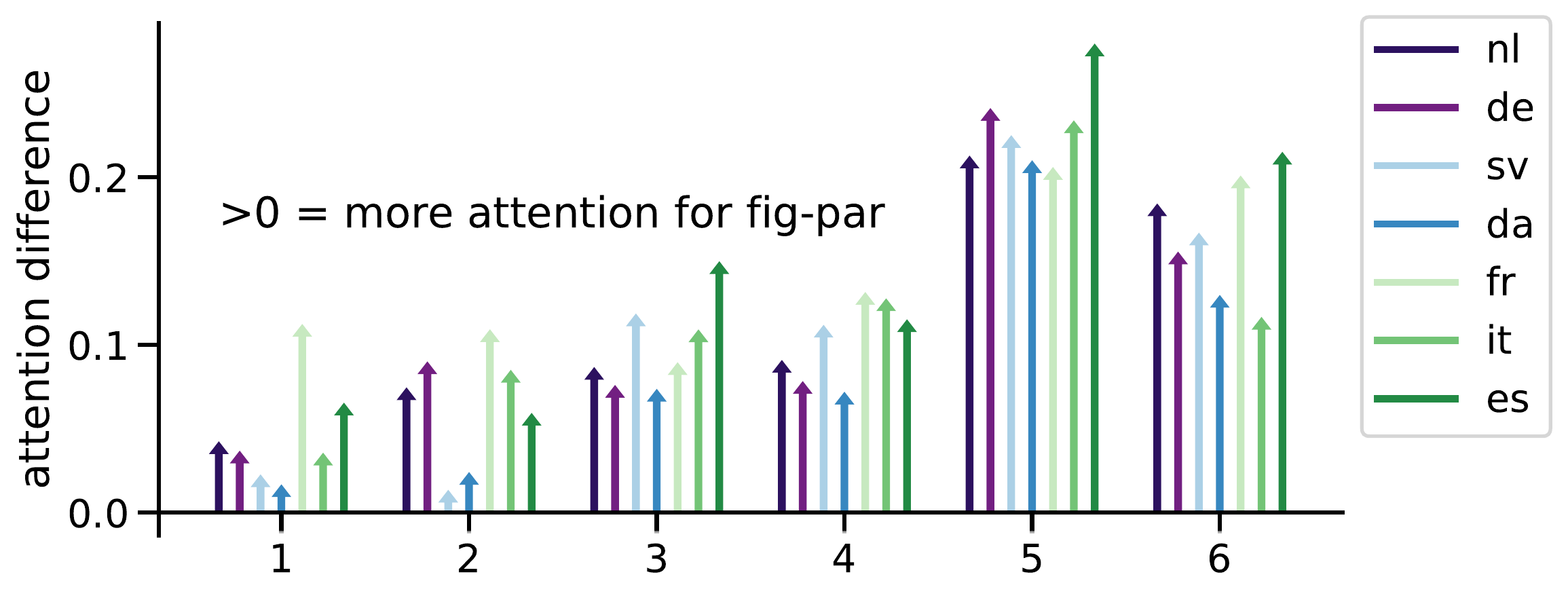}
    \caption{Cross-attention from target PIE noun to \texttt{</s>}}
    \end{subfigure}
    \centering
    \caption{The differences in cross-attention between \textit{fig-par} and \textit{lit-wfw} visualised per layer, per language.}
    \label{fig:ap_languages_cross_attention}
\end{figure}

\begin{figure}[!h]\centering
    \begin{subfigure}[b]{\columnwidth}\centering
    \includegraphics[width=0.8\columnwidth]{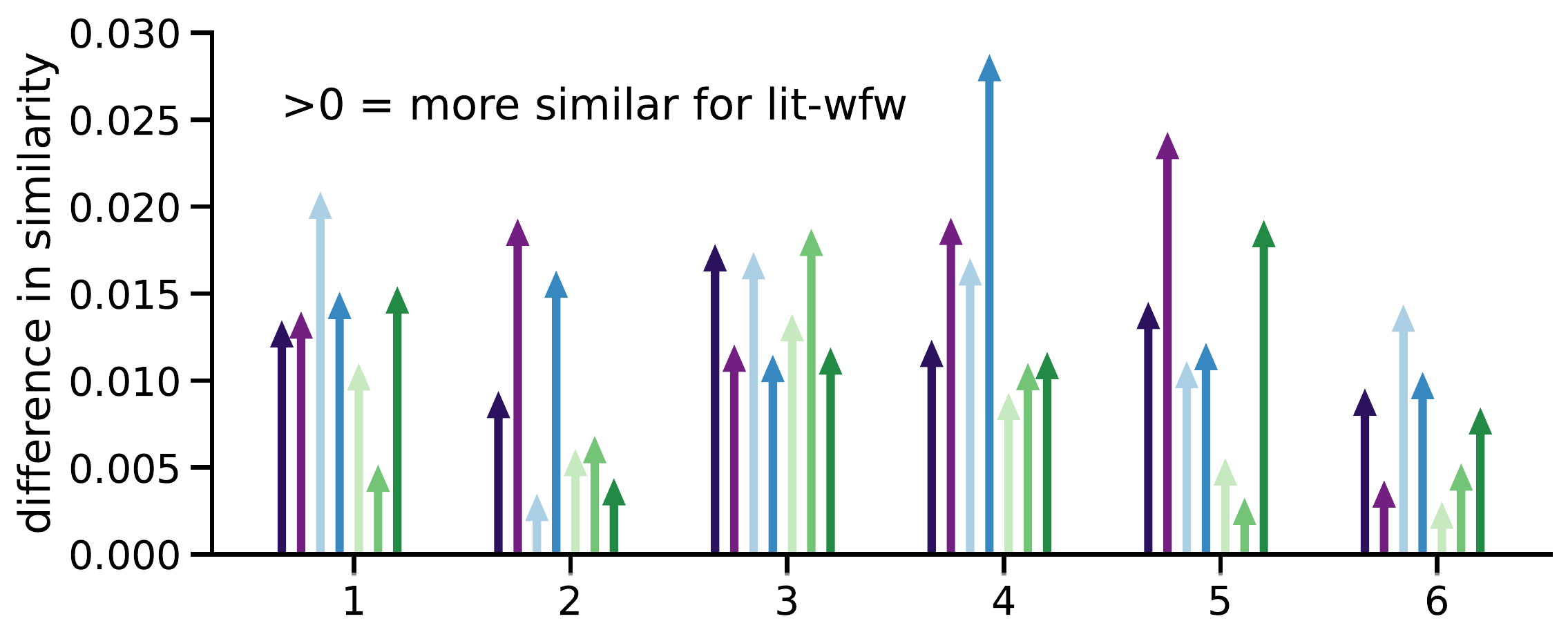}
    \caption{Impact on PIE tokens when masking PIE noun}
    \end{subfigure}
    \begin{subfigure}[b]{\columnwidth}\centering
    \includegraphics[width=0.8\columnwidth]{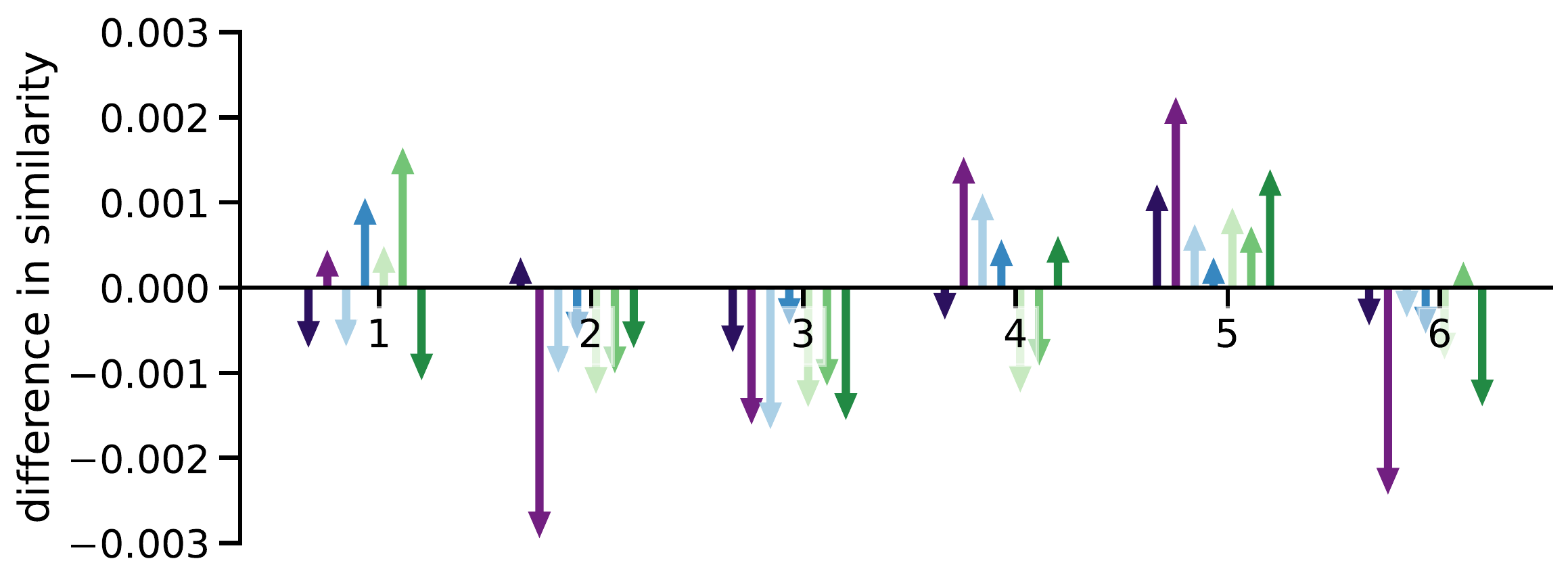}
    \caption{Impact on the context when masking PIE noun}
    \end{subfigure}
    \begin{subfigure}[b]{\columnwidth}\centering
    \includegraphics[width=0.8\columnwidth]{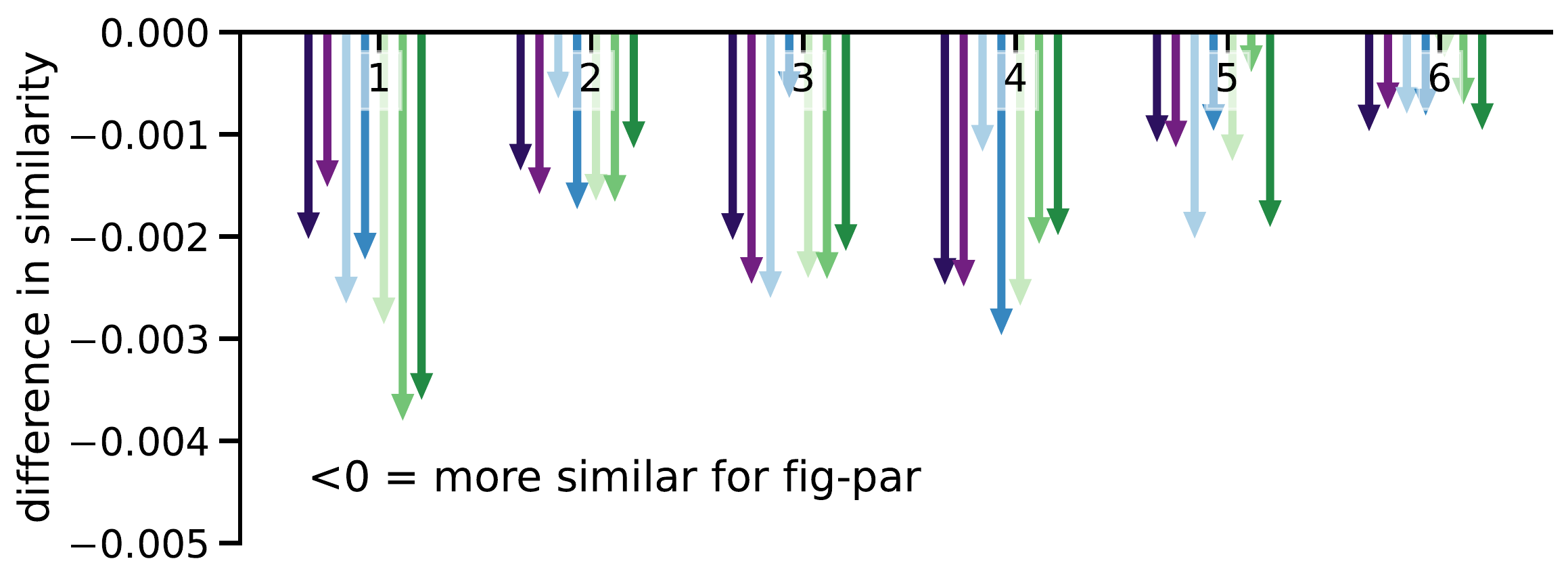}
    \caption{Impact on the PIE token when masking in the context}
    \end{subfigure}
    \begin{subfigure}[b]{\columnwidth}\centering
    \includegraphics[width=0.8\columnwidth]{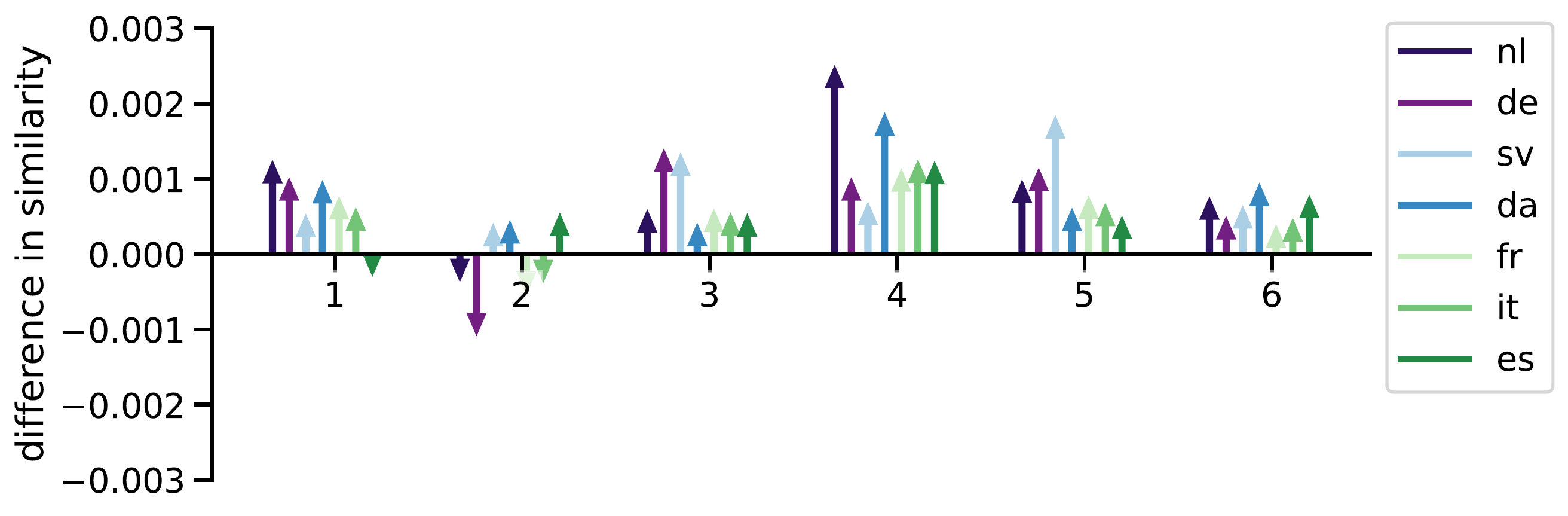}
    \caption{Impact on the context token when masking in the context}
    \end{subfigure}
    \centering
    \caption{The differences in CCA similarity between \textit{lit-wfw} and \textit{fig-par} visualised per layer, per language. Here, ``more similar'' means that the impact of masking is \underline{smaller}.}
    \label{fig:ap_languages_svcca}
    \vspace{-0.5cm}
\end{figure}

Figures~\ref{fig:ap_languages_attention} and~\ref{fig:ap_languages_cross_attention} present the results per layer, for the (cross-)attention graphs from \S\ref{sec:attention}.
Figure~\ref{fig:ap_languages_svcca} present the results per layer, for the CCA similarity graphs from \S\ref{sec:hidden_states}.

\clearpage

\section{Two-step CCA}
\label{ap:cca}

CCA can be used to compare representations over different layers of the same network or different networks in a way that is invariant to affine transformations \citep{raghu2017svcca}.
The CCA similarity expresses the extent to which two representations contain the same information while accounting for transformations in these two views of the data. Nonetheless, the similarity depends on the data used to perform CCA. Even with a dataset that is at least an order of magnitude larger than the number of dimensions in the hidden representations, the composition of the dataset impacts the outcome. Particularly relevant in the context of our work is the vocabulary size that impacts CCA computations.

\begin{figure}[!h]
\centering
\begin{subfigure}[b]{0.49\columnwidth}
    \centering
    \includegraphics[width=\textwidth]{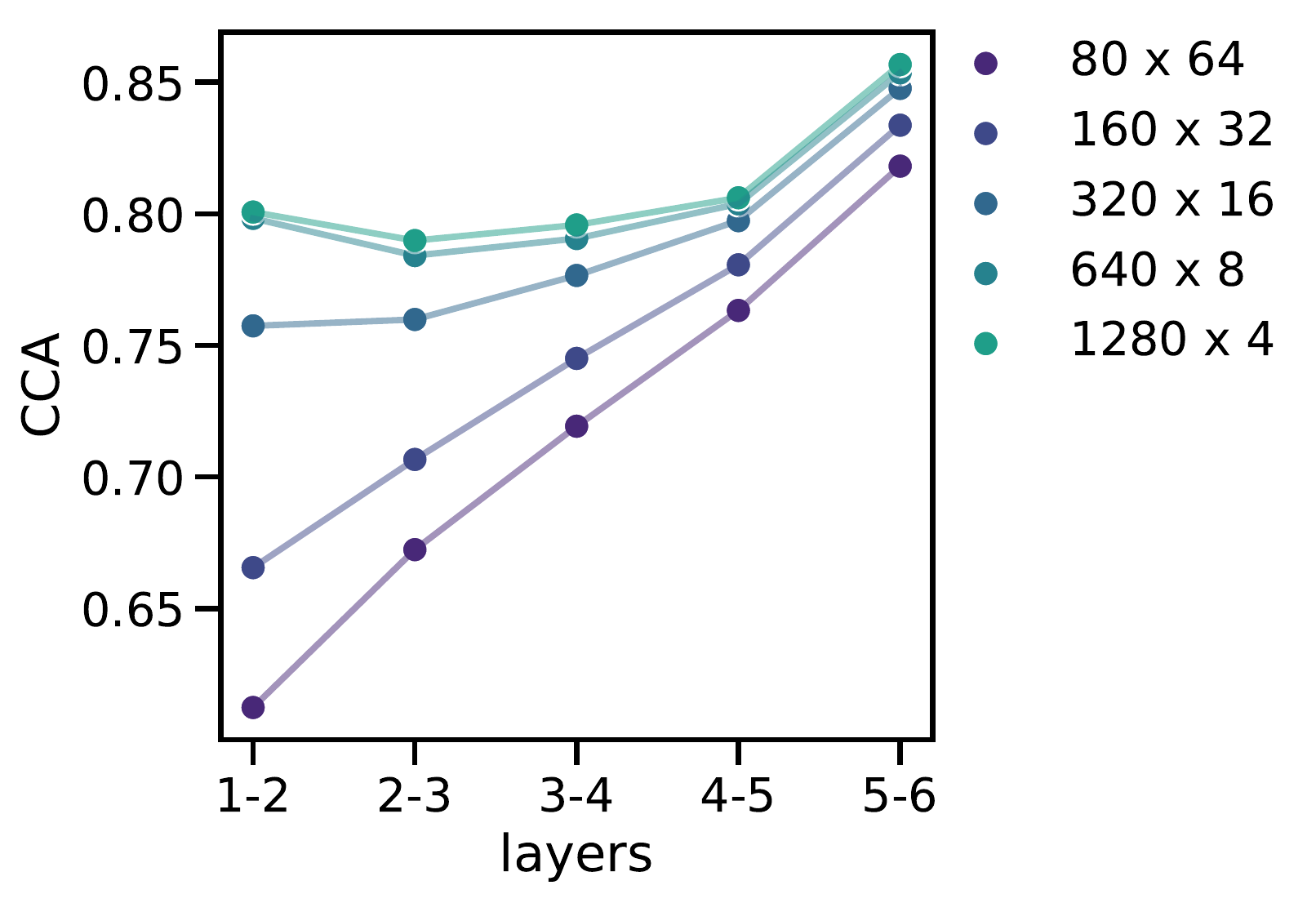}
    \caption{CCA per subset}
    \label{fig:cca}
\end{subfigure}
\begin{subfigure}[b]{0.49\columnwidth}
    \centering
    \includegraphics[width=\textwidth]{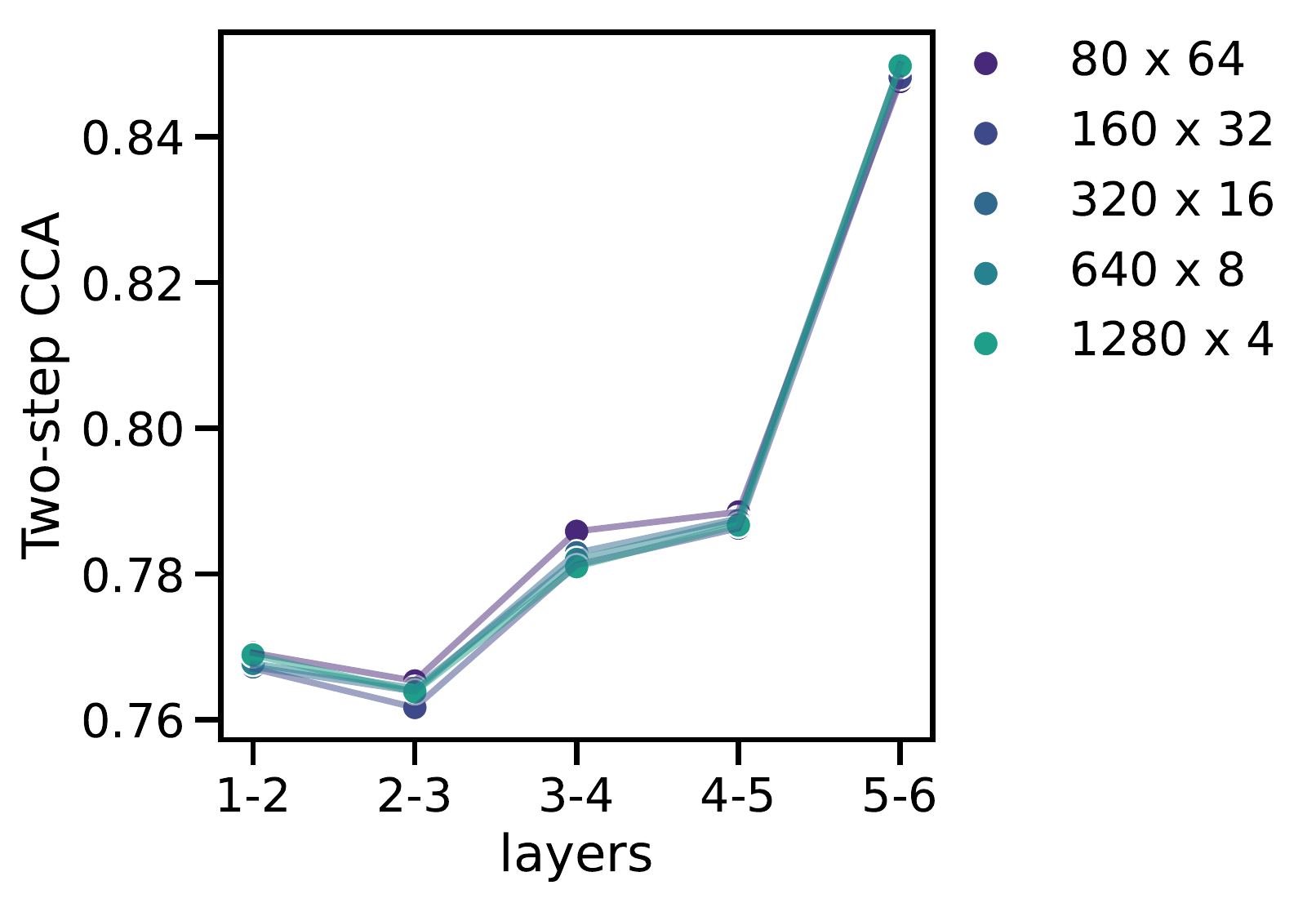}
    \caption{Two-step CCA}
    \label{fig:two_step_cca}
\end{subfigure}
\caption{Illustration of the impact of recomputing CCA with data subsets of differently composed vocabularies for a dataset size of 5k.}
\end{figure}

We illustrate this by measuring how hidden representations change over layers, randomly sampling tokens and considering multiple dataset compositions, varying from 64 occurrences of 80 unique tokens, to 4 occurrences of 1280 unique tokens. Recomputing CCA per subset yields the similarities shown in Figure~\ref{fig:cca}. Although the overall pattern of lower similarity between lower layers and higher similarity between higher layers is present for all subsets, the absolute similarity measures differ between subsets. In Figure~\ref{fig:two_step_cca}, however, where the projection matrix is computed on a separate dataset, subsets show comparable similarities. The differences between the methods decrease as the number of hidden representations used to perform CCA grows.

Performing CCA separately per (relatively small) subset of the MAGPIE corpus could thus reflect vocabulary differences rather than systematic differences due to figurativeness. We merely want to apply CCA to account for differences between layers and differences with and without masking attention, and thus apply two-step CCA, computing projection matrices on a separate dataset.

\section{Amnesic probing}
\label{ap:amnesic_probing}

\begin{figure}[b]
    \centering
    \begin{subfigure}[b]{\columnwidth}\centering
    \includegraphics[width=0.9\textwidth]{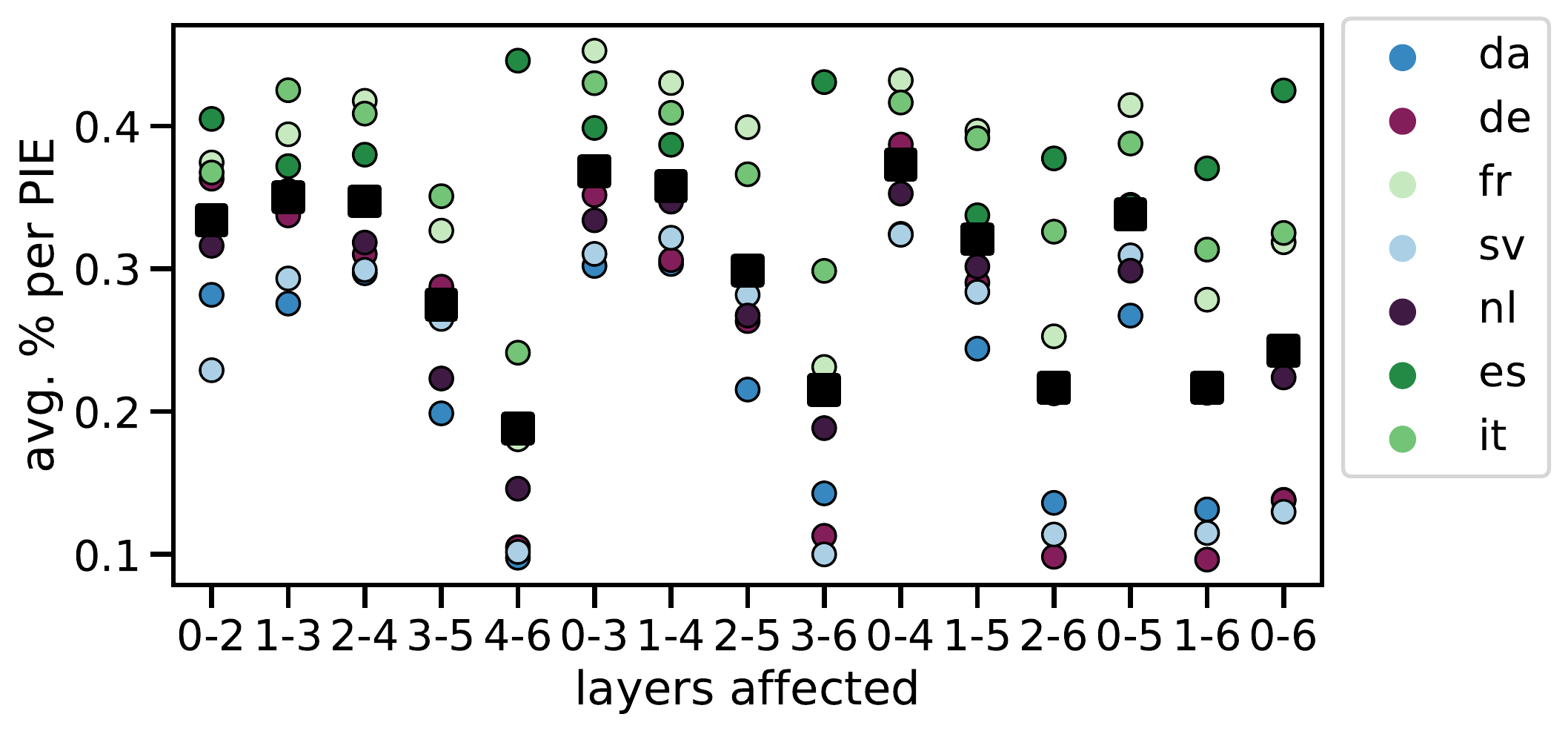}
    \end{subfigure}
    \caption{Impact of the selection of layers affected by INLP. Dots represented different languages, the squares indicate the mean \%.}
    \label{fig:inlp_params}
\end{figure}

Amnesic probing \citep{elazar2021amnesic} evaluates the behavioural influence of information recovered from hidden representations $H$ by probes, by removing that information from the representation and measuring the change in behaviour on the main task.
INLP, proposed by \citet{ravfogel-etal-2020-null}, is used to remove this information from the representations, by training $k$ classifiers to predict a property from input vectors. After training probe $i$, parametrised by $W_i$, the vectors are projected onto the nullspace of $W_i$, using projection matrix $P_{N(W_i)}$, such that $W_iP_{N(W_i)}H=0$.
The projection matrix of the intersection of all $k$ null spaces can then remove features found by the $k$ classifiers.

Using INLP, we train $50$ classifiers to detect figurative, paraphrased PIEs from figurative PIEs translated word for word from the hidden state. Afterwards, we apply the projection matrices while the model processes previously paraphrased translations. We separate the PIEs into five folds, using one for parameter estimation. For every fold $\frac{3}{5}$ is used to train INLP's probes, $\frac{1}{5}$ is used to measure whether the performance of the $k$ probes decreases and $\frac{1}{5}$ is used to measure the changed percentage. Dependent on where one intervenes in the model, amnesic probing may be more or less successful, since not every layer encodes the linguistic property and higher layers could recover information removed from lower layers \citep{elazar2021amnesic}. The parameter estimation performed measures the impact of different combinations of layers as the average success rate per PIE type (success means achieving a word-for-word translation).
As shown in Figure~\ref{fig:inlp_params}, there is quite some variation among languages, but generally intervening in the lower layers of Transformer is the most successful, and including the sixth layer is quite detrimental.
The results in the main body of the paper are computed by intervening on the hidden states of PIE tokens in $l\in\{0, 1, 2, 3, 4\}$.

\section{Idioms in OPUS}
\label{ap:idioms_in_opus}
To understand whether the model's translations reflect target translations from its training corpus, we extract up to 500 identical matches per idiom from OPUS for the \texttt{En-Nl} model.
These target translations are labelled heuristically, resulting in 54\% of paraphrased instances, which is substantially higher than the percentage of paraphrased instances in the model's translations.
This may be the result of infrequent idioms contained in OPUS, for which the model fails to learn the correct implicit meaning, even though the corpus does provide paraphrases.
Table~\ref{tab:opus} illustrates how the predicted translations' labels relate to the labels of target translations and provides BLEU scores per subset. Samples with a paraphrased target translation score substantially lower compared to those with a word-for-word or copied target translation, emphasising the negative impact of idioms on translation quality.

\begin{table}[!h]
    \centering\small
    \begin{tabular}{lc|ccc}
    \toprule
    \textbf{OPUS} &  & \multicolumn{2}{c}{\textbf{Predicted Translations}} \\
    & & \textit{Paraphrase} & \textit{Word for word} \\ \midrule 
    \multicolumn{4}{l}{(a) Translation type frequency (\%)}\\
    Paraphrase   & 54 & 49 & 51  \\
    Word for word   & 46 & 7 & 93 \\ \midrule
    \multicolumn{4}{l}{(b) BLEU scores} \\
    Paraphrase   && 27.2 & 19.9  \\
    Word for word && 25.6 & 38.2 \\
    \bottomrule
    \end{tabular}
    \caption{Distribution of translation labels for idiom occurrences in OPUS, along with their BLEU scores.}
    \label{tab:opus}
    \vspace{-0.5cm}
\end{table}

\end{appendices}